\documentclass[runningheads]{llncs}
\usepackage{graphicx}

\usepackage{tikz}
\usepackage{comment}
\usepackage{amsmath,amssymb} 
\usepackage{color}
\usepackage{url}  
\usepackage[pagebackref=true,breaklinks=true,letterpaper=true,pagebackref=true,colorlinks,bookmarks=false]{hyperref}
\usepackage{booktabs,lipsum}
\usepackage{wrapfig}
\usepackage{mathtools}

\usepackage[accsupp]{axessibility}  

\usepackage[width=122mm,left=12mm,paperwidth=146mm,height=193mm,top=12mm,paperheight=217mm]{geometry}

\newcommand{\mpage}[2]
{
\begin{minipage}{#1\linewidth}\centering
#2
\end{minipage}
}

\newcommand{\topic}[1]
{
\vspace{1mm}\noindent\textbf{#1}
}
\newcommand{\figref}[1]{Figure~\ref{fig:#1}} 

\DeclareMathOperator*{\argmin}{\arg\!\min}

\begin{document}
\pagestyle{headings}
\mainmatter

\title{Temporally Consistent Semantic Video Editing} 

\author{Yiran Xu\inst{1} \and
Badour AlBahar\inst{2} \and
Jia-Bin Huang\inst{1}}
\institute{University of Maryland, College Park \and
Virginia Tech}

\maketitle
\vspace{-10mm}
\begin{figure}
\begin{center}
\centering

\mpage{0.01}{\raisebox{60pt}{\rotatebox{90}{Input}}} 
\frame{\includegraphics[trim=350 0 0 0, clip,width=.235\textwidth]{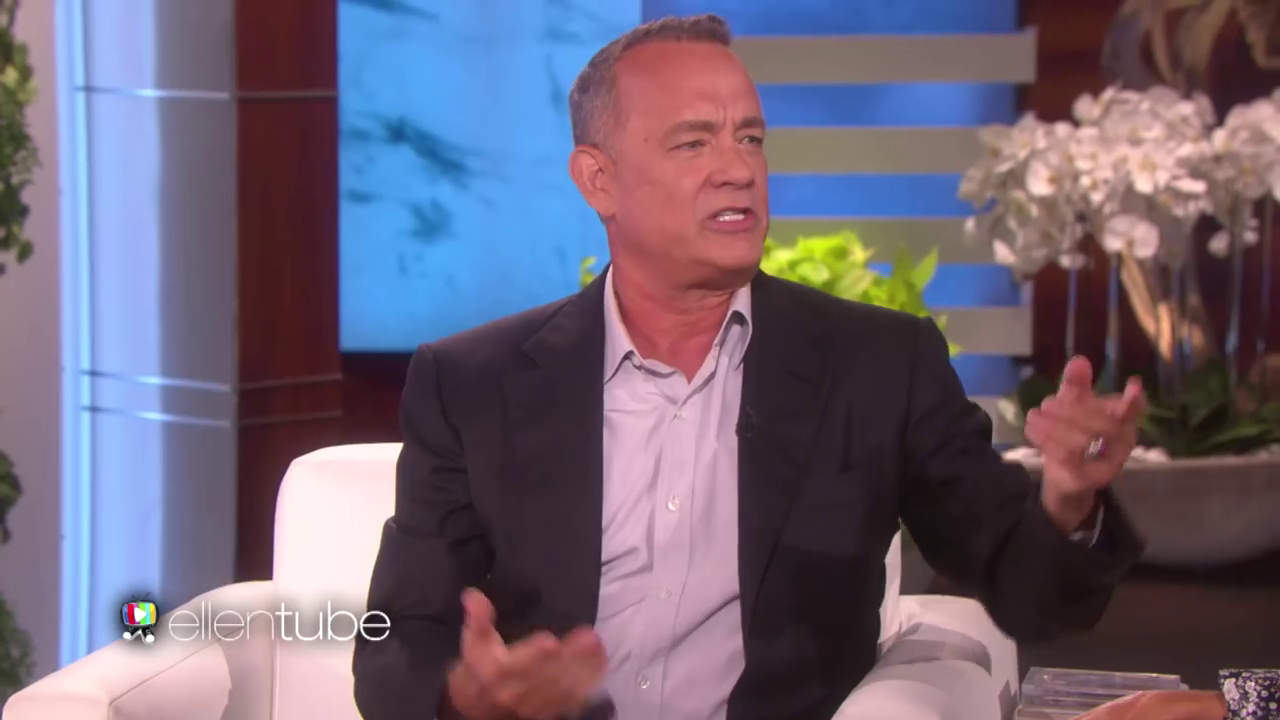}}\hspace{-20mm}\hfill
\frame{\includegraphics[trim=350 0 0 0, clip,width=.235\textwidth]{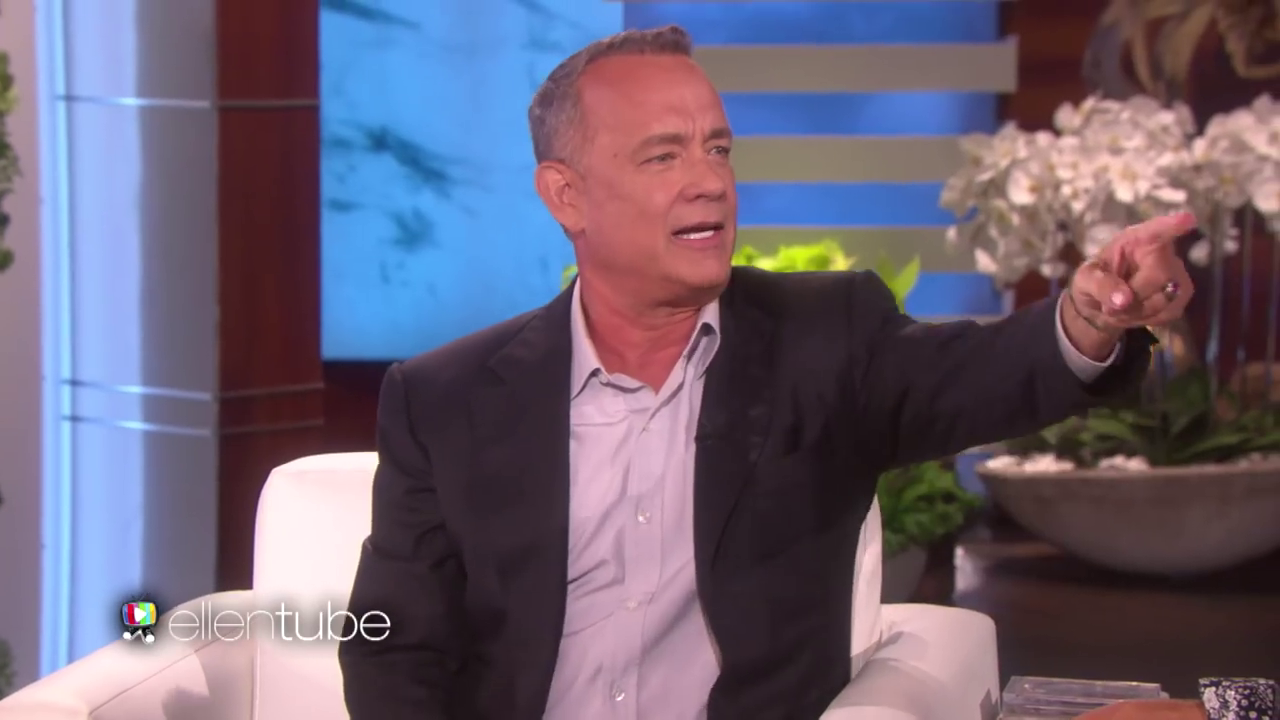}}\hspace{-20mm}\hfill
\frame{\includegraphics[trim=350 0 0 0, clip,width=.235\textwidth]{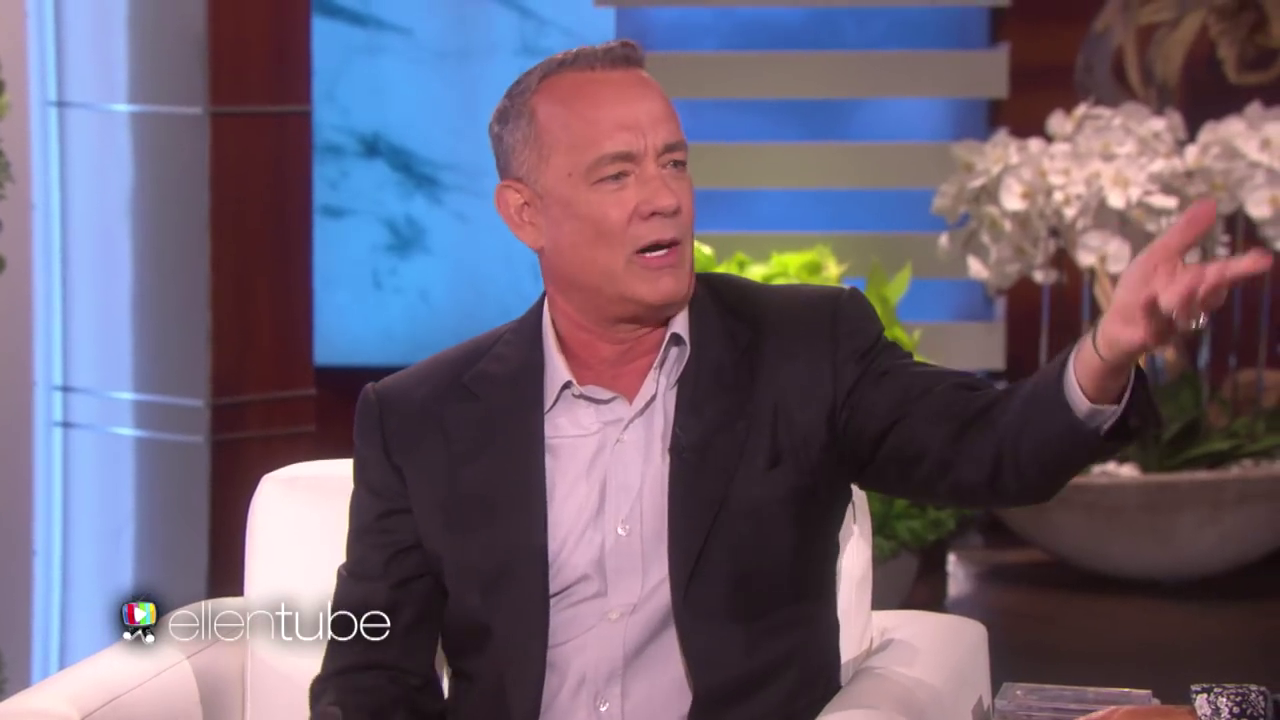}}\hspace{-20mm}\hfill
\frame{\includegraphics[trim=350 0 0 0, clip,width=.235\textwidth]{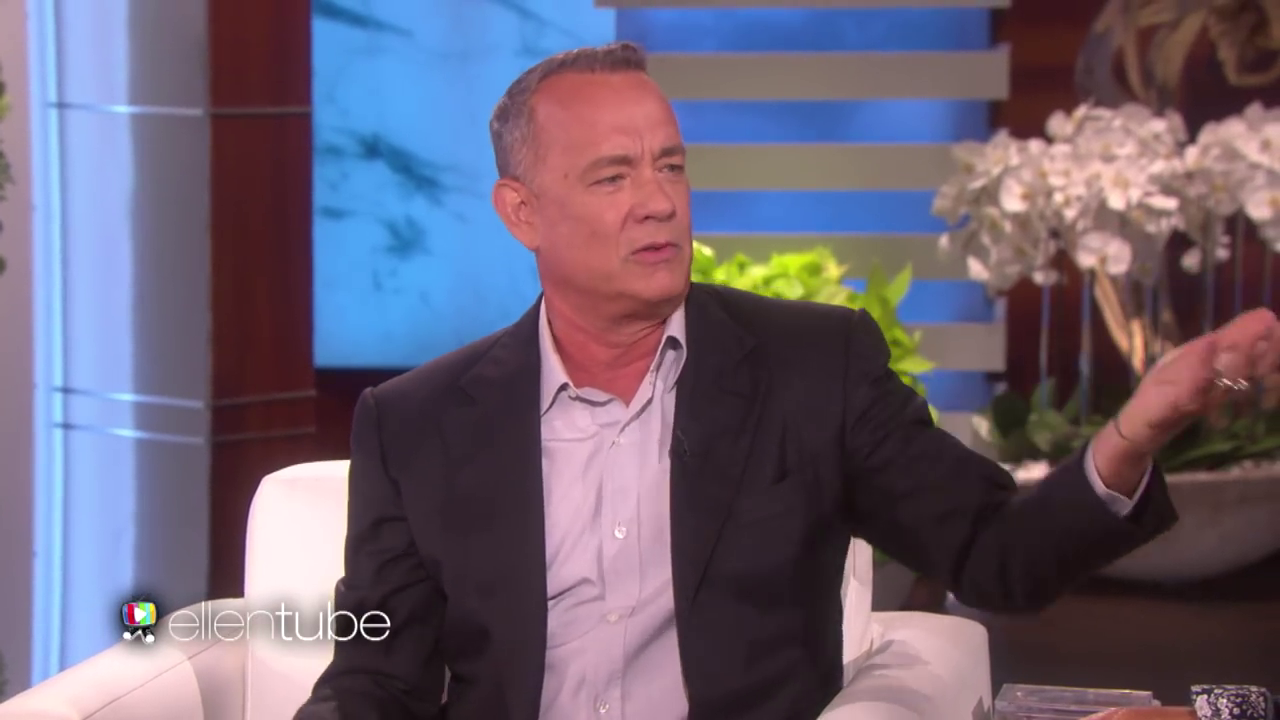}}\\

\vspace{-13mm}

\mpage{0.01}{\raisebox{60pt}{\rotatebox{90}{``angry''}}} 
\frame{\includegraphics[trim=350 0 0 0, clip,width=.235\textwidth]{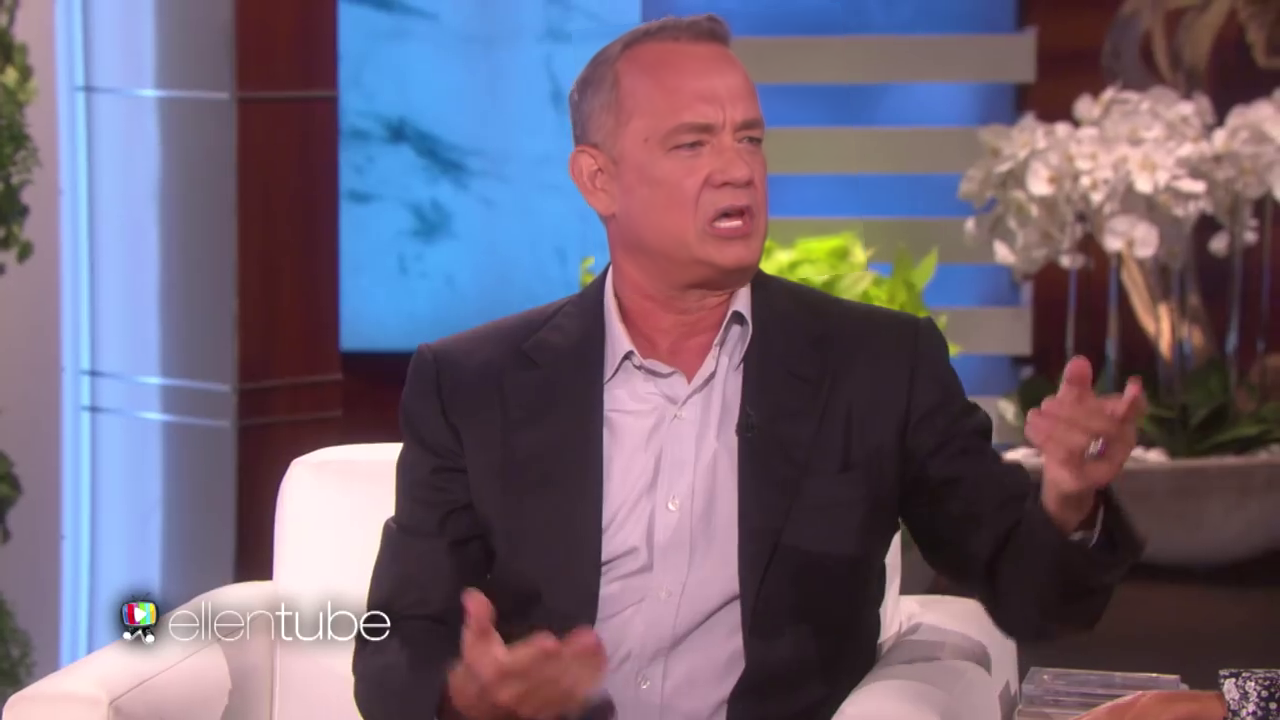}}\hspace{-20mm}\hfill
\frame{\includegraphics[trim=350 0 0 0, clip,width=.235\textwidth]{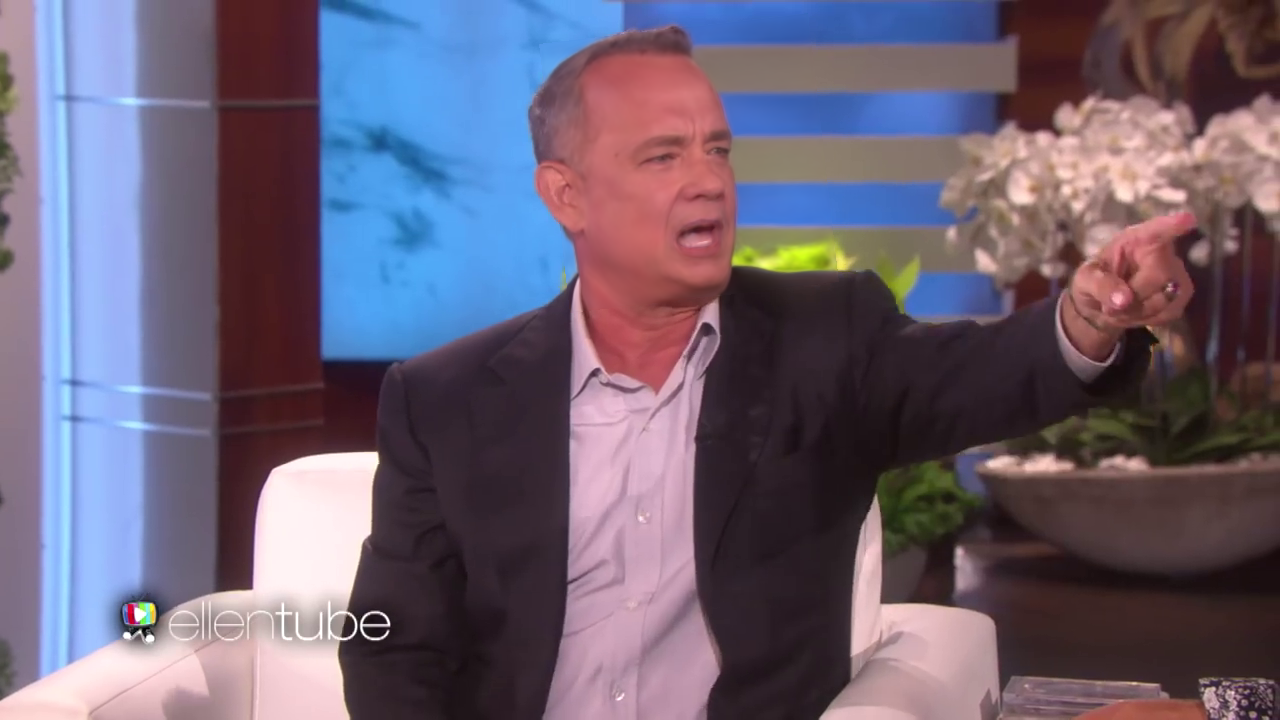}}\hspace{-20mm}\hfill
\frame{\includegraphics[trim=350 0 0 0, clip,width=.235\textwidth]{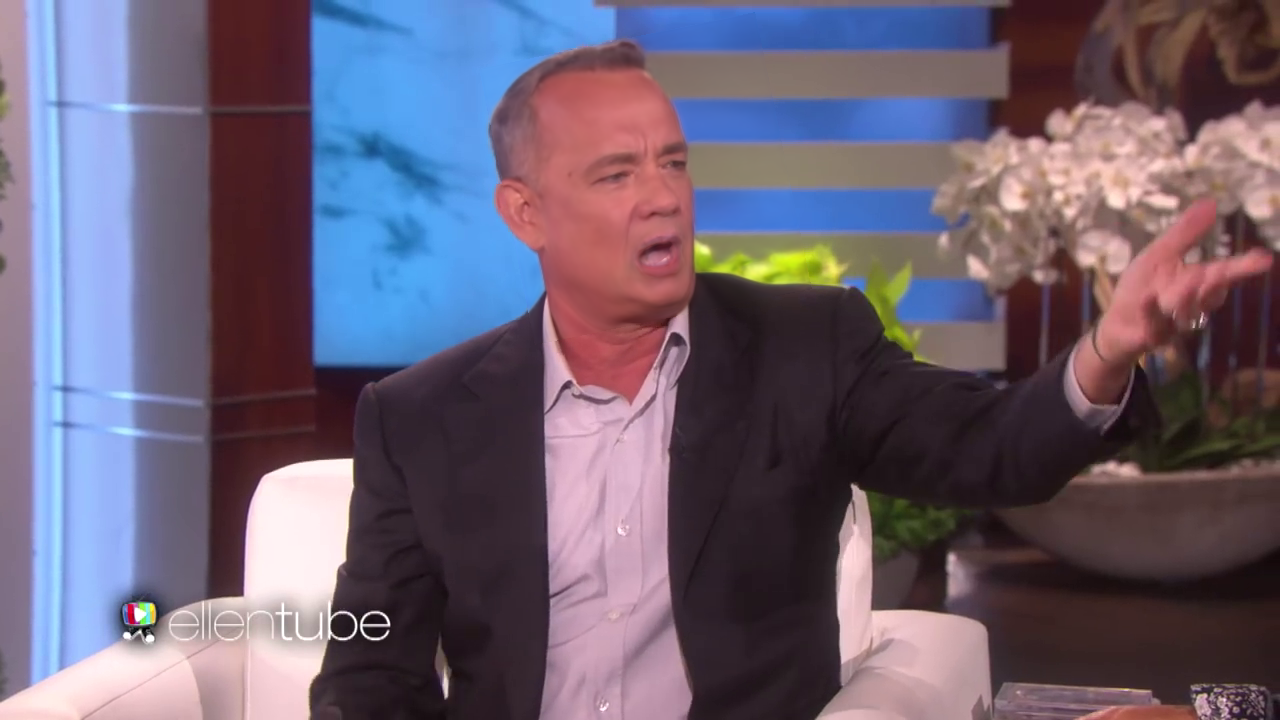}}\hspace{-20mm}\hfill
\frame{\includegraphics[trim=350 0 0 0, clip,width=.235\textwidth]{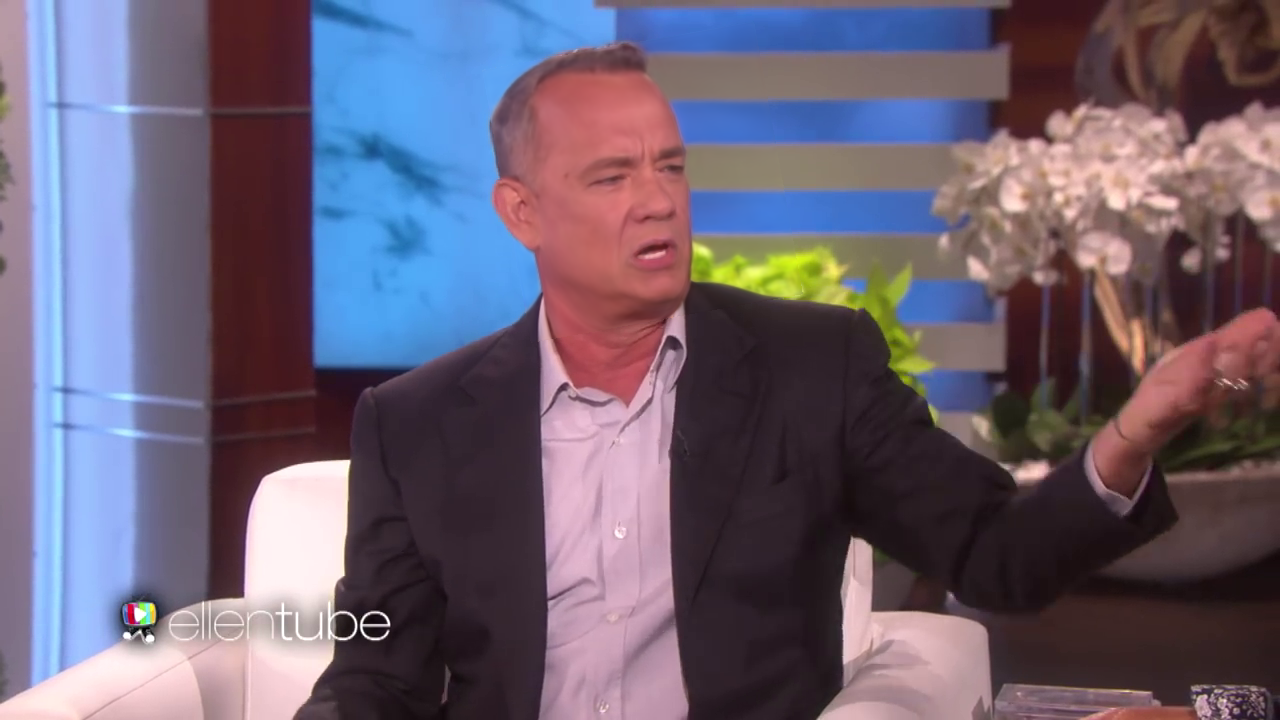}}\\

\vspace{-14.7mm}

\mpage{0.01}{\raisebox{60pt}{\rotatebox{90}{``eyeglasses''}}}
\frame{\includegraphics[trim=350 0 0 0, clip,width=.235\textwidth]{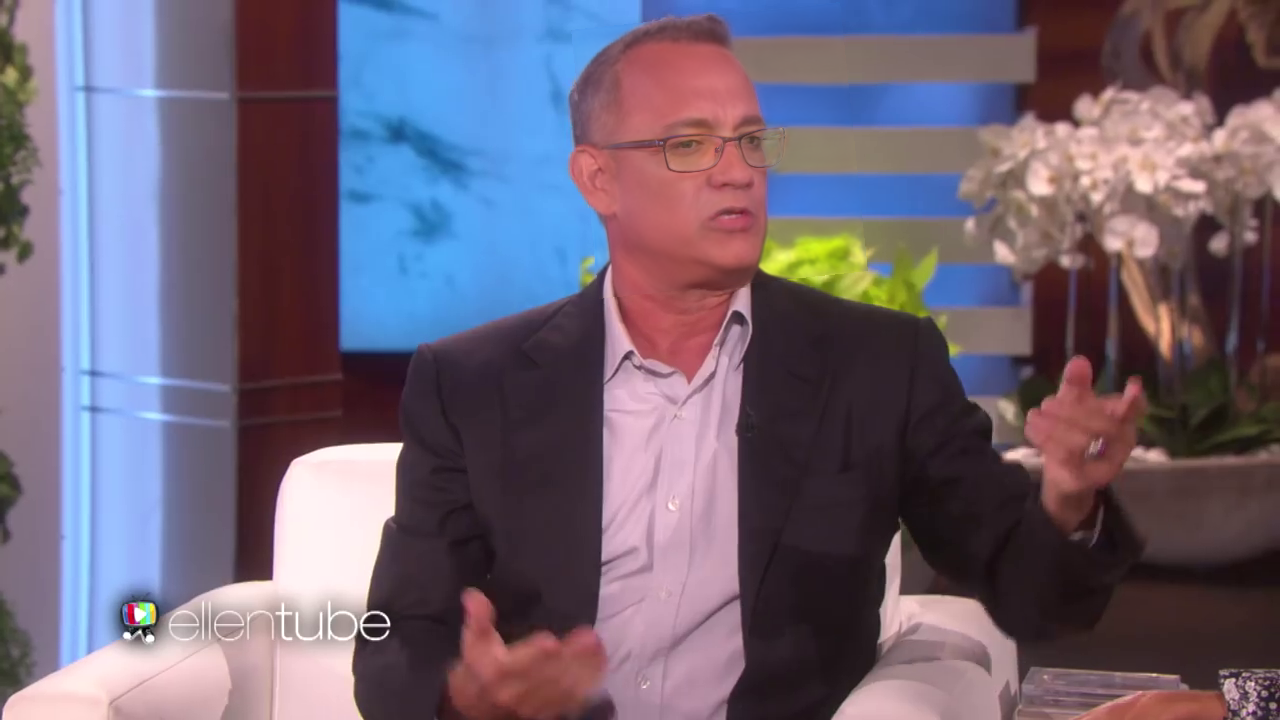}}\hspace{-20mm}\hfill
\frame{\includegraphics[trim=350 0 0 0, clip,width=.235\textwidth]{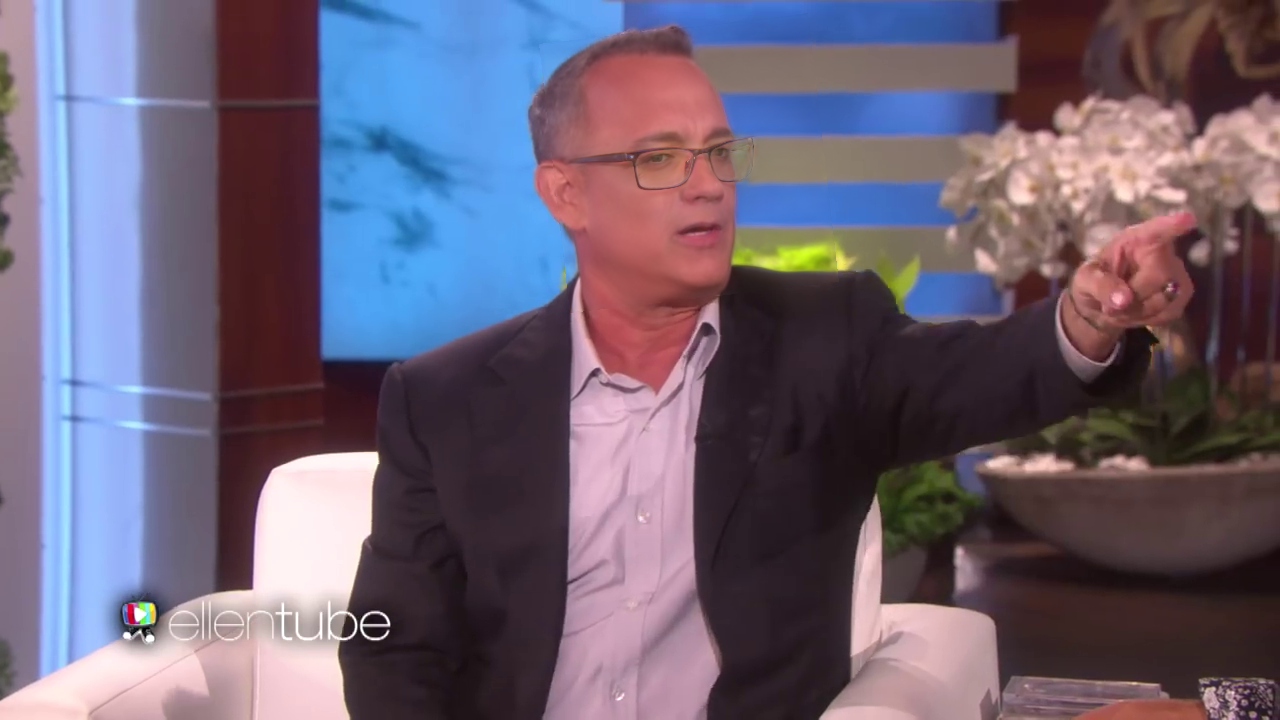}}\hspace{-20mm}\hfill
\frame{\includegraphics[trim=350 0 0 0, clip,width=.235\textwidth]{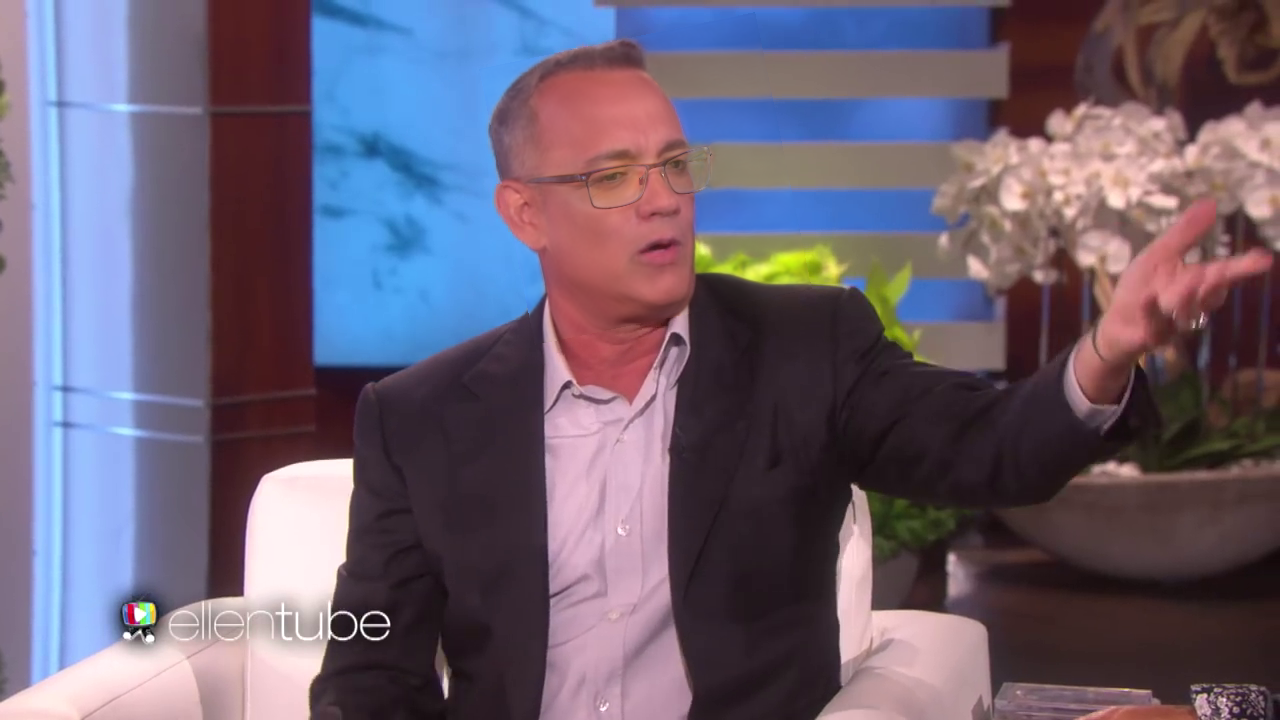}}\hspace{-20mm}\hfill
\frame{\includegraphics[trim=350 0 0 0, clip,width=.235\textwidth]{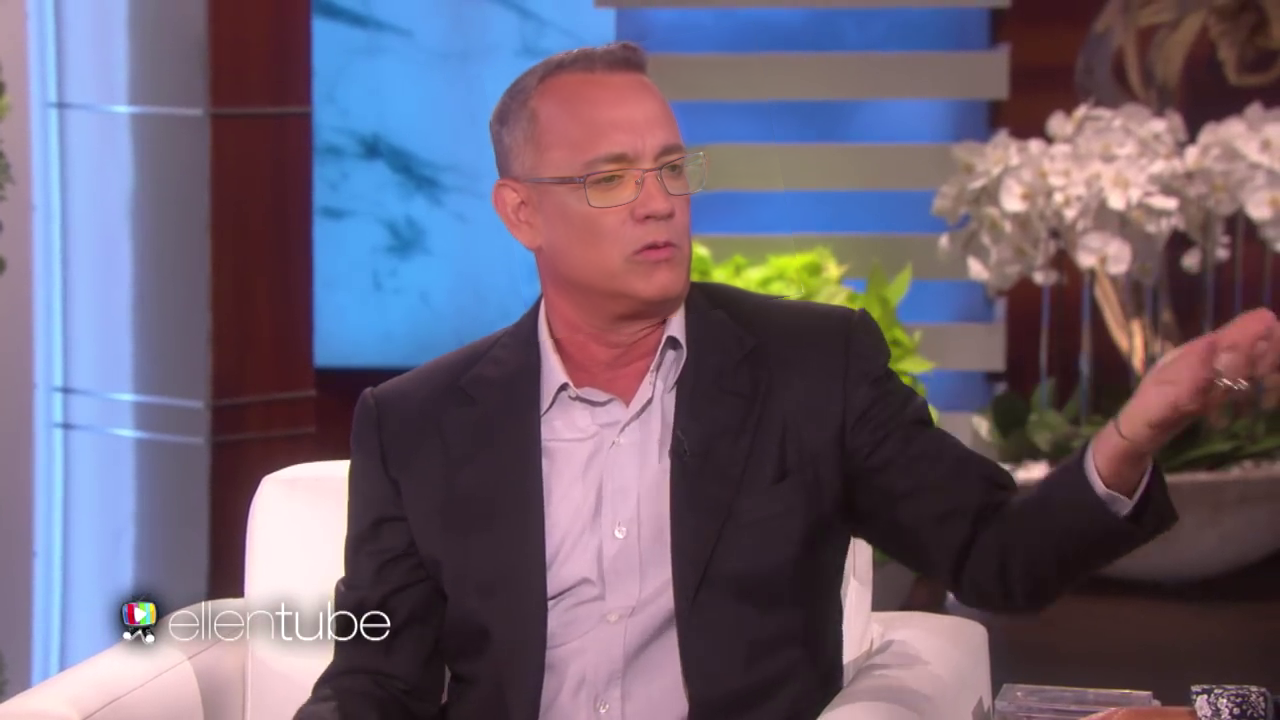}}\\

\vspace{-20mm}

\caption{
\textbf{Temporally consistent video semantic editing.} 
We present a method for editing the semantic attributes of a video using a pre-trained StyleGAN model.
Here we showcase free-form text based editing from SytleCLIP~\cite{Patashnik_2021_ICCV} to make the person appear ``angry" (2nd row) or wear ``eyeglasses" (3rd row).
}
\label{fig:teaser}
\end{center}
\vspace{-15mm}
\end{figure}

\begin{abstract}
Generative adversarial networks (GANs) have demonstrated impressive image generation quality and semantic editing capability of real images, e.g., changing object classes, modifying attributes, or transferring styles. 
However, applying these GAN-based editing to a video independently for each frame inevitably results in temporal flickering artifacts.
We present a simple yet effective method to facilitate temporally coherent video editing. 
Our core idea is to minimize the temporal photometric inconsistency by optimizing both the latent code and the pre-trained generator. 
We evaluate the quality of our editing on different domains and GAN inversion techniques and show favorable results against the baselines.
\end{abstract}

\section{Introduction}
\label{sec:intro}
Generative adversarial models (GANs)~\cite{goodfellow2014generative} have shown remarkable ability to generate photorealistic images in various domains such as faces and scenes~\cite{brock2018large,karras2019style,karras2020analyzing}.
GANs take a latent code (usually sampled from a Gaussian distribution) as input and produce an image as the output. 
\emph{GAN inversion} techniques allow us to project a \emph{real image} onto the latent space of a pretrained GAN and retrieve its corresponding latent code. 
The pretrained GAN generator can then reconstruct that image using the estimated latent code.
Modifying this estimated latent code opens up exciting new opportunities to perform a wide range of high-level editing tasks that are traditionally challenging, e.g., changing semantic object classes, modifying high-level attributes of the object/scene, or even applying 3D geometric transformations. 
We refer to the modification of the latent code with a semantic change in the image as \emph{semantic editing}, e.g., changing the semantic attributes of an object.

\begin{wrapfigure}{r}{0.4\textwidth}
\vspace{-5mm}


\frame{\includegraphics[trim=0 0 0 0, clip,width=.12\textwidth]{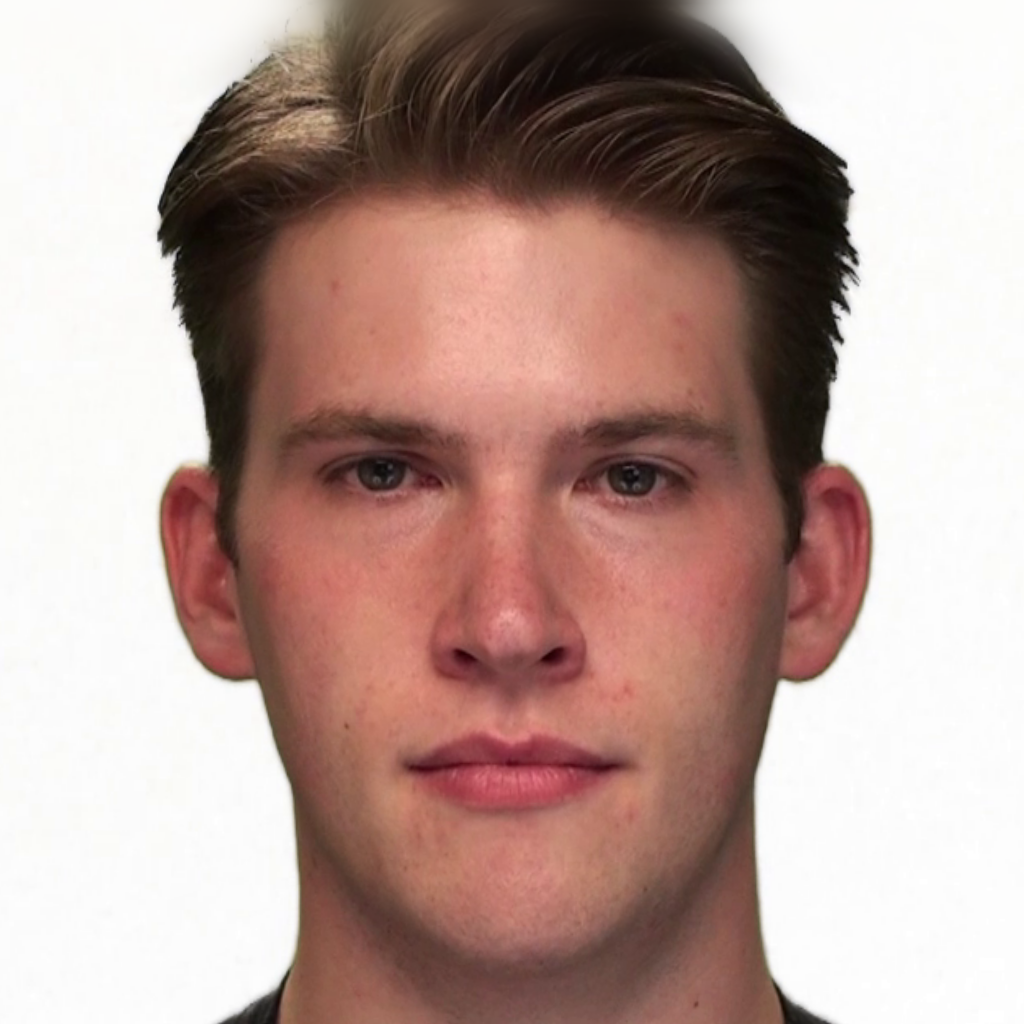}}\hfill
\frame{\includegraphics[trim=0 0 0 0, clip,width=.12\textwidth]{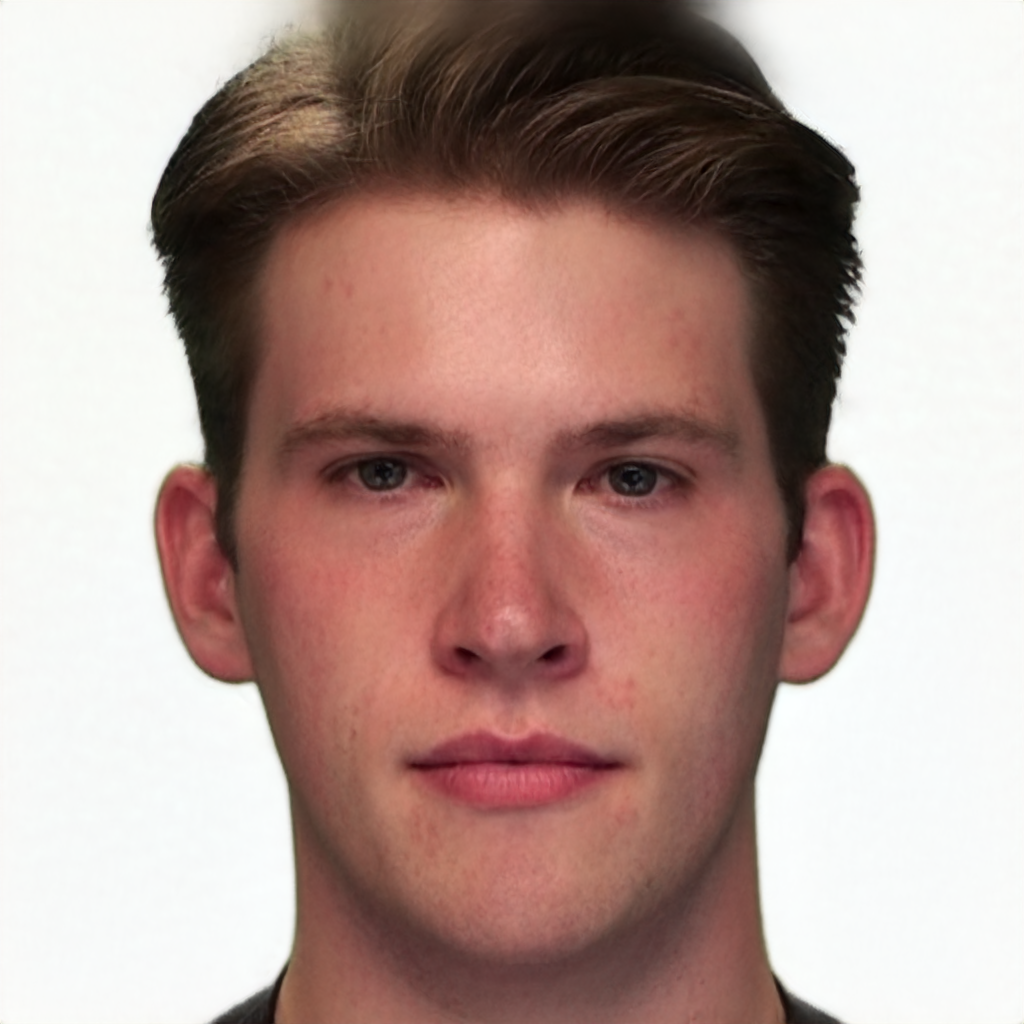}}\hfill
\frame{\includegraphics[trim=0 0 0 0, clip,width=.12\textwidth]{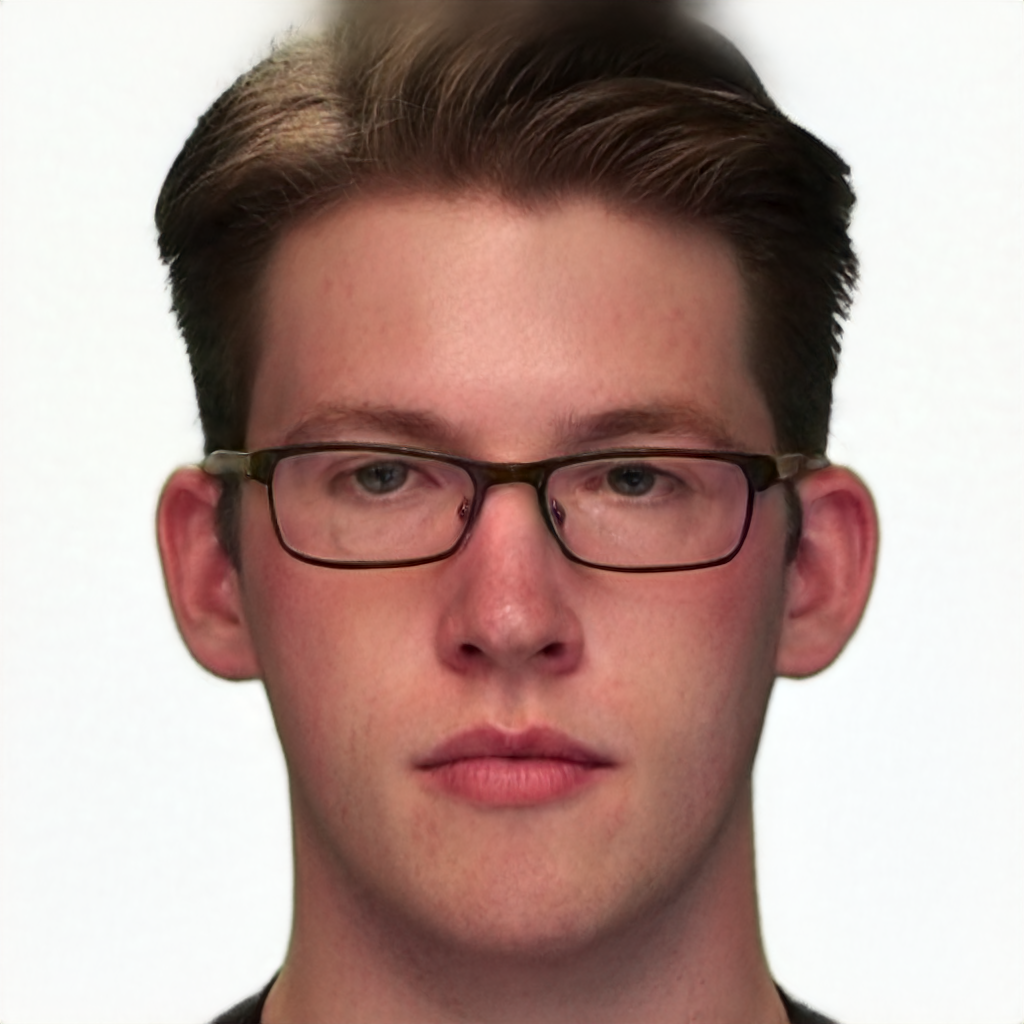}}\\

\vspace{-4mm}

\frame{\includegraphics[trim=0 0 0 0, clip,width=.12\textwidth]{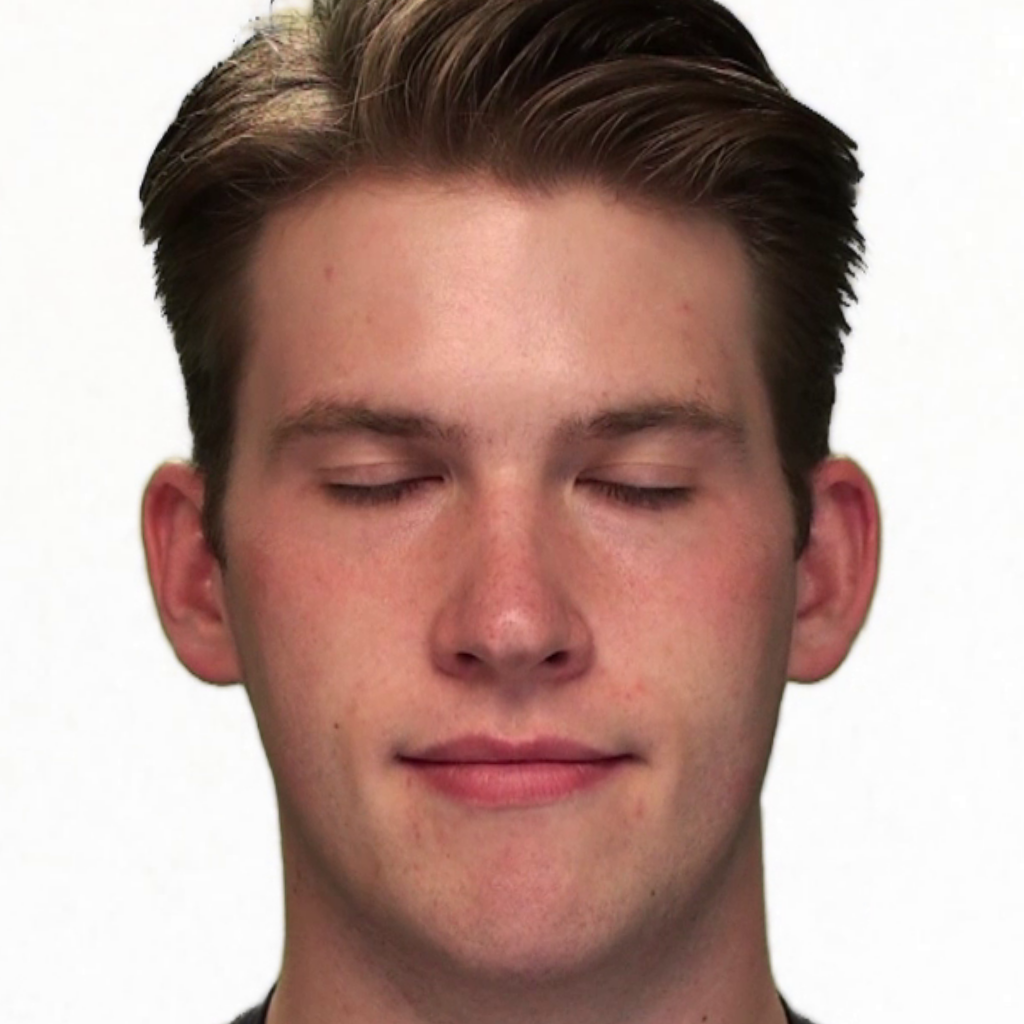}}\hfill
\frame{\includegraphics[trim=0 0 0 0, clip,width=.12\textwidth]{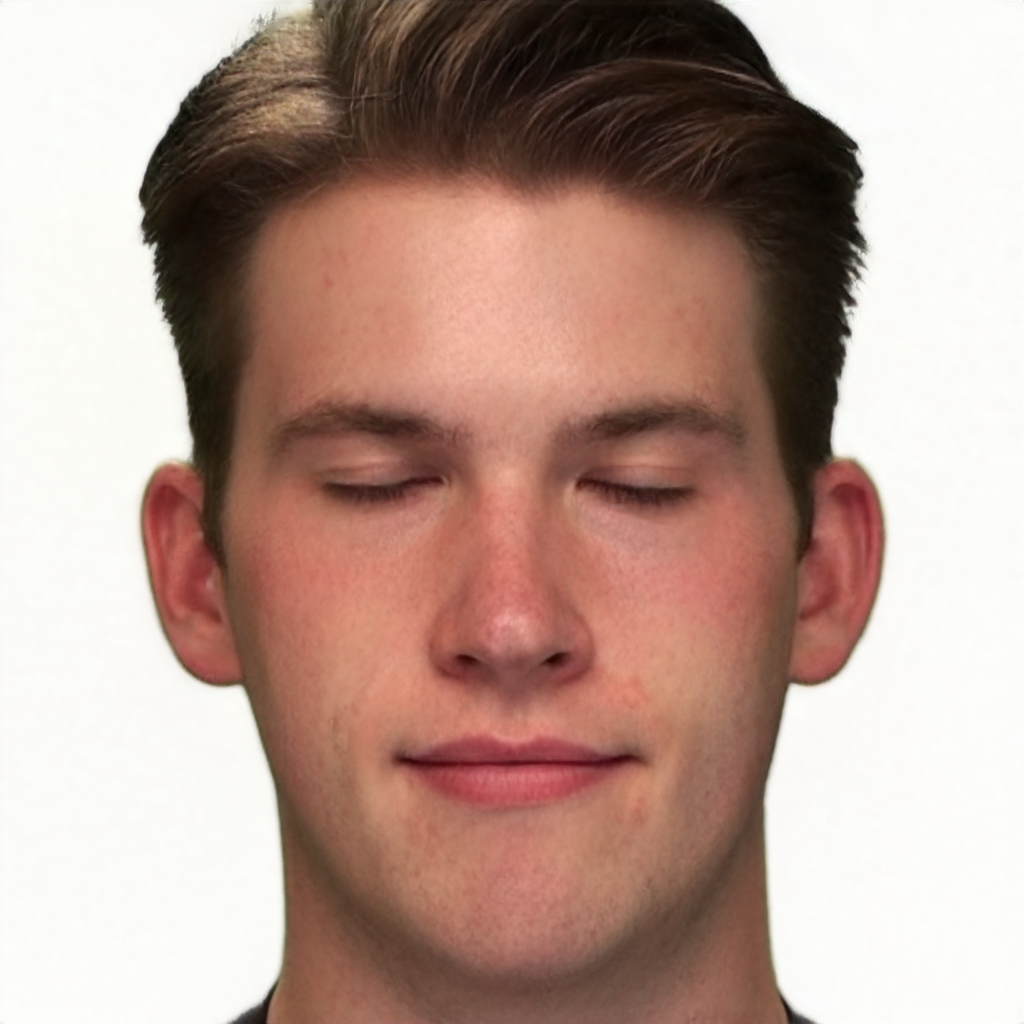}}\hfill
\frame{\includegraphics[trim=0 0 0 0, clip,width=.12\textwidth]{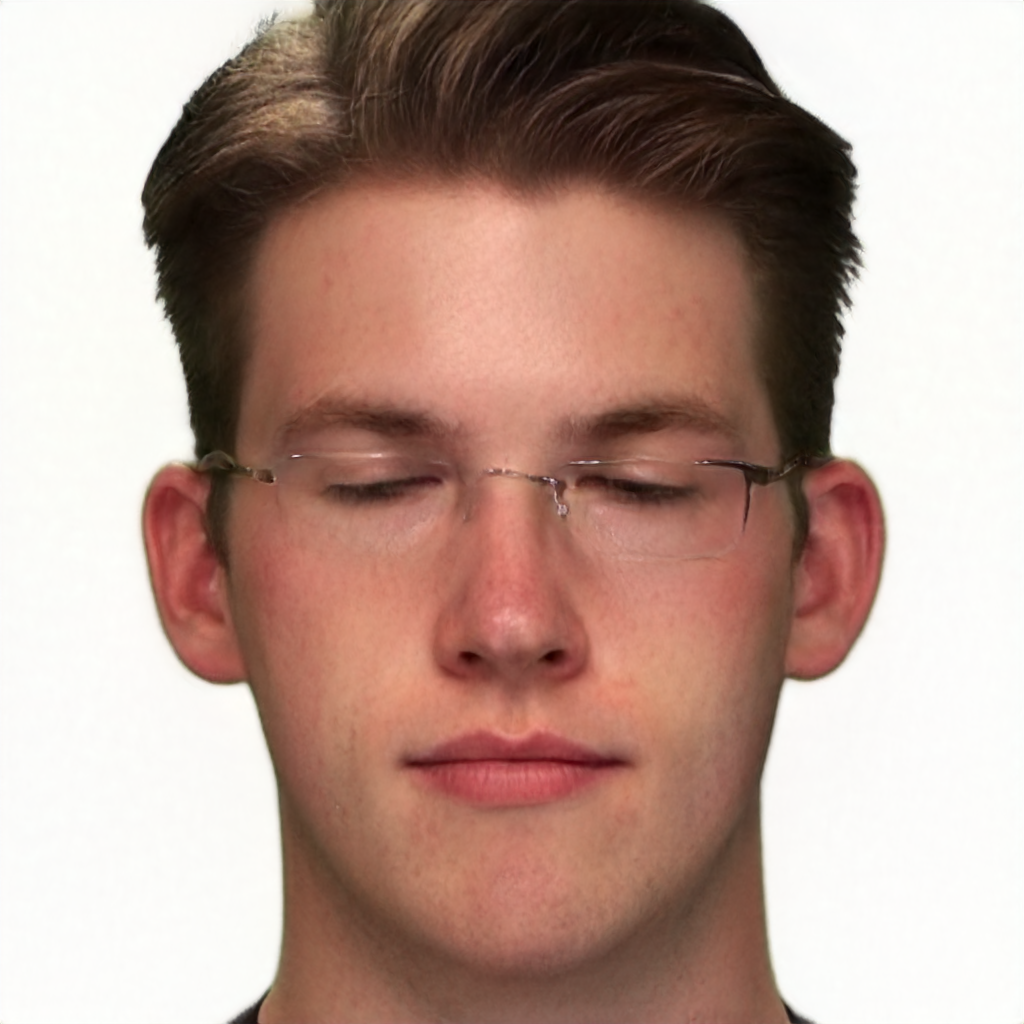}}\\

\vspace{-5mm}

\mpage{0.3}{\small{Input}}\hfill
\mpage{0.3}{\small{Inverted}}\hfill
\mpage{0.3}{\small{Edited}}
\vspace{-5mm}

\caption{
\textbf{Issues with per-frame editing.}
While current methods achieve faithful inversion and photorealistic editing, the 
results are inconsistent across frames (\emph{eyeglasses}) and may fail to preserve details of the input video (\emph{lips}).
%
}
\label{fig:motivation}
\end{wrapfigure}
\topic{Semantic editing in images.} 
A recent line of research work~\cite{zhu2016generative,abdal2019image2stylegan,abdal2020image2stylegan++,huh2020ganprojection,zhu2020indomain,alaluf2021restyle,roich2021pivotal} has shown promising results in reconstructing an input image by either optimizing the latent code (or latent variables) or directly predicting the latent code via an image encoder.
These GAN inversion techniques enable interesting semantic photo editing applications. 
For image-level editing applications, several approaches~\cite{haerkoenen2020ganspace,shen2020interfacegan,shen2021closedform} find specific semantic directions in the latent space, e.g., changing poses, colors, or age, while others~\cite{gal2021stylegannada} aim to change the global style, e.g., photo $\rightarrow$ sketch, photo $\rightarrow$ cartoon. 
We denote them as \emph{In-domain} and \emph{Out-of-domain} editing, respectively.
With these \emph{image} GAN inversion-based semantic editing approaches, 
how can we extend them to \emph{videos}?

\topic{Per-frame editing.} 
One straightforward approach is to apply existing GAN inversion techniques~\cite{huh2020ganprojection,zhu2020indomain,alaluf2021restyle,roich2021pivotal} for each frame in a video \emph{independently}.
\figref{motivation} shows an example of applying a StyleCLIP mapper~\cite{Patashnik_2021_ICCV} on two frames.
The input and the independently reconstructed frames look plausible when viewed individually. 
However, the two edited frames exhibit inconsistency (e.g., the frame of the eyeglasses).
Very recently, Yao et al.~\cite{yao2021latent} learns to predict per-frame semantic editing directions for editing face videos. 
However, the edited videos suffer from apparent temporal flickering and fail to preserve facial identity. 



\topic{Our work.} 
In this paper, we present a method for \emph{temporally consistent} video semantic editing. 
We start from the existing GAN inversion approaches~\cite{alaluf2021restyle,roich2021pivotal} to obtain the latent code for each frame.
We first modify the latent code to achieve the initial per-frame editing results.
However, such a direct editing approach results in temporal inconsistencies in the modified video's appearance or style. 
To deal with this challenge, we propose to compute bi-directional optical flow
estimated from a frame pair sampled from the video.
We can then adjust the latent code and the generator to minimize the photometric loss (along with valid flow vectors).
We present a two-phase optimization strategy. 
In the first phase, we update only the latent codes via an MLP (with generator parameters frozen) to adjust the consistency of the detailed appearance.
In the second phase, we finetune the generator with a local regularization to maintain the editability of the latent space. 
Our two-phase optimization approach helps achieve significantly improved temporal consistency while preserving the edited contents.

\topic{Concurrent work.}
Two concurrent work~\cite{tzaban2022stitch,alaluf2022third} also apply StyleGAN for video editing. 
These methods either use per-frame pivot tuning~\cite{roich2021pivotal} for maintaining the similarity between the edited and input frame~\cite{tzaban2022stitch} or apply latent vector smoothing~\cite{alaluf2022third} with StyleGAN3~\cite{Karras2021}.
Our method differs in 1) the use of explicit temporal consistency optimization and 2) the applicability of performing both in-domain and out-of-domain editing.

\topic{Our contributions}
\begin{itemize}
\item We tackle a task on GAN-based semantic editing in videos. 
We propose a simple yet effective flow-based approach to mitigate the temporal inconsistency of a directly (frame by frame) edited video. 
\item We present a two-phase optimization approach for updating the latent code \emph{and} generator to preserve the video details. 
\item Our method is agnostic and can be applied to different GAN inversion and editing approaches.
\end{itemize}
\vspace{-6mm}

\section{Related Work}
\label{sec:related}
\topic{Generative adversarial networks.}
The quality and resolution of generated images have been achieved rapidly in recent years~\cite{karras2019style,karras2020analyzing,karras2020training,Karras2021,brock2018large}. 
These GAN models can map a random latent code (a noise vector) to a photorealistic image. 
Many recent efforts have been devoted to improving the generator architectures~\cite{karras2018progressive,karras2019style,karras2020analyzing,Karras2021}, 
training strategies~\cite{brock2018large}, loss function designs~\cite{mao2017least,gulrajani2017improved}, and regularization~\cite{miyato2018spectral}.
Our work builds upon existing pretrained StyleGAN models as they demonstrate disentangled latent space for editing.
Instead of \emph{generating synthetic images}, our goal is to \emph{edit real videos}.

\topic{GAN inversion.}
GAN inversion~\cite{zhu2016generative,xia2021gan} allows us to reconstruct real images by projecting them onto a pretrained GAN's latent space. 
These techniques facilitate interesting photo editing applications. 
They can be split into
encoder-based~\cite{luo2017learning,tewari2020stylerig,Nitzan2020FaceID,viazovetskyi2020stylegan2,richardson2021encoding,alaluf2021restyle,tov2021designing,chai2021latent,richardson2021encoding,tov2021designing}, 
optimization-based ~\cite{raj2019gan,abdal2019image2stylegan,abdal2020image2stylegan++,huh2020ganprojection,tewari2020pie,gu2020image,collins2020editing,daras2020your}, 
and hybrid methods~\cite{zhu2020indomain,bau2020semantic,roich2021pivotal}. 
Our method is \emph{agnostic} to different GAN inversion approaches for initializing the latent code. 
For example, our experiments explore using PTI inversion~\cite{roich2021pivotal} for in-domain editing and Restyle encoder~\cite{alaluf2021restyle} for out-of-domain editing.

\topic{Semantic image editing in latent space.}
Semantic image manipulation and editing allow us to change the content and style of an image. It can be grouped into In-Domain editing and Out-of-Domain editing.
Targeting at finding semantic directions in the latent space of a pretrained generator, 
in-domain editing~\cite{shen2020interfacegan,haerkoenen2020ganspace,shen2021closedform,Wu_2021_CVPR,yuksel2021latentclr,li2021dystyle,Patashnik_2021_ICCV,abdal2021styleflow,alaluf2021only,wu2021coarse,afifi2021histogan,saha2021loho} 
manipulates the attributes of the object, but keeps the same style. 
Out-of-domain~\cite{kwong2021unsupervised,jang2021stylecarigan,gal2021stylegannada}, however, aims to change the style of the image. 
These techniques usually perform well on a single image but fail to maintain temporal consistency if applied to a video. 

\topic{Semantic video editing.}
Recent and concurrent work~\cite{yao2021latent,tzaban2022stitch,alaluf2022third} explore \emph{video editing} with a pre-trained StyleGAN.
The methods in~\cite{yao2021latent,tzaban2022stitch} apply per-frame editing and show coherent editing without using any temporal information. 
However, these methods support only in-domain editing.
For \emph{localized editing} (e.g., adding eyeglasses), we find that the method in~\cite{yao2021latent} produces inconsistency and fails to preserve identity.
The work~\cite{alaluf2022third} applies temporal smoothing on the \emph{inverted latent vectors} in StyleGAN3~\cite{Karras2021}. 
Our approach, in contrast, directly minimizes the temporal photometric inconsistency at the \emph{synthesized frames}.



\topic{Video editing and temporal consistency.}
Temporal consistency is one critical criterion in video editing. 
Existing methods achieve temporal consistency often by enforcing the output videos to satisfy the constraints imposed by 2D optical flow~\cite{chen2017coherent,huang2016temporally}.
Alternatively, several methods first estimate an unwarped 2D texture map (either explicitly~\cite{rav2008unwrap} or implicitly~\cite{kasten2021layered}) and then perform editing, e.g., adding a pattern or changing the style of the 2D unwarped textures.
The editing can then be propagated to the original video via the estimated UV mapping.
Several \emph{blind} methods enhance the temporal consistency as a \emph{post-processing} step~\cite{bonneel2015blind,lai2018learning,lei2020dvp}. 
However, they typically have difficulty handling videos with significant appearance changes.
Our work shares similar ideas with these methods to enforce temporal consistency, using the optical flow fields estimated from the initial edited video.
Instead of directly optimizing the \emph{pixel values}, our core idea is to leverage the pretrained generator, and update the latent code and generator to achieve temporal consistent \emph{and} photorealistic results. 

\vspace{-3mm}

\section{Method}
\label{sec:method}
\begin{figure*}[t]
\centering
\includegraphics[trim=0 0 0 0, clip,width=\textwidth]{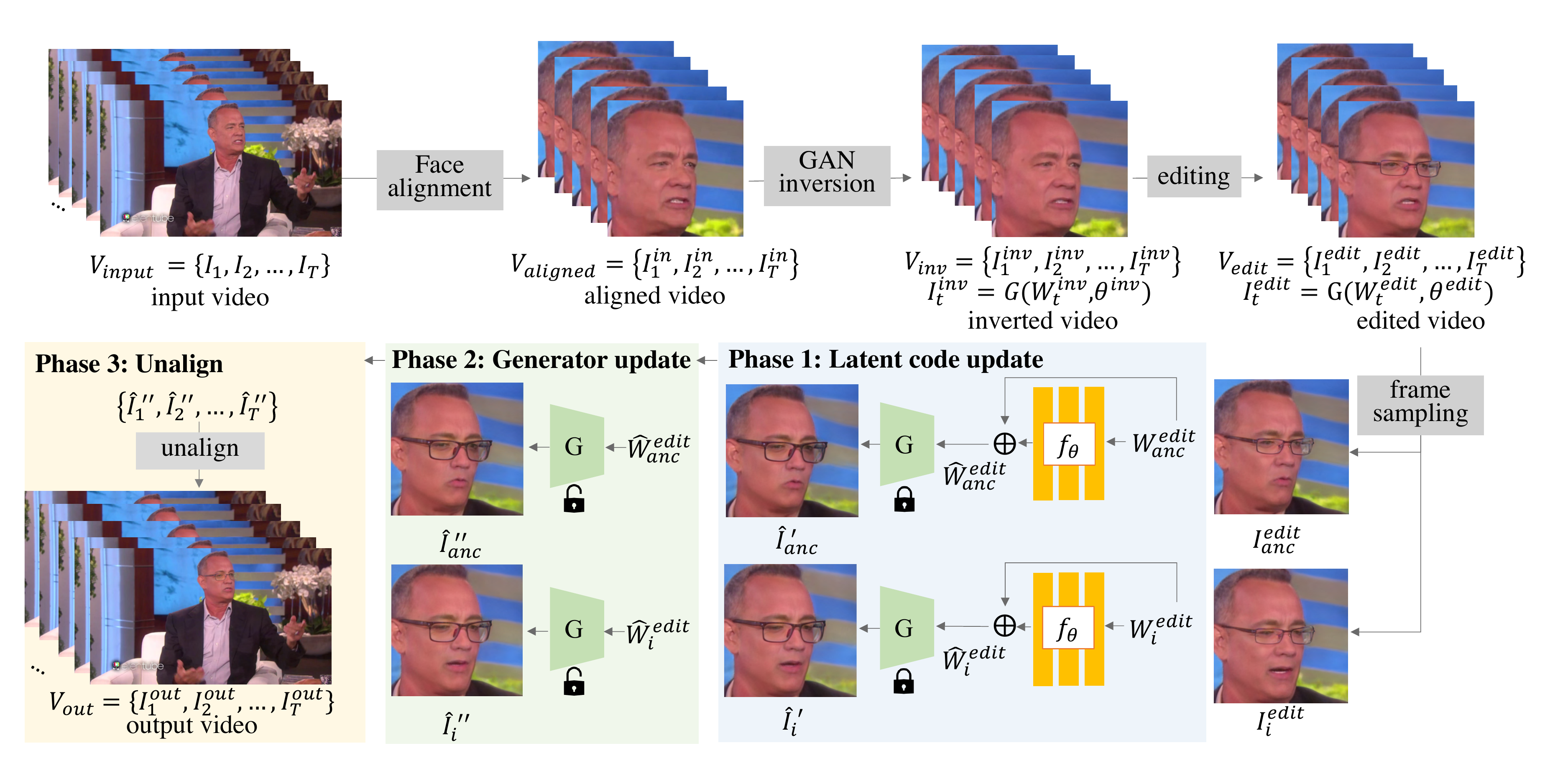}
\vspace{-8mm}
\caption{\textbf{Video editing with flow-based temporal consistency.} 
Given an input video of $T$ frames $V_{input}$, we first spatially align the video frames using an off-the-shelf face landmark detector.
We then use existing GAN inversion techniques~\cite{roich2021pivotal,alaluf2021restyle} to obtain the inverted frames $\{I^{inv}_1, I^{inv}_2, \cdots, I^{inv}_T\}$ and their corresponding latent code in the $\mathcal{W^{+}}$-space of StyleGAN $\{W^{inv}_1, W^{inv}_2, \cdots, W^{inv}_T\}$.
We independently perform semantic editing on these inverted frames to obtain $\{I^{edit}_{1}, I^{edit}_{2}, \cdots, I^{edit}_{T}\}$ and their corresponding latent code $\{W^{edit}_1, W^{edit}_2, \cdots, W^{edit}_T\}$.
To achieve temporal consistency, we choose an anchor frame $I^{edit}_{anc}$ as the reference frame, and each time sample another frame $I^{edit}_{i}$ from the edited video.
To generate a temporally consistent edited video, we first refine the latent codes of the directly edited video $W^{edit}_{anc}$ and $\{W^{edit}_{i}\}_{i \neq anc}$ to $\hat{W}^{edit}_{anc}$ and $\{\hat{W}^{edit}_{i}\}_{i \neq anc}$ by optimizing an MLP $f_\theta$ (phase 1). 
These refined latent codes result in the temporally consistent frames $\hat{I}^{'}_{anc}$ and $\hat{I}^{'}_{i}$.
To further improve the temporal consistency, we keep the refined latent codes $\hat{W}^{edit}_{anc}$ and $\hat{W}^{edit}_{i}$ and only update the generator parameters (phase 2). 
This will generate $\hat{I}^{''}_{anc}$ and $\hat{I}^{''}_{i}$ with improved temporal consistency.
After our two phase optimization, we finally unalign the frames to generate our final edited video $V_{out}$ (phase 3).
}
\vspace{-8mm}
\label{fig:approach}
\end{figure*}
\begin{figure*}[t]
\centering
\includegraphics[trim=0 0 0 0, clip,width=\textwidth]{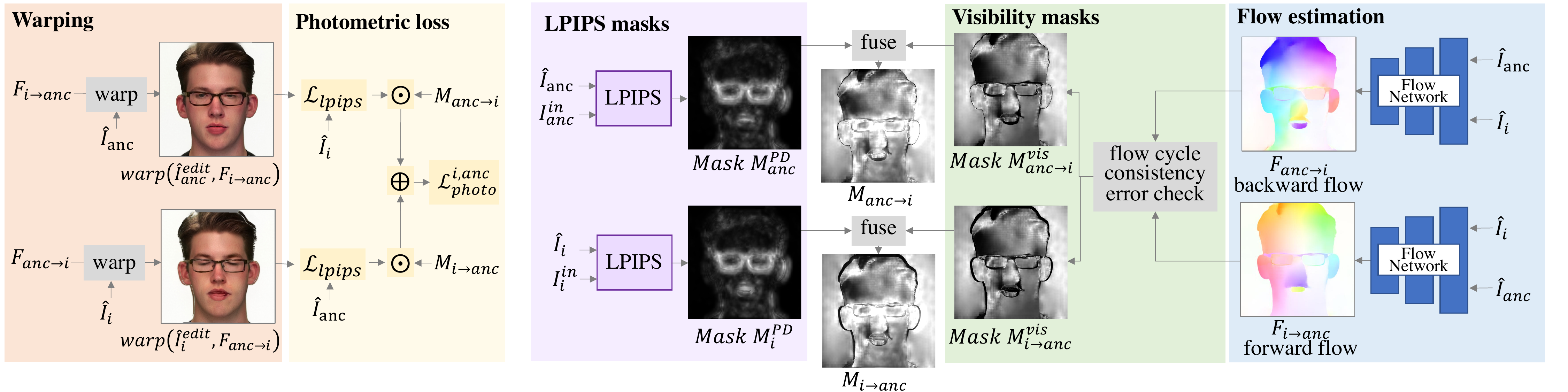}
\mpage{0.34}{\small{(a) $\mathcal{L}_{photo}$ computation}}\hfill
\mpage{0.64}{\small{(b) Flow and mask computation}}\\
\vspace{-2mm}
\caption{\textbf{Photometric loss for temporal consistency.} 
Given a frame pair $\hat{I}_{i}$ and $\hat{I}_{anc}$ (either from phase 1 or phase 2), we compute the forward and backward flows $F_{i \rightarrow anc}$ and $F_{anc \rightarrow i}$ using RAFT~\cite{teed2020raft}. 
We then use these two flow fields to compute the visibility masks by performing a forward-backward and backward-forward flow consistency error check.
For in-domain editing, we also use LPIPS to obtain a semantic mask that highlights the difference between the aligned input frames $I^{in}_{i}$ and $I^{in}_{anc}$ and our edited frames $\hat{I}_{i}$ and $\hat{I}_{anc}$.
We then fuse both the LPIPS semantic masks and the visibility masks to get our final masks $M_{anc \rightarrow i}$ and $M_{i \rightarrow anc}$.
To compute the photometric loss (Eqn.~\ref{eq:photo_loss}), we use the flows to warp the directly edited frames and utilize the fuzed masks as shown in (a).
}
\vspace{-7mm}
\label{fig:photoloss}
\end{figure*}

\subsection{Overview}
\topic{GAN Inversion.}
Given an input video $V_{input} = \{I_1, \cdots, I_T\}$ of $T$ frames, our goal is to semantically edit all the video frames while preserving the temporal coherence of the edited video. 
To edit the input video $V_{input}$, we first align its frames by using a facial alignment method~\cite{guo2020towards}. 
Then we use existing GAN inversion techniques (e.g., ~\cite{roich2021pivotal,alaluf2021restyle}) to invert the frames back to the latent code such that the inverted frame $I^{inv}_t = G(W^{inv}_t; \theta^{inv})$ is similar to the input frame: $I^{inv}_t \approx I_t$.
With the inverted frames, we can edit the inverted video $V_{inv} = \{I^{inv}_1, I^{inv}_2, \cdots, I^{inv}_T\}$ by \emph{independently} editing its frames $I^{inv}_t$.
We denote this frame-by-frame editing approach as ``direct editing''.

\topic{In-domain and out-of-domain GAN-based editing.}
Commonly used \emph{image-based} editing techniques via a GAN include (1) in-domain and (2) out-of-domain editing. 
We refer to an \emph{in-domain editing}~\cite{shen2020interfacegan,haerkoenen2020ganspace,shen2021closedform,Wu_2021_CVPR} as the editing that only manipulates the latent code, given a \emph{fixed} pretrained generator. 
That is, the generator parameters $\theta^{inv}$ remain frozen ($\theta^{inv} = \theta^{edit}$), and only the latent code  $W^{edit}_t$ is updated.
The in-domain editing usually changes semantic attributes such as {color, age, or facial expressions}. 
On the other hand, out-of-domain editing may involve updating the pretrained generator to produce an entirely new style (as shown in ~\cite{gal2021stylegannada}). 
Here, the latent code remains the same $W^{edit}_t=W^{inv}_t$ and only the generator $\theta^{edit}$ changes.

\begin{figure}[t]
\frame{\includegraphics[trim=0 0 0 0, clip,width=.2\textwidth]{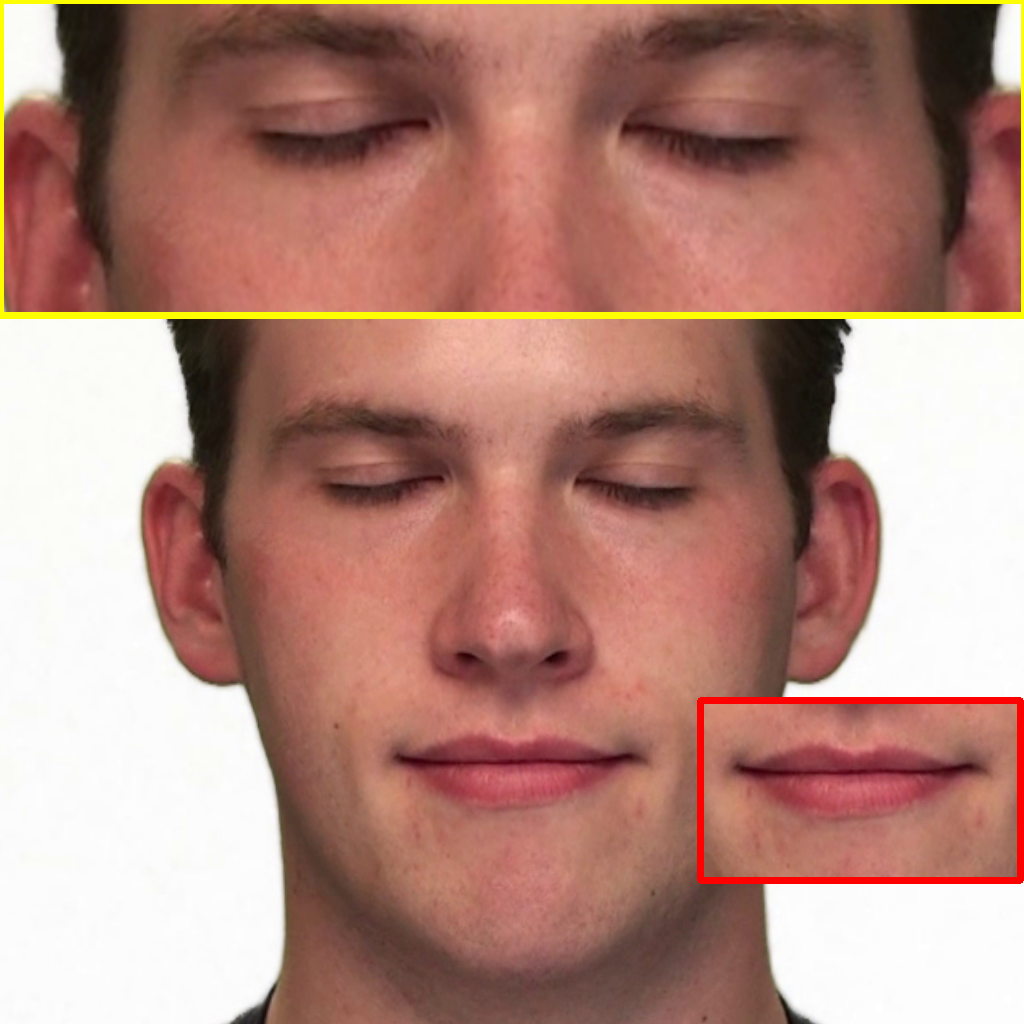}}\hfill 
\frame{\includegraphics[trim=0 0 0 0, clip,width=.2\textwidth]{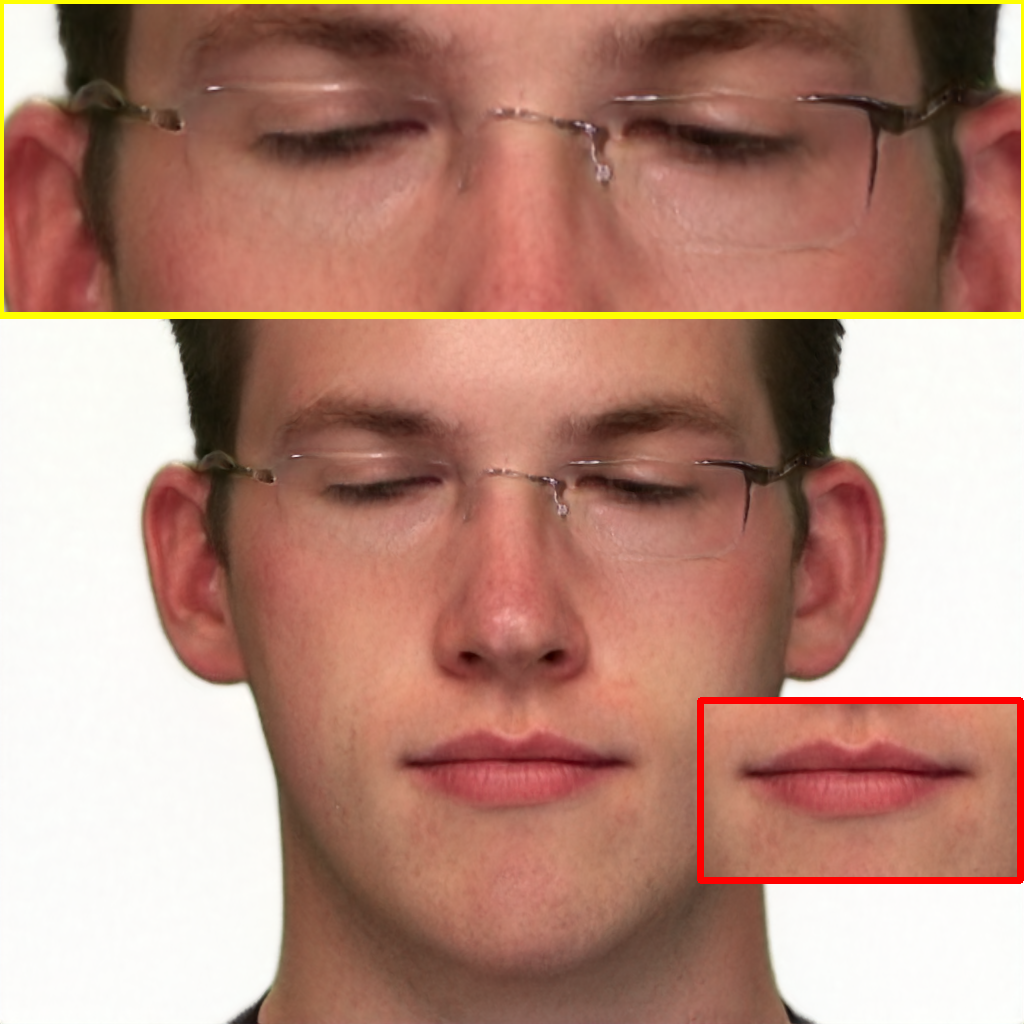}}\hfill 
\frame{\includegraphics[trim=0 0 0 0, clip,width=.2\textwidth]{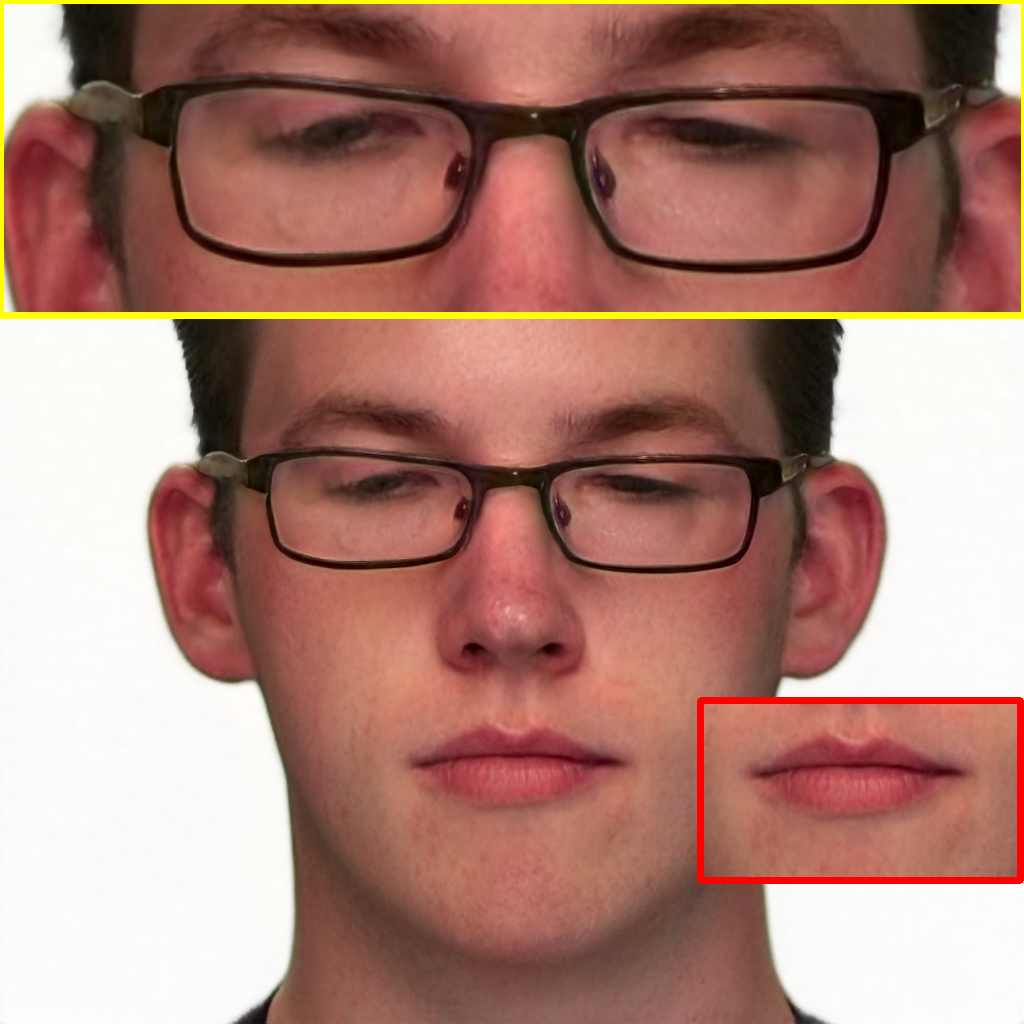}}\hfill
\frame{\includegraphics[trim=0 0 0 0, clip,width=.2\textwidth]{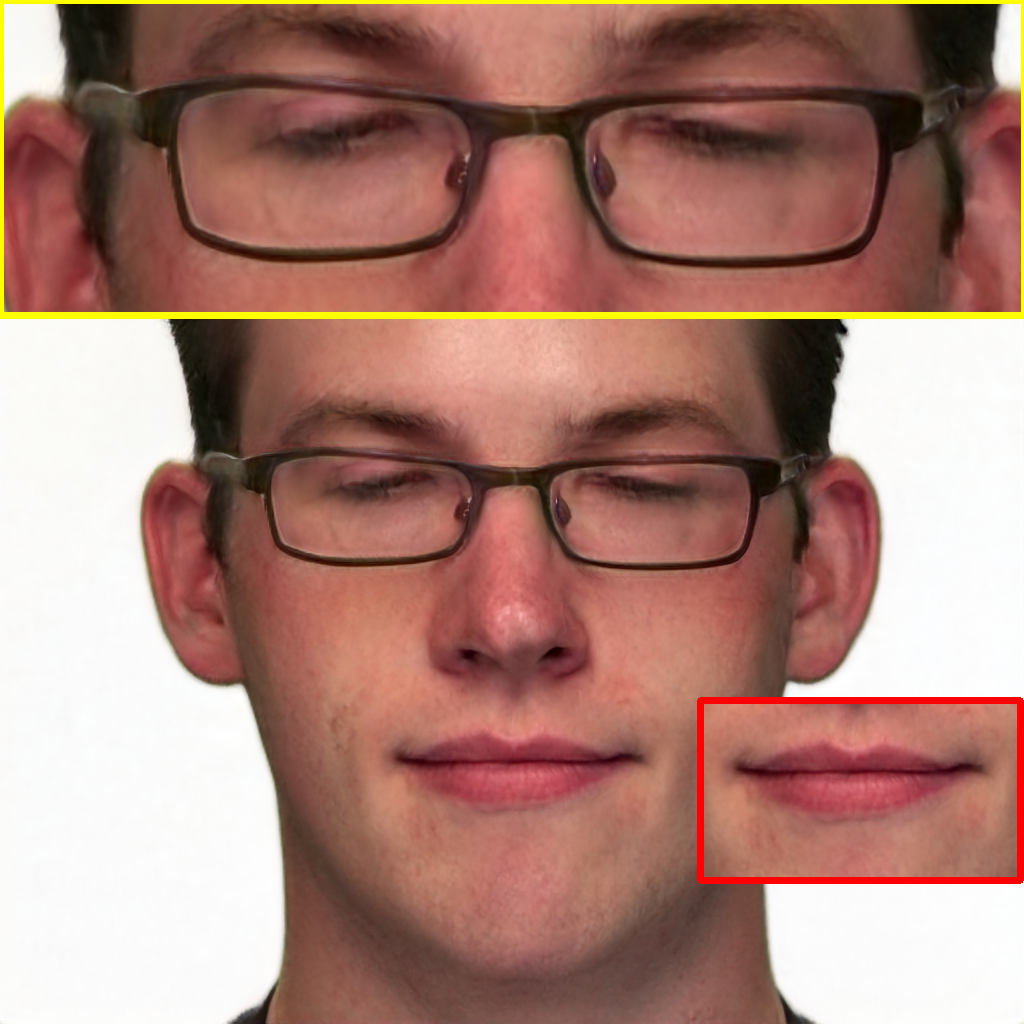}}\\

\vspace{-5mm}
\mpage{0.2}{\footnotesize{Input}}\hfill
\mpage{0.2}{\footnotesize{Direct Editing}}\hfill
\mpage{0.2}{\footnotesize{$W$ updated}}\hfill
\mpage{0.2}{\footnotesize{$W, G$ updated}}

\vspace{-3mm}
\caption{
\textbf{Motivation for two-phase optimization.}
Updating latent code $W$ brings in the eyeglasses, and tuning $G$ with the perceptual difference mask recovers the expression in the input.
}
\vspace{-1mm}
\label{fig:w_G_comp}
\end{figure}

\topic{Direct editing on a video.}
When applying both types of editing techniques to a video independently for each frame, we obtain an edited video 
$V_{edit} = \{I^{edit}_{1}, I^{edit}_{2}, \cdots, I^{edit}_{T}\}$. 
For each directly edited frame $I^{edit}_{t}$, there is a corresponding latent code $W^{edit}_t$ such that $I^{edit}_t = G(W^{edit}_t; \theta^{edit})$. 
Due to the per-frame, independent process, the edited video $V_{edit}$ often suffers from temporal inconsistency.
Moreover, due to the poor disentanglement of this per-frame editing, not only will the edited attributes differ among frames, but other existing facial attributes also change (see the change in mouth in Fig.~\ref{fig:w_G_comp}).
Our goal is to ensure that the edited attributes remain temporally consistent while preserving the other details from the input video. 


\topic{Overview of our approach.}
To achieve this goal, we propose a two-phase optimization approach: phase 1 updates the  \emph{latent code} via an MLP and phase 2 updates the \emph{generator}.
In both phases, we optimize the temporal photometric loss across frames.
With the finetuned latent code and generator, we unalign the edited frames to produce an edited video.
Figure~\ref{fig:approach} outlines our workflow. 
Below, we describe the details and the losses of our approach. 
\vspace{-3mm}
\subsection{Flow-based temporal consistency}\label{sec:method_flow_based}
We present a flow-based approach to explicitly encourage temporal consistency in the edited video $V_{edit}$. 

\topic{Frame sampling.}
As we cannot fit an entire video into the GPU memory, we choose to perform our optimization from a \emph{pair of frames} at a time.
We choose to use an anchor frame $I^{edit}_{anc}$ as one in the pair, which we set as the middle frame of the video.
This is inspired by recent video representation work~\cite{rho2022neural}, where a video is represented by a key frame and a flow network. 
At each iteration, we sample a latent code $W^{edit}_i$, corresponding to the frame $I^{edit}_{i}$ and optimize the pair of frames $\{I^{edit}_{anc}, I^{edit}_{i}\}$.
We perform our optimization in two phases (Section~\ref{sec:two_phase}). 
In phase 1, we generate temporally consistent pairs $\{\hat{I}^{'}_{anc}, \hat{I}^{'}_{i}\}_{i \neq anc}$ as a result. 
In phase 2, we further improve the temporal consistency, recover other affected attributes brought by the per-frame editing due to the poor disentanglement, and generate the pairs $\{\hat{I}^{''}_{anc}, \hat{I}^{''}_{i}\}_{i \neq anc}$. 

\topic{Flow estimation and warping.}
We use RAFT~\cite{teed2020raft} to compute the forward and backward flows $F_{i \rightarrow anc}$ and $F_{anc \rightarrow i}$ of the pair $\{\hat{I}_{anc}, \hat{I}_{i}\}$. 
This pair is either the output of phase 1 $\{\hat{I}^{'}_{anc}, \hat{I}^{'}_{i}\}$ or phase 2 $\{\hat{I}^{''}_{anc}, \hat{I}^{''}_{i}\}$.
We then use these two flows to warp the pair of frames $\{\hat{I}_{anc}, \hat{I}_{i}\}$.

\topic{Visibility masks.}
To highlight the \emph{non-occluded} regions, we compute the visibility masks $M^{vis}_{anc \rightarrow i}$ and $M^{vis}_{i \rightarrow anc}$ $\in [0,1]$. 
This mask shows lower weights for occluded pixels and higher weights for the non-occluded pixels (Figure~\ref{fig:photoloss}).
To compute the visibility masks, we first compute forward-backward and backward-forward flow consistency error maps $\epsilon_{anc \rightarrow i}$ and $\epsilon_{i \rightarrow anc}$ and compute the error map by
$\epsilon_{i \rightarrow anc}(p) = ||p - F_{anc \rightarrow i}(p + F_{anc \rightarrow j}(p))||_2 \,,$
where $p$ is a pixel in the flow field. 
These resultant error maps are mapped to $[0, 1]$ using an exponential function such that $M^{vis}_{anc \rightarrow i} = \exp(-10\epsilon_{anc \rightarrow i})$ and $M^{vis}_{i \rightarrow anc} = \exp(-10\epsilon_{i \rightarrow anc})$. 

\topic{Perceptual difference mask.}
For in-domain editing, because the introduced editing is temporally inconsistent, we observe that the visibility masks do \emph{not} emphasize those edited parts (e.g., eyeglasses).
To highlight those edited parts, we compute the soft semantic perceptual difference masks $M^{PD}_{anc}$ and $M^{PD}_{i}$ between the pair of frames and their corresponding aligned input frames using LPIPS~\cite{zhang2018unreasonable} (Figure~\ref{fig:photoloss}).
Due to the significant appearance differences, we cannot use these semantic perceptual difference masks for out-of-domain editing.

\topic{Fused masks.}
For in-domain editing, we fuse the visibility masks and the semantic perceptual difference masks such that $M_{anc \rightarrow i} = (M^{vis}_{anc \rightarrow i} \oplus M^{PD}_{i})$ and $M_{i \rightarrow anc} = (M^{vis}_{i \rightarrow anc} \oplus M^{PD}_{anc})$. The masks will also be clamped to [0, 1]. This fusion is shown in Figure~\ref{fig:photoloss}.
On the other hand, for out-of-domain editing, $M_{anc \rightarrow i} = M^{vis}_{anc \rightarrow i}$ and $M_{i \rightarrow anc} = M^{vis}_{i \rightarrow anc}$.

\topic{Bi-directional photometric loss.} 
We use the warped frames and the final computed masks to compute the bi-directional photometric loss to achieve a temporally consistent video.
This loss measures the difference between the two frames to calculate the deviation in the non-occluded parts. 
\vspace{-2mm}
\begin{equation}\label{eq:photo_loss}
\begin{split}
    \mathcal{L}_{photo} &= \sum_{\hat{I}_{i}, \hat{I}_{anc} \in P} M_{i \rightarrow anc} \mathcal{L}_{LPIPS}(\hat{I}_{anc}, warp(\hat{I}_{i}, F_{anc \rightarrow i})) \\
    &+ M_{anc \rightarrow i} \mathcal{L}_{LPIPS}(\hat{I}_{i}, warp(\hat{I}_{anc}, F_{i \rightarrow anc})) \,,
\end{split}
\vspace{-4mm}
\end{equation}
where $\hat{I}_{t}$ is either the output of phase 1 $\hat{I}^{'}_{t}$ or phase 2 $\hat{I}^{''}_{t}$. 
Intuitively, this bi-directional photometric loss ensures colors along the valid (forward-backward or backward-forward consistent) vectors across frames are as similar as possible.
\vspace{-3mm}

\subsection{Two-phase optimization strategy} \label{sec:two_phase}
We split our optimization into two phases. 
In the first phase, we refine the latent codes $\{W_t^{edit}\}$ by only optimizing an MLP $f_\theta$. 
While in the second phase, we only update the generator weights $\theta^{edit}$. 

\topic{Motivation.}
We use a two-phase optimization approach for in-domain editing because we observe that only refining the latent codes (phase 1) sometimes introduces undesired changes to \emph{other} facial attributes. 
We show an example in Fig.~\ref{fig:w_G_comp}.
When we only update the latent codes, we achieve temporal consistency of the introduced glasses; however, the mouth expression of the person changes. 
To address this in the case of in-domain editing, we update the generator weights (phase 2) using the perceptual difference mask to enforce the pixels outside the mask to be the same as the input. 
This will maintain the facial expression of the aligned input frame.
The primary source of inconsistency for out-of-domain editing is the global inconsistency (e.g., background). 
Hence, updating the generator (phase 2) introduces this desired global change.

\topic{Phase 1: Latent code update.}
In this phase, we update the latent code $W^{edit}_t$ using a Multi-layer Perceptron
 (MLP) $f_\theta=(w;\theta_{f})$ implicitly. 
 We use the same architecture as StyleCLIP mapper~\cite{Patashnik_2021_ICCV}. 
 We use this MLP to predict a residual for the latent codes and update the parameters of the MLP instead of directly optimizing the latent codes explicitly, such that:
\begin{equation} \label{eq:mlp_update}
    \hat{W}^{edit}_t = W^{edit}_t + \alpha f_\theta(W^{edit}_t;\theta_{f}) \,,
\end{equation}
then for a pair of directly edited frames $\{I^{edit}_{anc}, I^{edit}_{i}\}$, we can get the updated frames $\hat{I}^{'}_{i} = G(\hat{W}^{edit}_i)$, $\hat{I}^{'}_{anc} = G(\hat{W}^{edit}_{anc})$.

Our goal is to minimize:
\begin{equation} \label{eq:first_phase}
    \argmin_{\theta_{f}}\mathcal{L}_{I} = \argmin_{\theta_{f}} \sum_{t \neq anc} \mathcal{L}_{photo} + \lambda_{rf}\mathcal{L}_{rf} + \lambda_{\epsilon}\mathcal{L}_{\epsilon} \,,
\end{equation}
where $\mathcal{L}_{photo}$ is the photometric loss, and 
\begin{equation}
    \mathcal{L}_{rf} = ||f_{\theta}(W^{edit}_t;\theta_{f})||_1 + ||f_{\theta}(W^{edit}_{anc};\theta_{f})||_1    
\end{equation}
is a regularization term to make sure we do not deviate too much from ${W}^{edit}_t$. We set $\lambda_{rf} = 0.1$ for the experiments.
$\mathcal{L}_{\epsilon} = ||\epsilon_{anc \rightarrow i}||_1 + ||\epsilon_{i \rightarrow anc}||_1$ is the norm of error maps, and we set $\lambda_{\epsilon}=10$.

\begin{figure}[t]

\mpage{0.45}{\tiny{x-t slice}}
\mpage{0.2}{\tiny{x-t slice}} 
\mpage{0.3}{\hspace{18mm}\tiny{x-t slice}}\\
\vspace{-2mm}

$\underbracket[1pt][2.0mm]{\includegraphics[trim=0 0 0 0, clip,width=.15\textwidth]{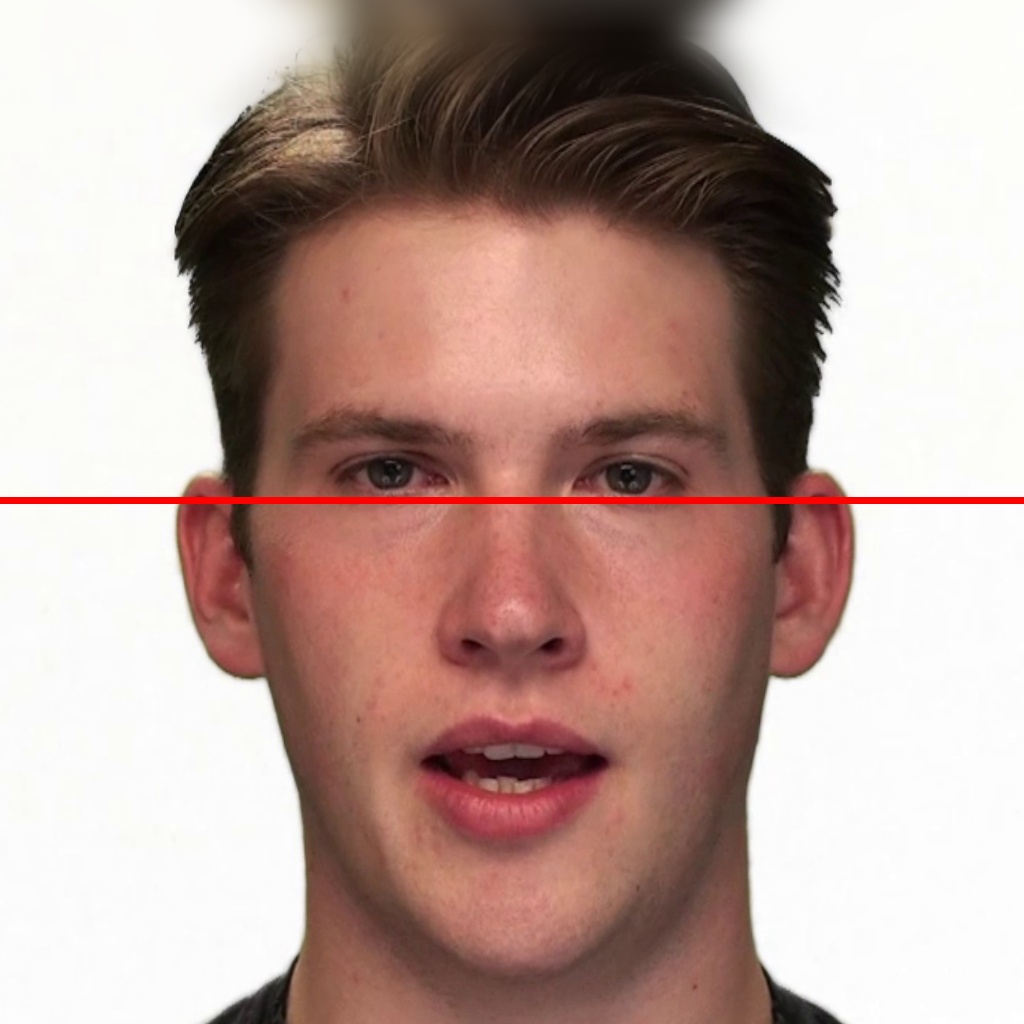}\hfill 
{\includegraphics[trim=0 0 0 0, clip,width=.15\textwidth]{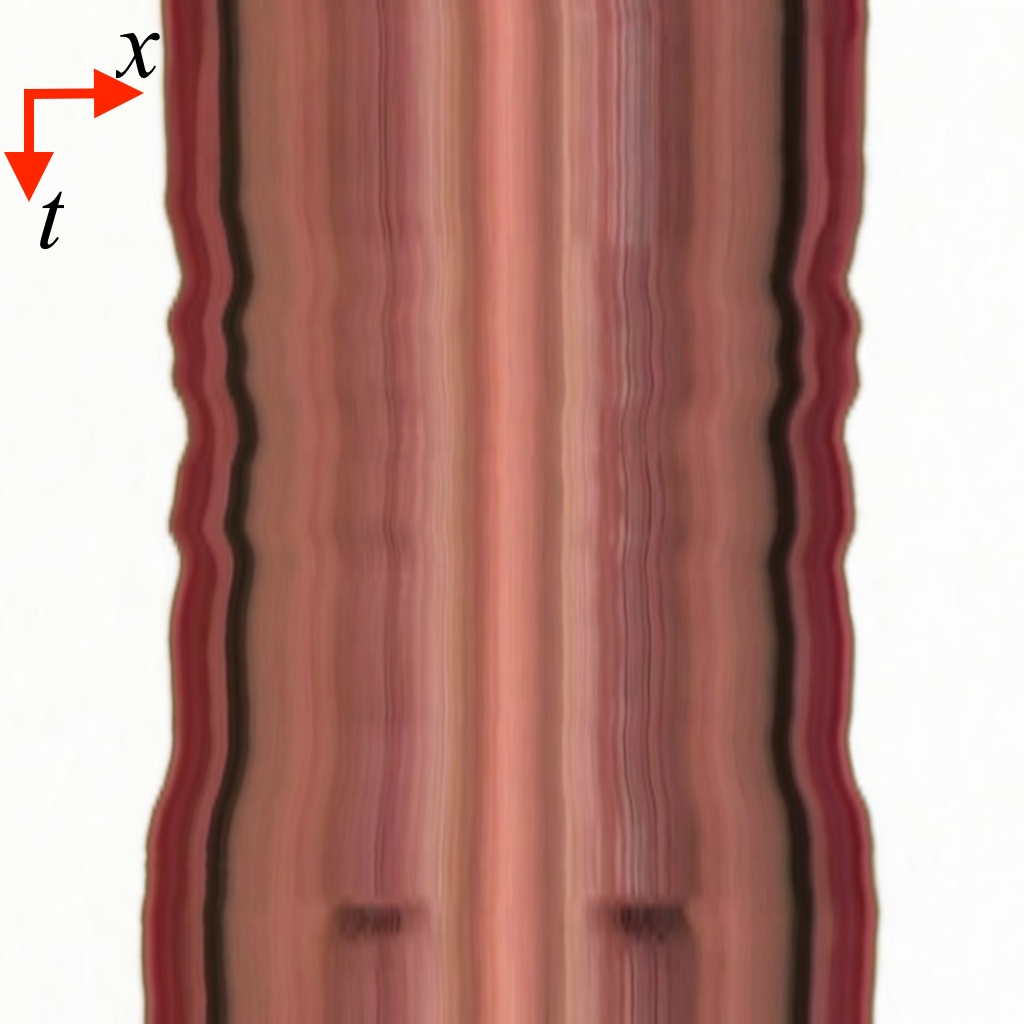}} \\
}_%
    {\substack{\vspace{-3.0mm}\\\colorbox{white}{Input}}}$
\vspace{2mm} \hfill
$\underbracket[1pt][2.0mm]{\includegraphics[trim=0 0 0 0, clip,width=.15\textwidth]{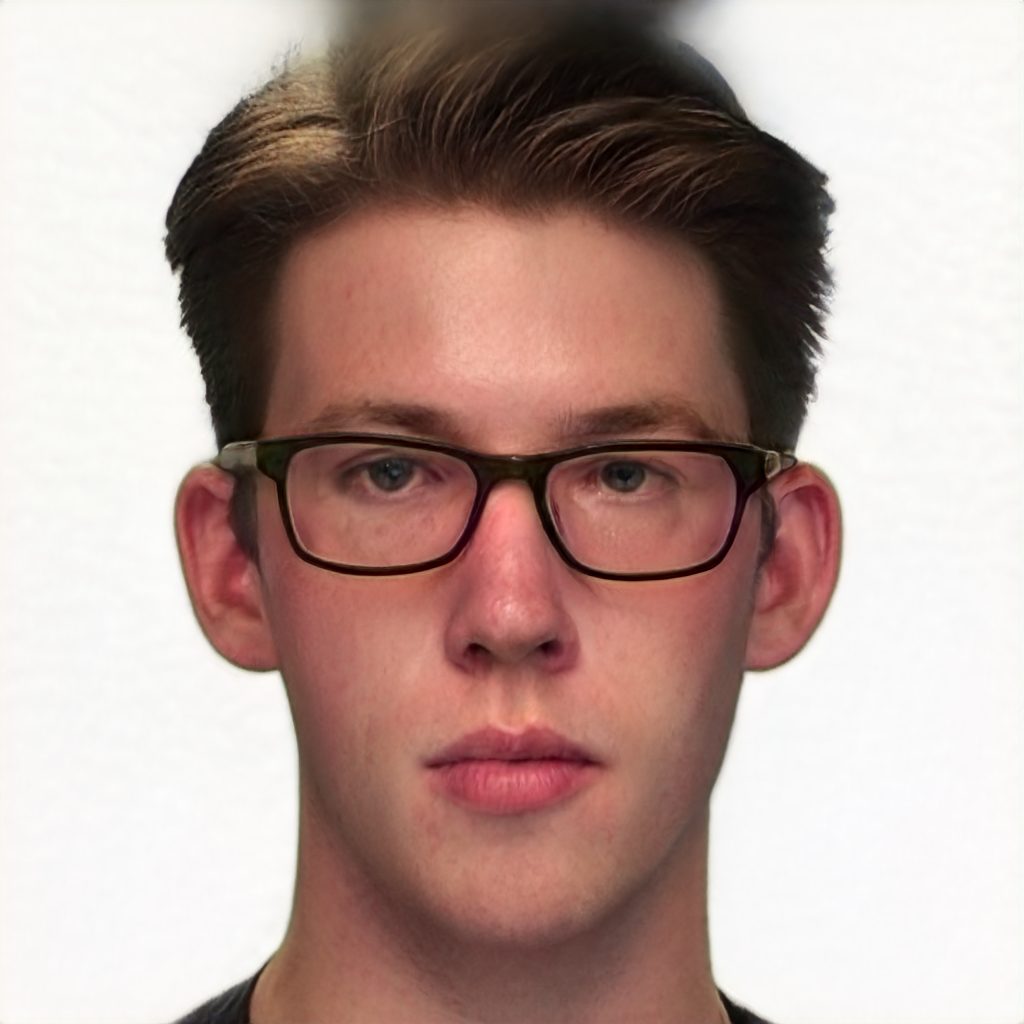}\hfill 
\includegraphics[trim=0 0 0 0, clip,width=.15\textwidth]{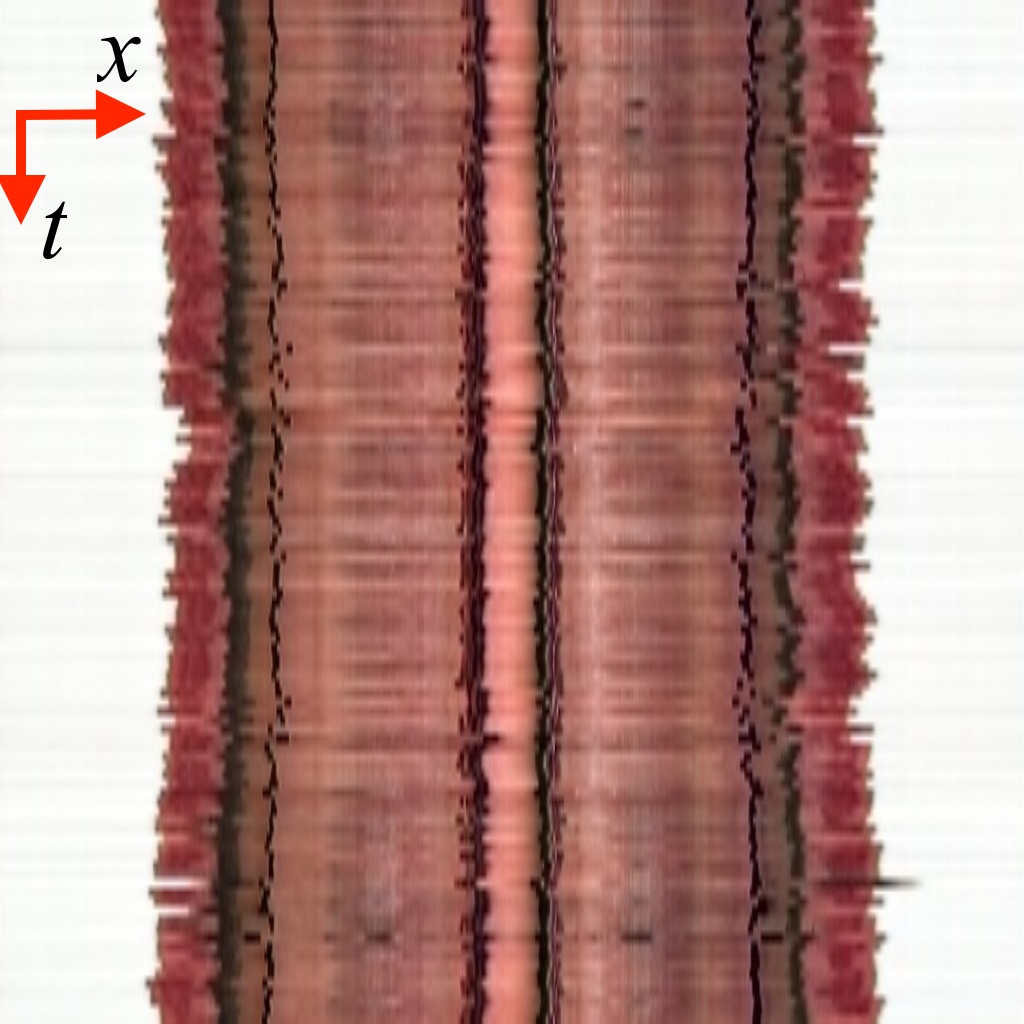}\hfill}_%
    {\substack{\vspace{-3.0mm}\\\colorbox{white}
    {Explicitly update $W$}}}$\vspace{2mm} \hfill
$\underbracket[1pt][2.0mm]{\includegraphics[trim=0 0 0 0, clip,width=.15\textwidth]{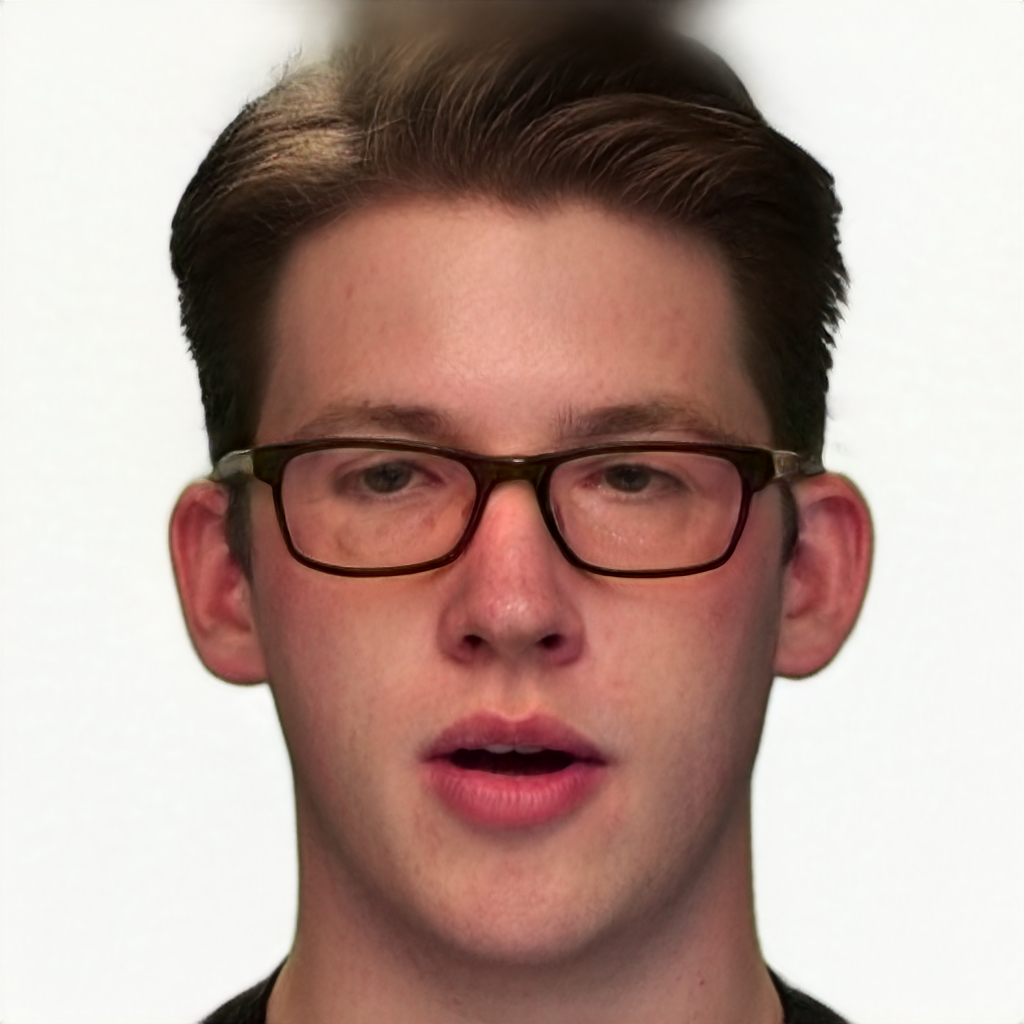}\hfill 
\includegraphics[trim=0 0 0 0, clip,width=.15\textwidth]{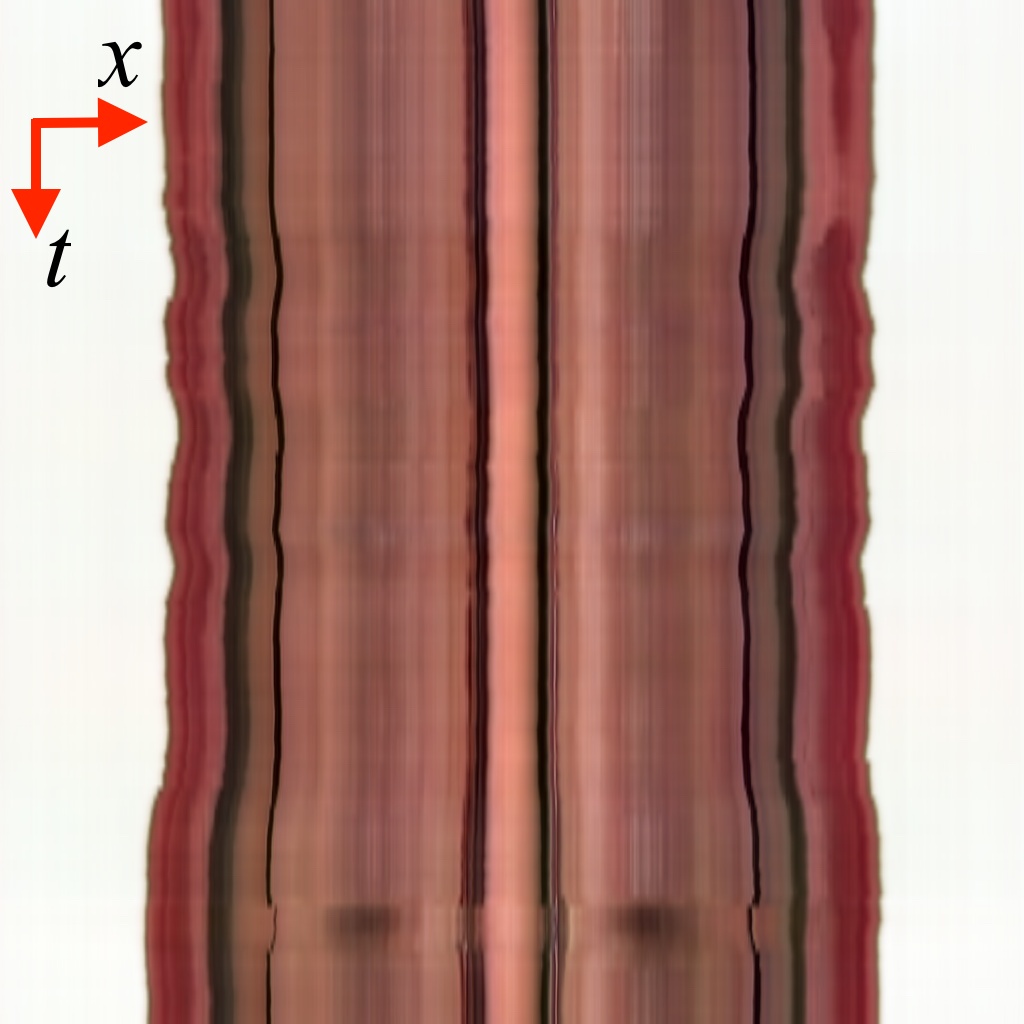}}_%
    {\substack{\vspace{-3.0mm}\\\colorbox{white}
    {Implicitly update $W$}}}$\vspace{2mm}

\vspace{-8mm}
\caption{
\textbf{x-t slices between updating latent codes explicitly and implicitly with an MLP.}
We visualize the optimized frames and an x-t slice at $y=500$. 
Explicitly updating latent code $W$ gives us an unstable x-t scanline, while updating $W$ implicitly with an MLP gives a smooth scanline.
}
\vspace{-8mm}
\label{fig:why_mlp}
\end{figure}

The reason we use an MLP to update the latent code \emph{implicitly} is that
we observe that \emph{explicitly} optimizing the latent codes results in an unstable optimization when using a large learning rate. 
However, the running time becomes too long when using a small learning rate. 
To address this, we introduce an MLP to predict the residual and update the latent codes \emph{implicitly}. 
This leads to a more stable optimization.
We show an example of x-t scanline in Fig.~\ref{fig:why_mlp} to demonstrate the effectiveness of introducing the MLP. 

\topic{Phase 2: Generator update.}
For in-domain editing, in this phase, we use the updated latent codes $\{\hat{W}^{edit}_t\}_{t=1}^{T}$ from phase 1, and our goal is to finetune the generator only to minimize:
\begin{equation}
    \hat{\theta}^{edit} = \argmin_{\hat{\theta}^{edit}}\mathcal{L}_{II}  = \argmin_{\hat{\theta}^{edit}}  \sum_{t \neq anc} \mathcal{L}_{photo} + \lambda_{\epsilon}\mathcal{L}_{\epsilon} + \lambda_{r}\mathcal{L}_{r} + \lambda_{M}\mathcal{L}_{M}\,,
\end{equation}
\vspace{-5mm}
\begin{equation}
    \mathcal{L}_{M} = (1 - M^{PD}_i)\mathcal{L}_{LPIPS}(\hat{I}^{''}_{i}, I^{in}_i) + (1 - M^{PD}_{anc})\mathcal{L}_{LPIPS}(\hat{I}^{''}_{anc}, I^{in}_{anc}) \,.
\end{equation}
$M^{PD}_i$ is the perceptual difference mask computed between $\hat{I}^{''}_{i} = G(\hat{W}^{edit}_t; \hat{\theta}^{edit})$ and aligned input $I^{in}_i$, and $\mathcal{L}_{LPIPS}(\cdot,\cdot)$ is the LPIPS distance loss~\cite{zhang2018unreasonable}. We initialize $\hat{\theta}^{edit}$ as ${\theta}^{edit}$.
The LPIPS term also plays a role to maintain the sharpness of the edited frames. 
This is because the consistency can be achieved by pushing all the frames to become blurry.

Here, $\mathcal{L}_{r}$ is the regularization loss for the generator and $\lambda_{r}$ is the strength of regularization. 
We introduce this loss to help prevent the generator $G$ from losing its latent space editability as we do not wish to \emph{ruin} its pretrained latent space. 
Therefore, similar to~\cite{roich2021pivotal}, we use this \emph{local regularization} to preserve the editing ability of our generator. 
More specifically, we first obtain a latent code $W_r$ by linearly interpolating between the current latent code $\hat{W}^{edit}_t$ and a randomly sampled code $W_z$ with an interpolation parameter $\alpha_{interp}$: $W_r = \hat{W}^{edit}_t + \alpha_{interp} \frac{W_z - \hat{W}^{edit}_t}{||W_z - \hat{W}^{edit}_t||_2}$.
This gives us a new latent code in a local region around $\hat{W}^{edit}_t$.
To ensure that we do not lose the editing capability of the original generator, we add a penalty on the distance between the generated image from the new generator and the old one such that:
\begin{equation}
    \mathcal{L}_{r} = \mathcal{L}_{LPIPS}(x_r, \hat{x}_r) + \lambda_{\ell_2}^{r}\mathcal{L}_{\ell_2}(x_r, \hat{x}_r) \,,
\end{equation}
where $x_r = G(W_r; \theta^{edit}), \hat{x}_r = G(W_r; \hat{\theta}^{edit})$, $\lambda_{\ell_2}^{r}$ is the weight for $\ell_2$ loss. This regularization can alleviate the side effects from updating $G$ within a local area. This is desirable since for a video, the latent codes for the same identity tend to gather locally.

For out-of-domain editing, unlike in-domain editing, we cannot rely on the perceptual difference mask, so the optimization goal reduces to:
\begin{equation}
    \hat{\theta}^{edit} = \argmin_{\hat{\theta}^{edit}}\mathcal{L}_{II} = \argmin_{\hat{\theta}^{edit}} \sum_{t \neq anc} \mathcal{L}_{photo} + \lambda_{r}\mathcal{L}_{r} + \lambda_{\epsilon}\mathcal{L}_{\epsilon} \,.
\end{equation}
To compensate for the regularization effect of the perceptual difference mask, we freeze the last eight layers of the synthesis network in $G$ to avoid blurry results.
As all the computations, including the GAN generator, flow estimation network, spatial warping, and photometric losses, are \emph{differentiable}, we can backpropagate the errors all the way back. 
After phase 1 and 2, we will have $\{\hat{W}^{edit}_t\}_{t=1}^T$ and $G(\cdot;\hat{\theta}^{edit})$ as a result. 


\subsection{Phase 3: Unalign}
After our two-phase optimization, we perform our final phase as post-processing. 
In this phase, we put the aligned frames back to the original video to generate our final edited video (see Figure~\ref{fig:approach}). 
We follow the \emph{stitch tuning} approach in \cite{tzaban2022stitch} by tuning the generator to reduce the edge artifact brought by editing.
Note that this is only feasible for the in-domain editing because the out-of-domain editing has a global appearance compared to the input video.

\section{Experimental Results}
\label{sec:result}
\subsection{Experimental setup}
\begin{figure*}[t]
\centering
\mpage{0.01}{\raisebox{60pt}{\rotatebox{90}{Input}}}  \hfill
\frame{\includegraphics[trim=0 20 0 40, clip,width=.14\textwidth]{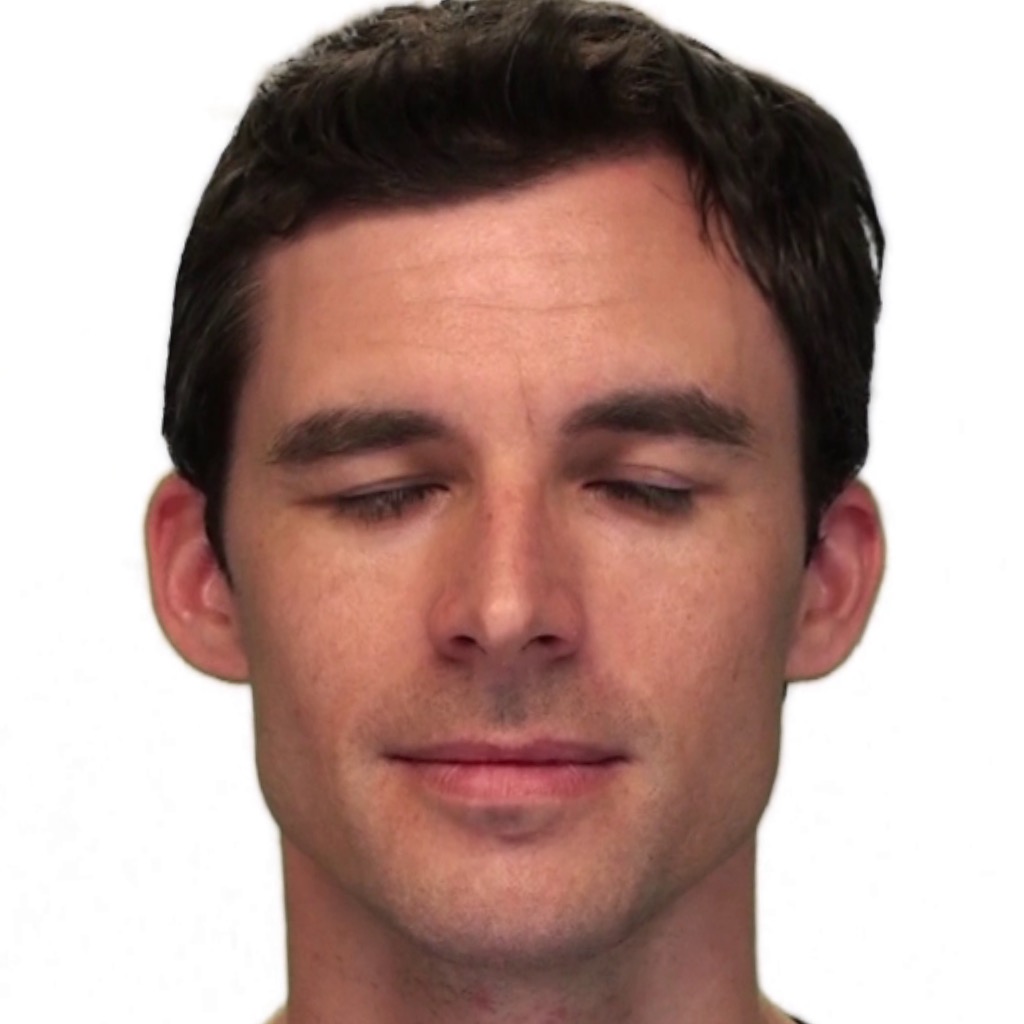}}\hfill \hspace{-5mm}
\frame{\includegraphics[trim=0 20 0 40, clip,width=.14\textwidth]{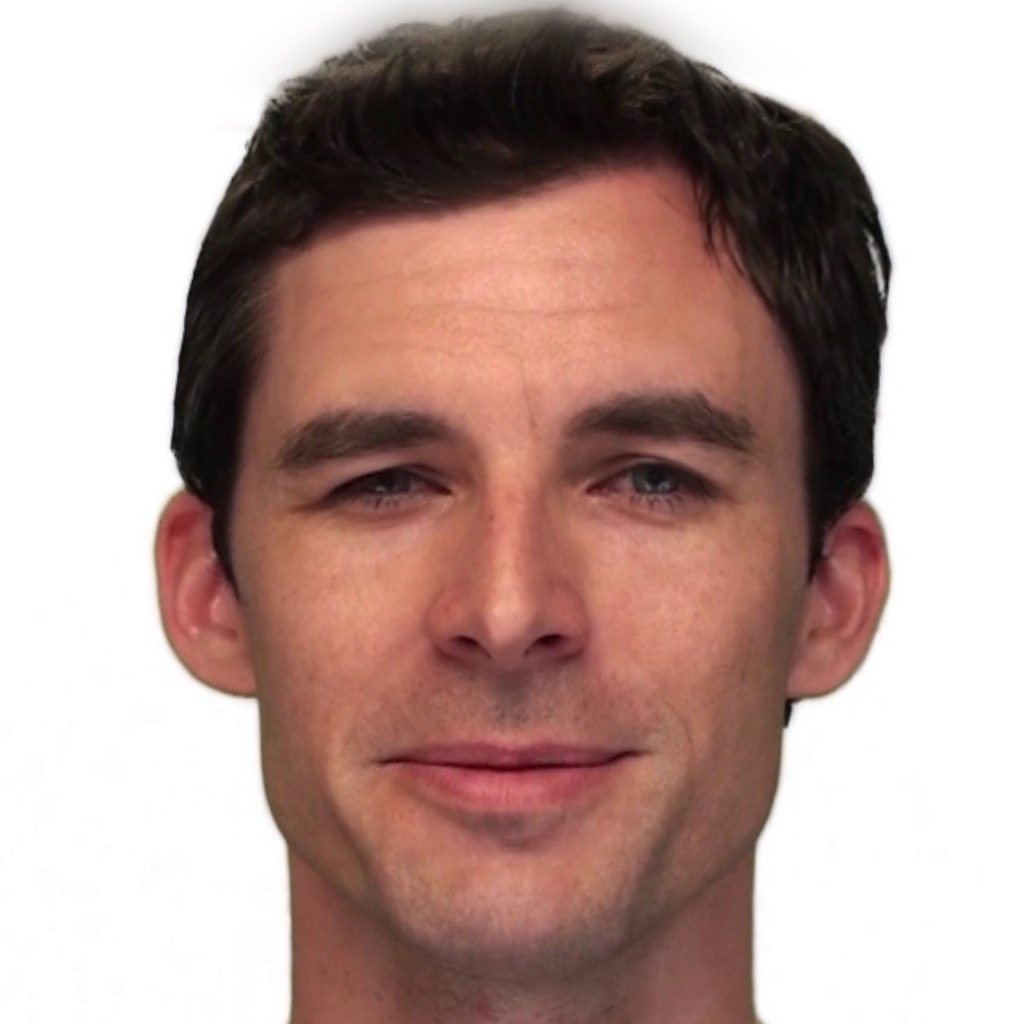}}\hfill \hspace{-5mm}
\frame{\includegraphics[trim=0 20 0 40, clip,width=.14\textwidth]{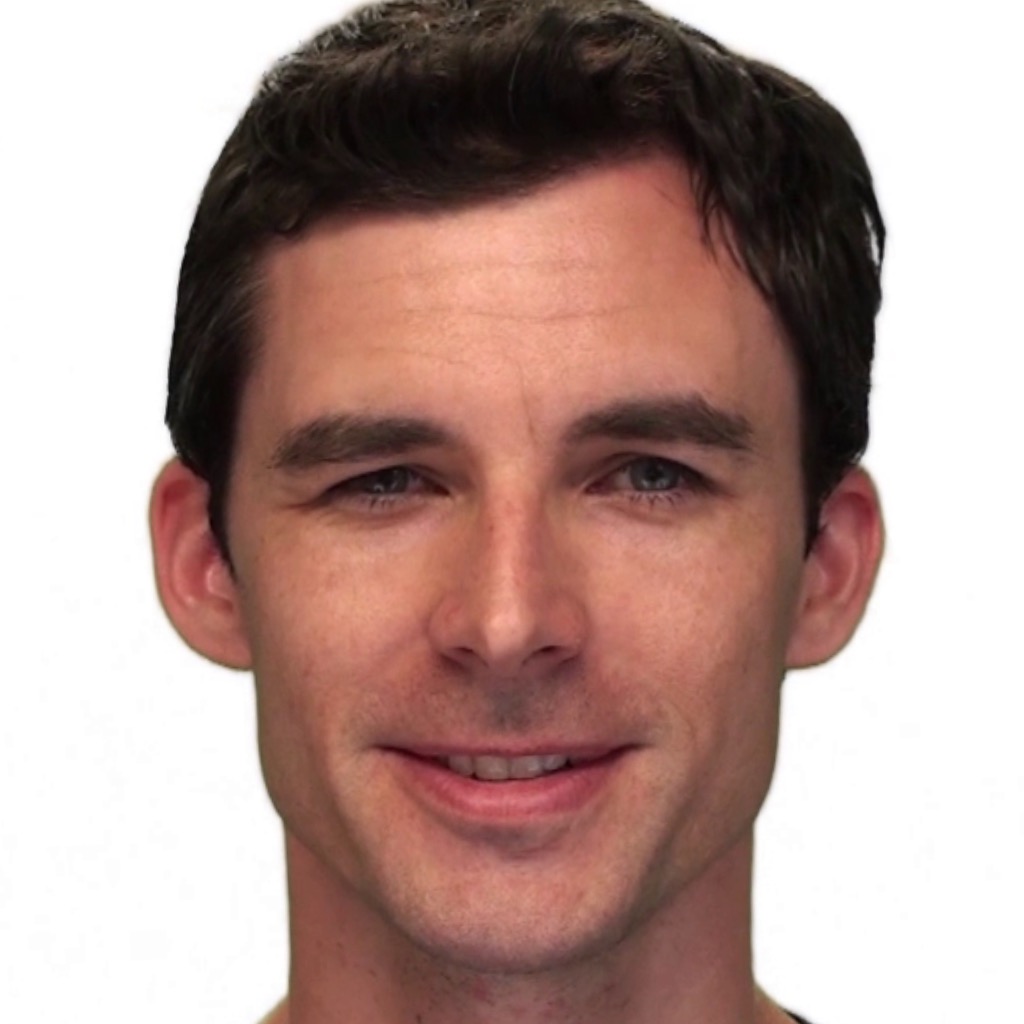}}\hfill
\hspace{2mm}
\frame{\includegraphics[trim=0 20 0 40, clip,width=.14\textwidth]{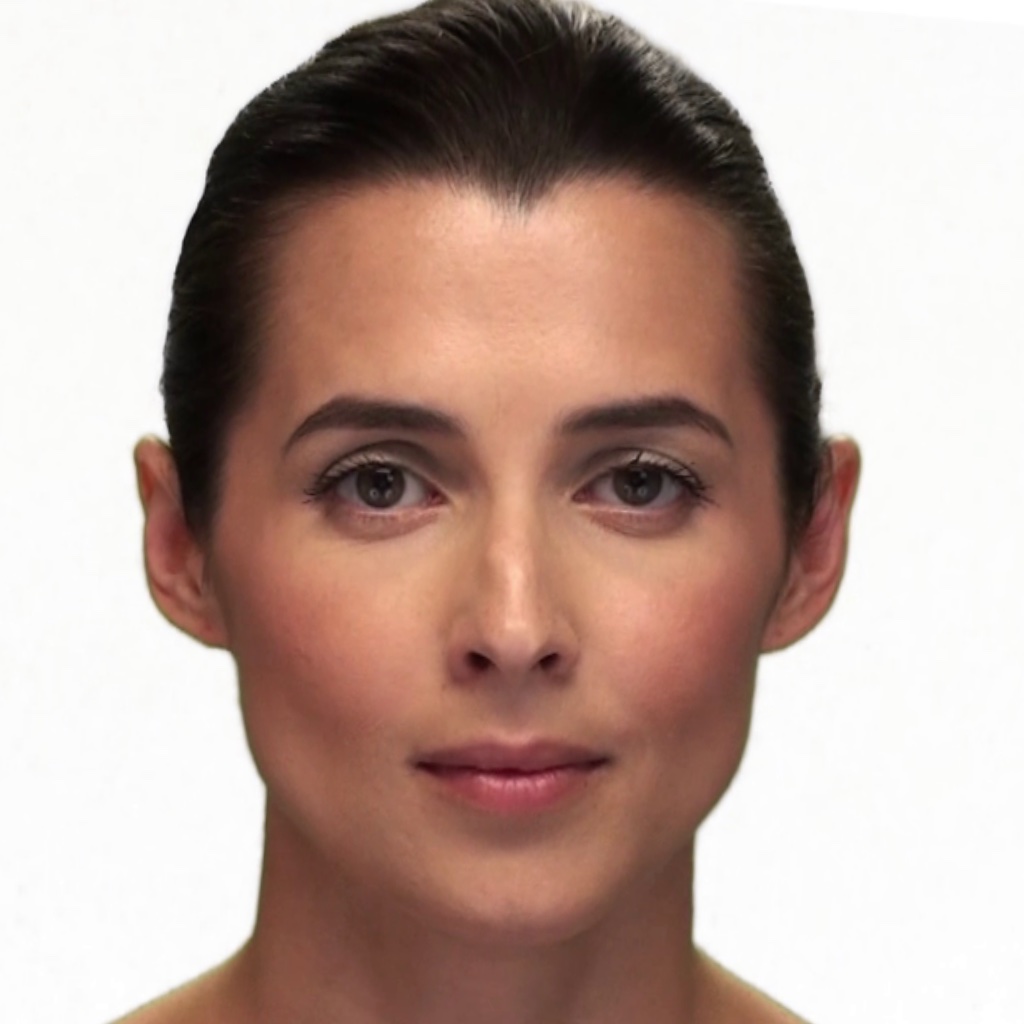}}\hfill\hspace{-5mm}
\frame{\includegraphics[trim=0 20 0 40, clip,width=.14\textwidth]{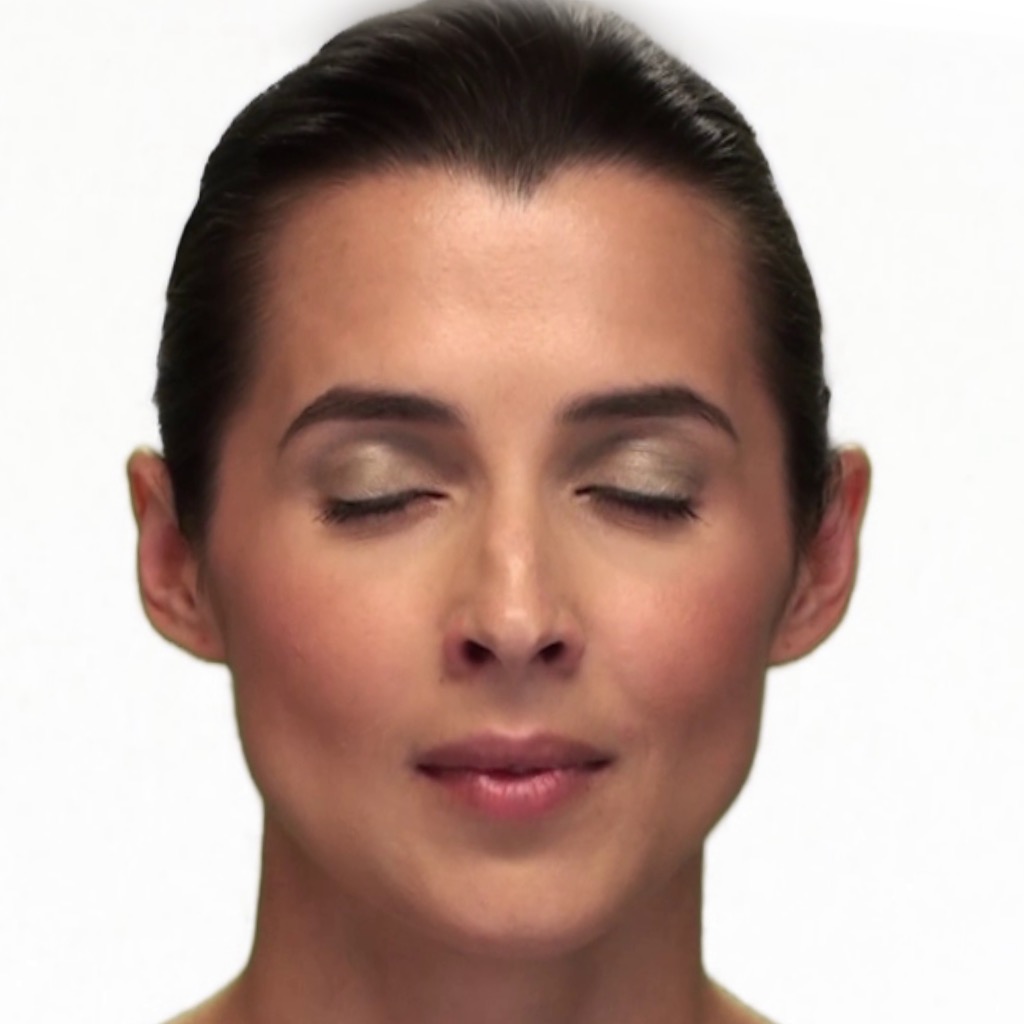}}\hfill\hspace{-5mm}
\frame{\includegraphics[trim=0 20 0 40, clip,width=.14\textwidth]{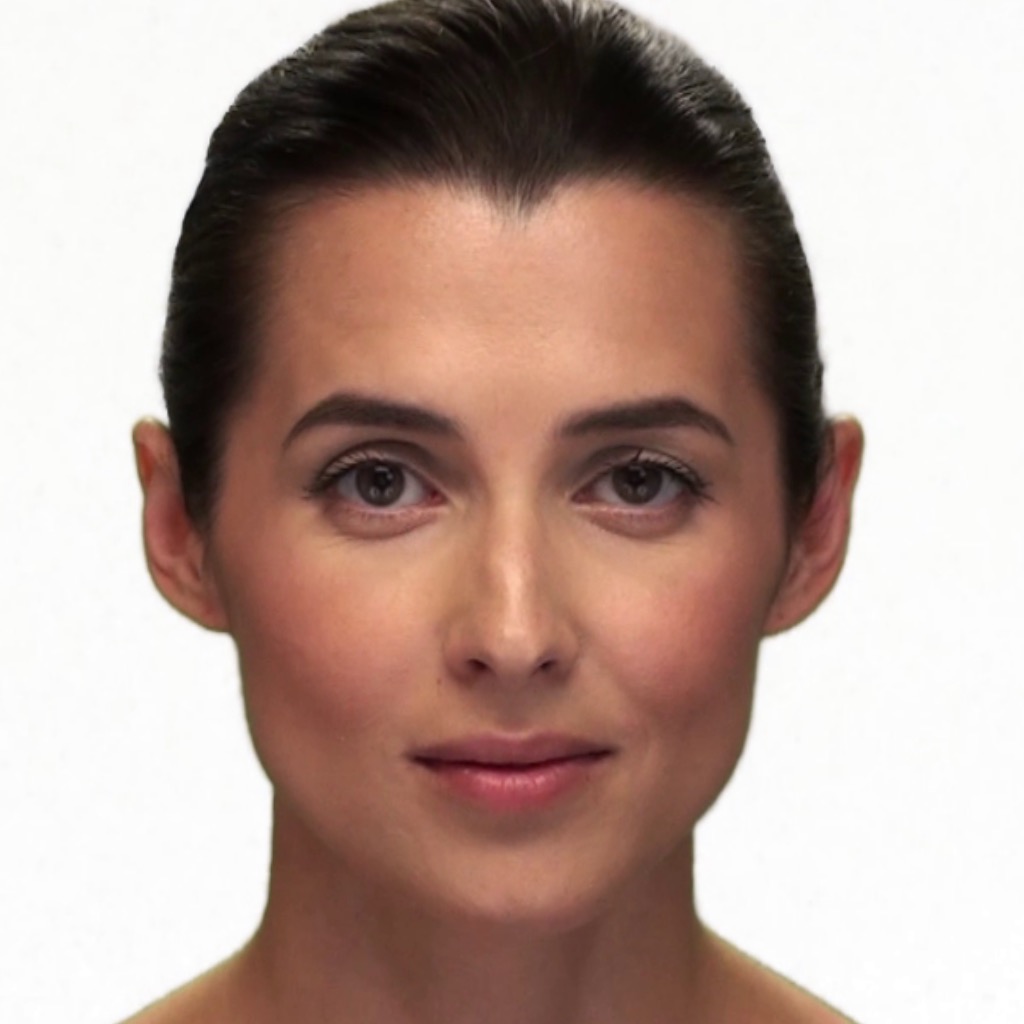}}\\

\vspace{-13.5mm}
\mpage{0.01}{\raisebox{60pt}{\rotatebox{90}{Ours}}}  \hfill
\frame{\includegraphics[trim=0 20 0 40, clip,width=.14\textwidth]{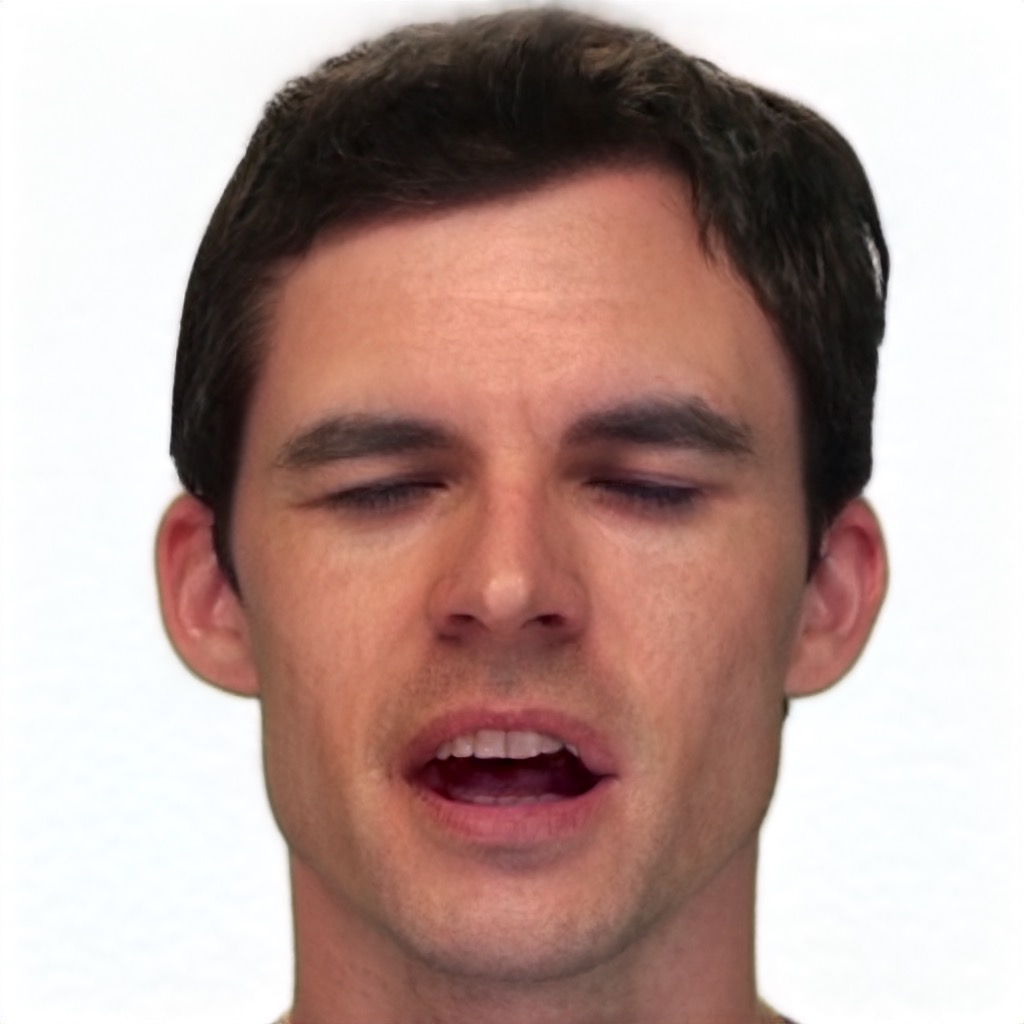}}\hfill\hspace{-5mm}
\frame{\includegraphics[trim=0 20 0 40, clip,width=.14\textwidth]{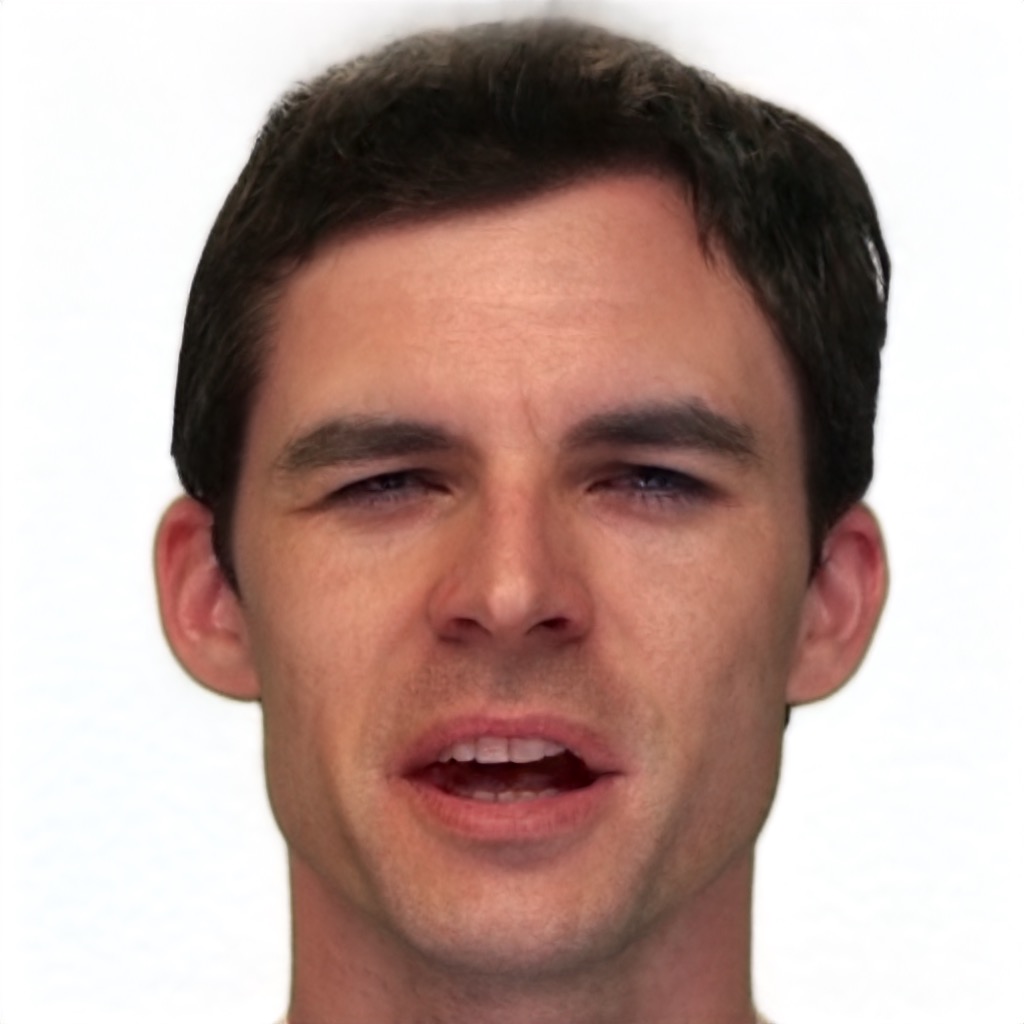}}\hfill\hspace{-5mm}
\frame{\includegraphics[trim=0 20 0 40, clip,width=.14\textwidth]{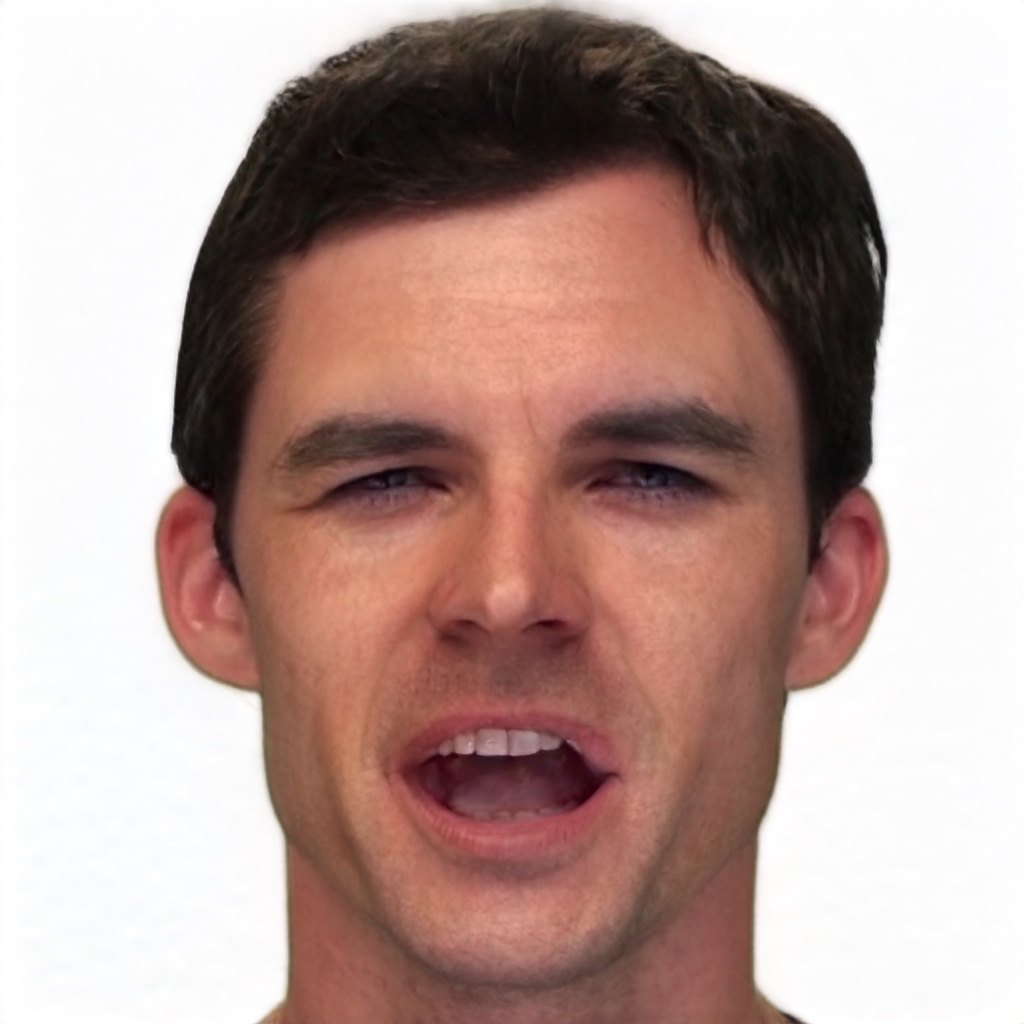}}\hfill
\hspace{2mm}
\frame{\includegraphics[trim=0 20 0 40, clip,width=.14\textwidth]{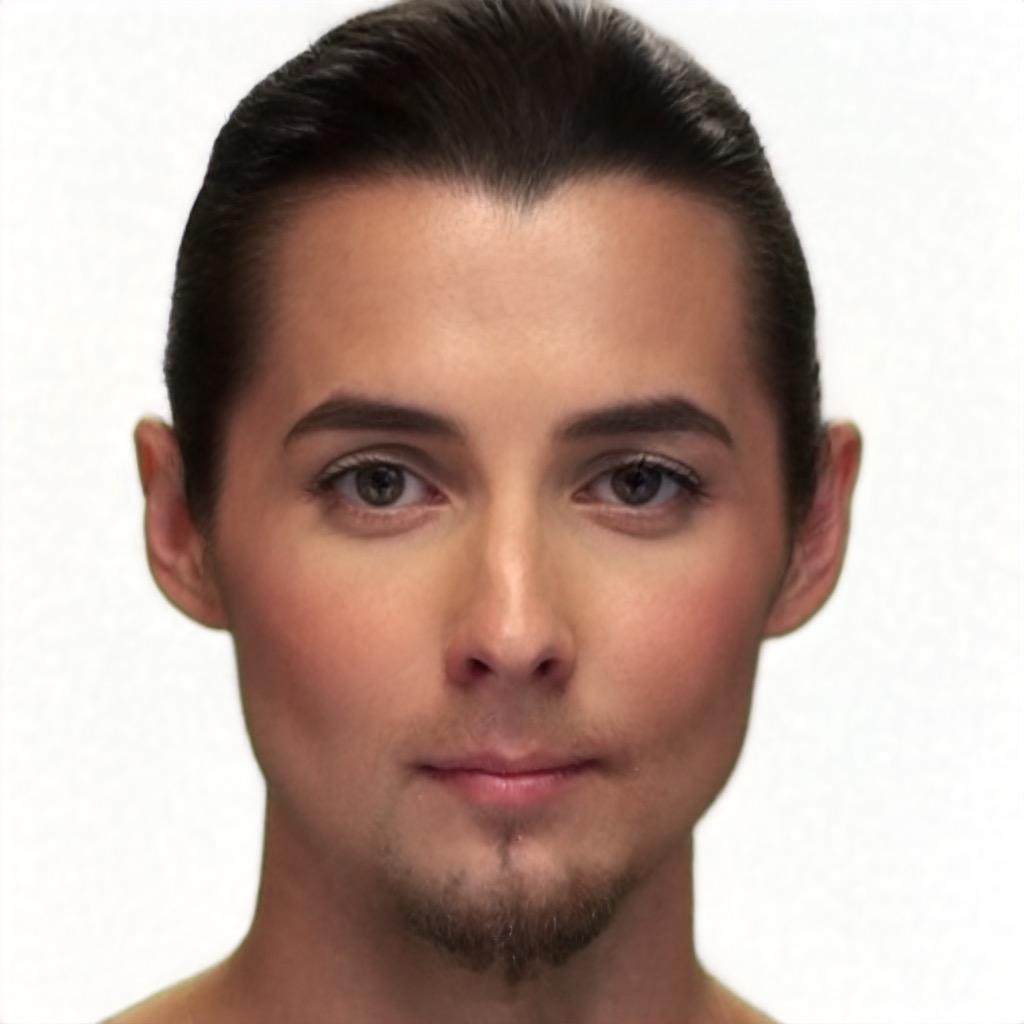}}\hfill\hspace{-5mm}
\frame{\includegraphics[trim=0 20 0 40, clip,width=.14\textwidth]{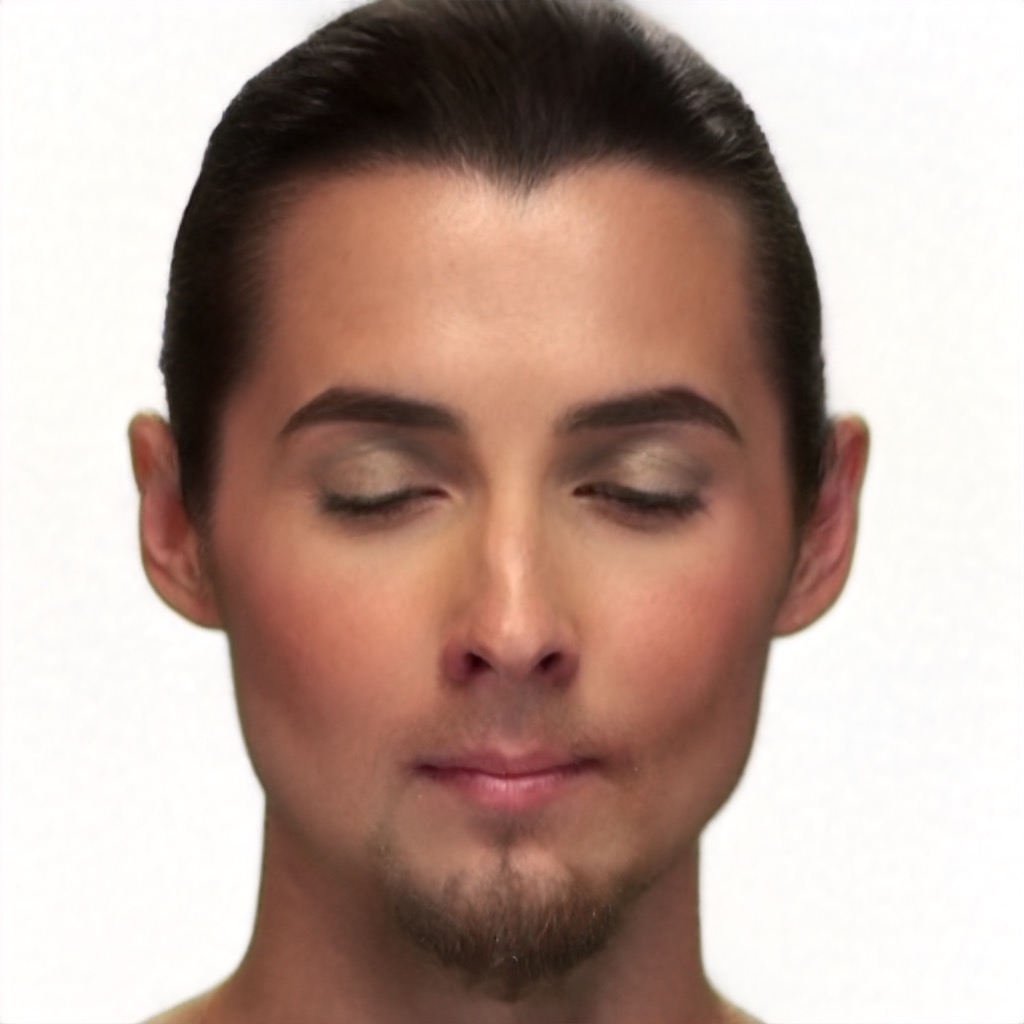}}\hfill\hspace{-5mm}
\frame{\includegraphics[trim=0 20 0 40, clip,width=.14\textwidth]{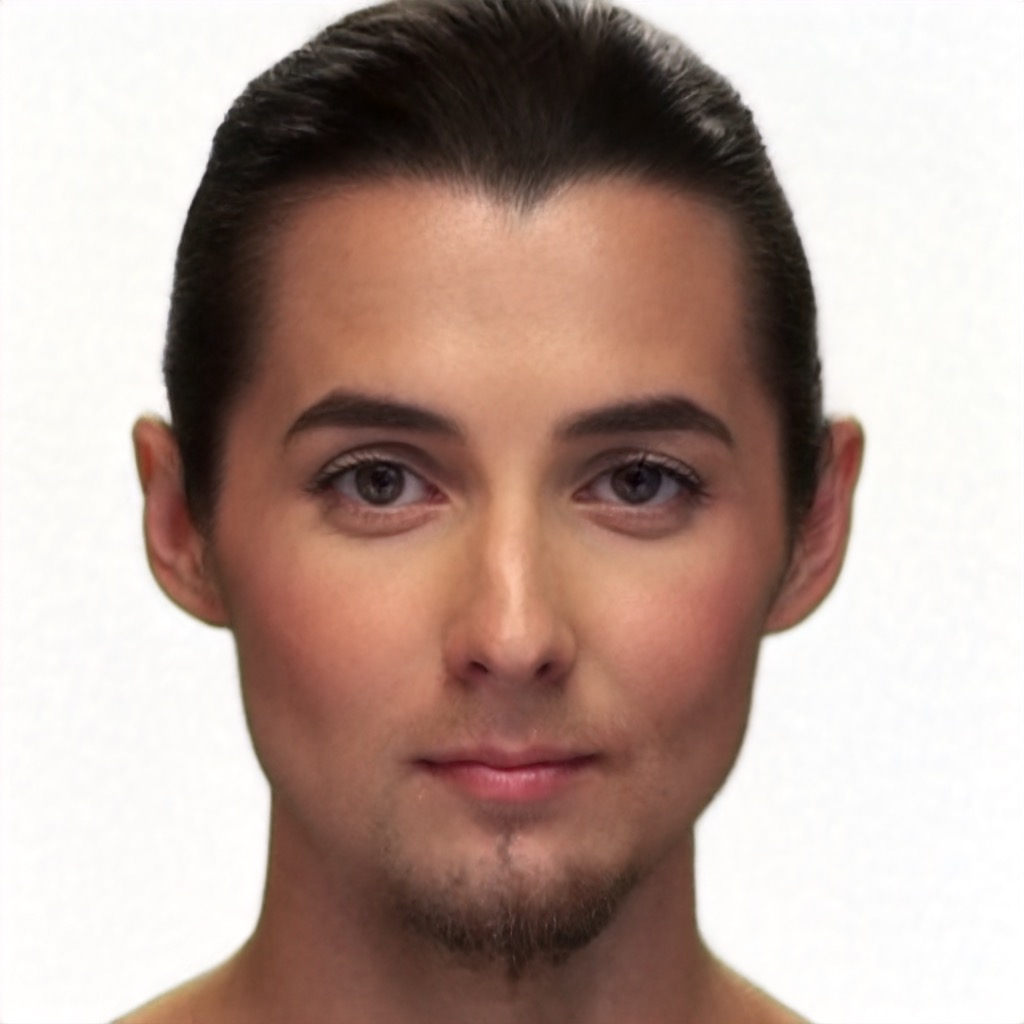}}\\

\vspace{-13mm}
\hspace{5mm}\mpage{0.45}{``angry"} \hfill
\mpage{0.45}{``beard"}
\vspace{0.5mm}

\mpage{0.01}{\raisebox{60pt}{\rotatebox{90}{Input}}}  \hfill
\frame{\includegraphics[trim=0 20 0 40, clip,width=.14\textwidth]{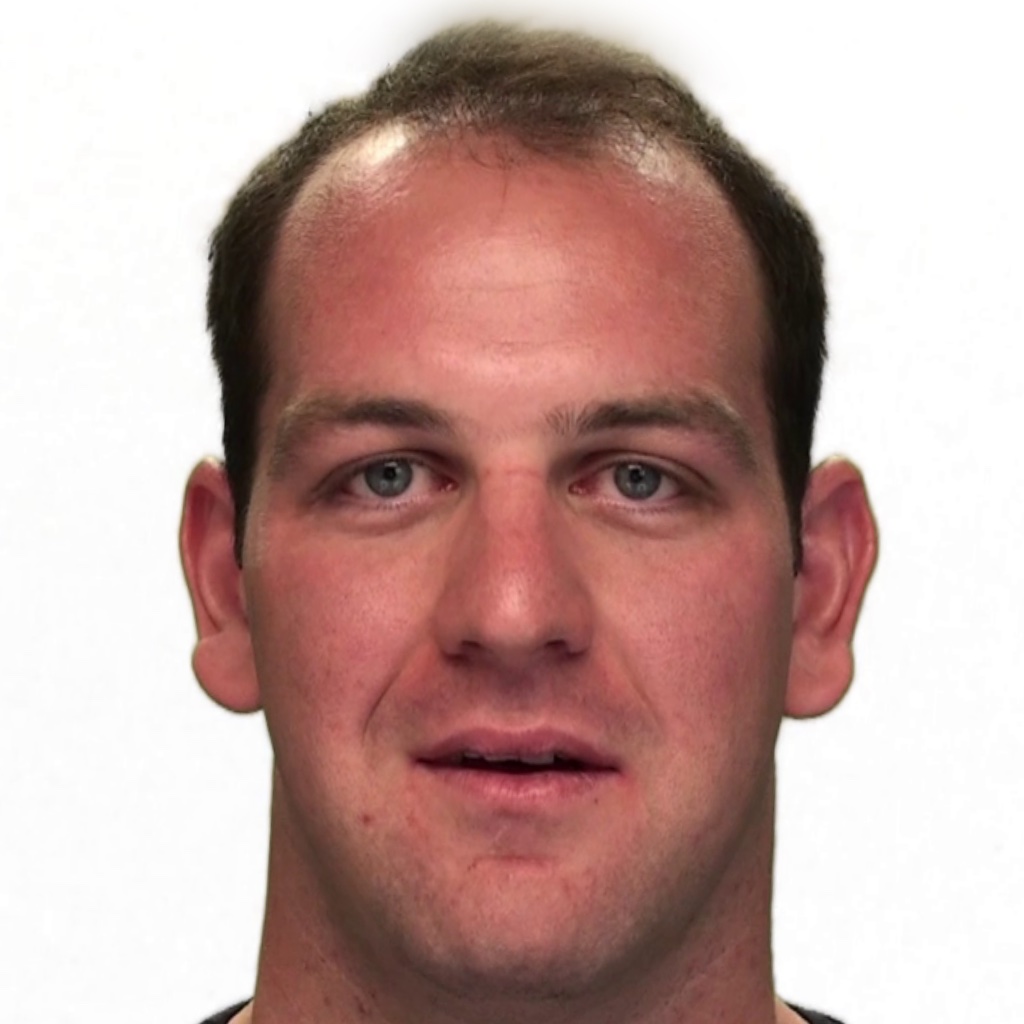}}\hfill\hspace{-5mm}
\frame{\includegraphics[trim=0 20 0 40, clip,width=.14\textwidth]{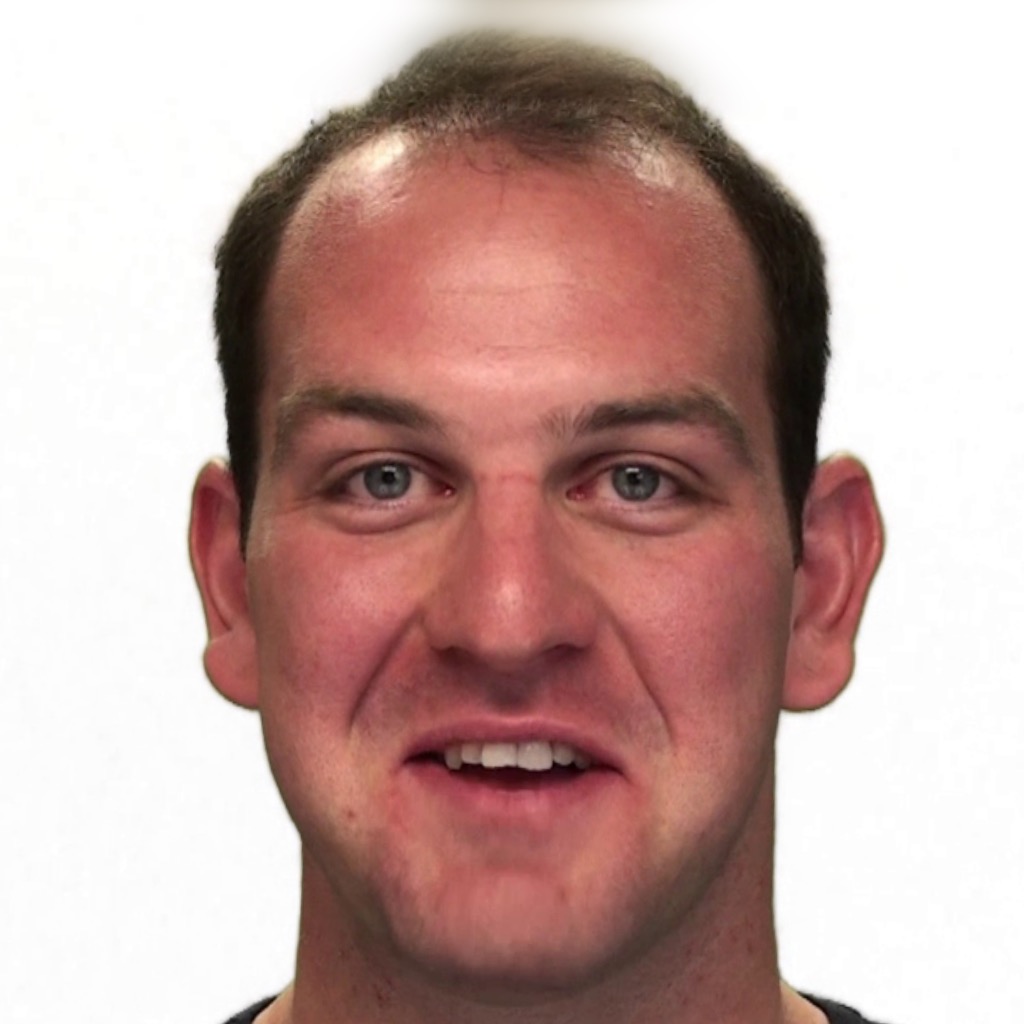}}\hfill\hspace{-5mm}
\frame{\includegraphics[trim=0 20 0 40, clip,width=.14\textwidth]{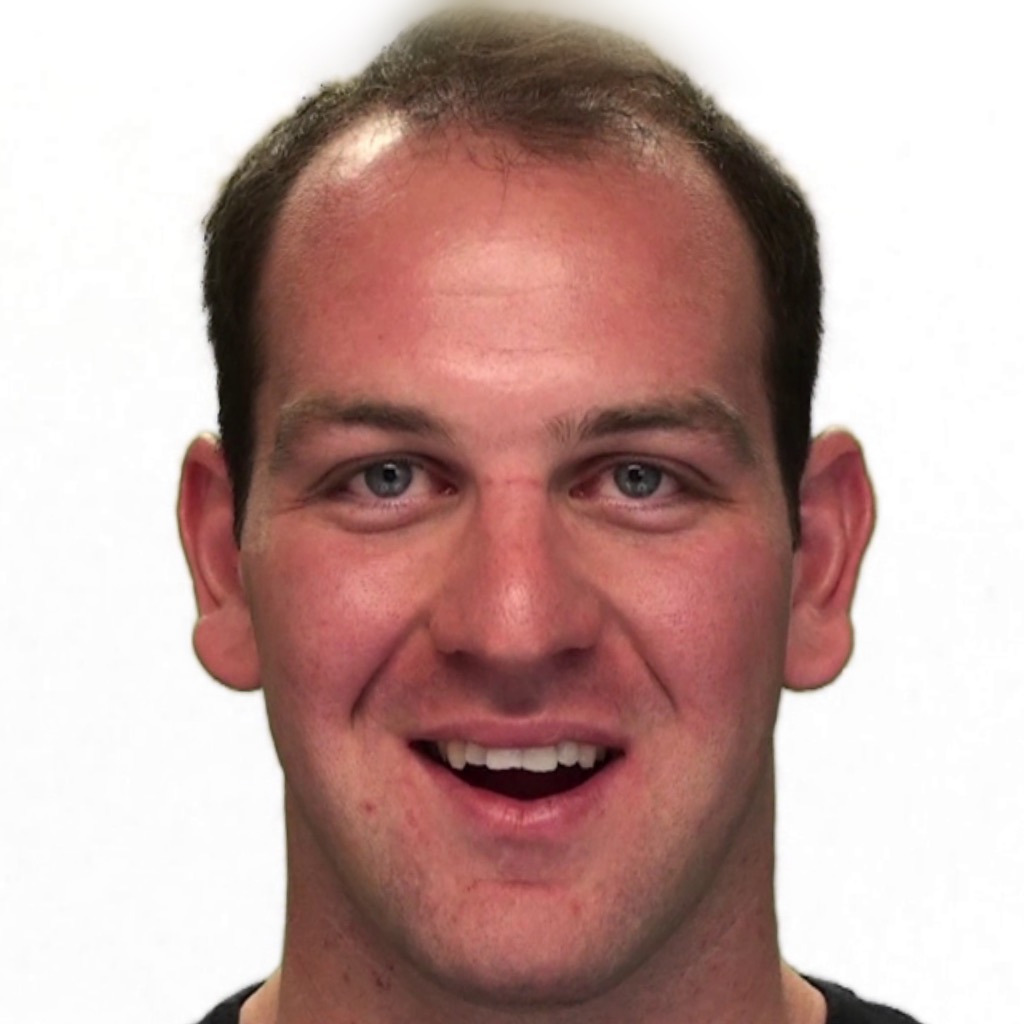}}\hfill
\hspace{2mm}
\frame{\includegraphics[trim=0 20 0 40, clip,width=.14\textwidth]{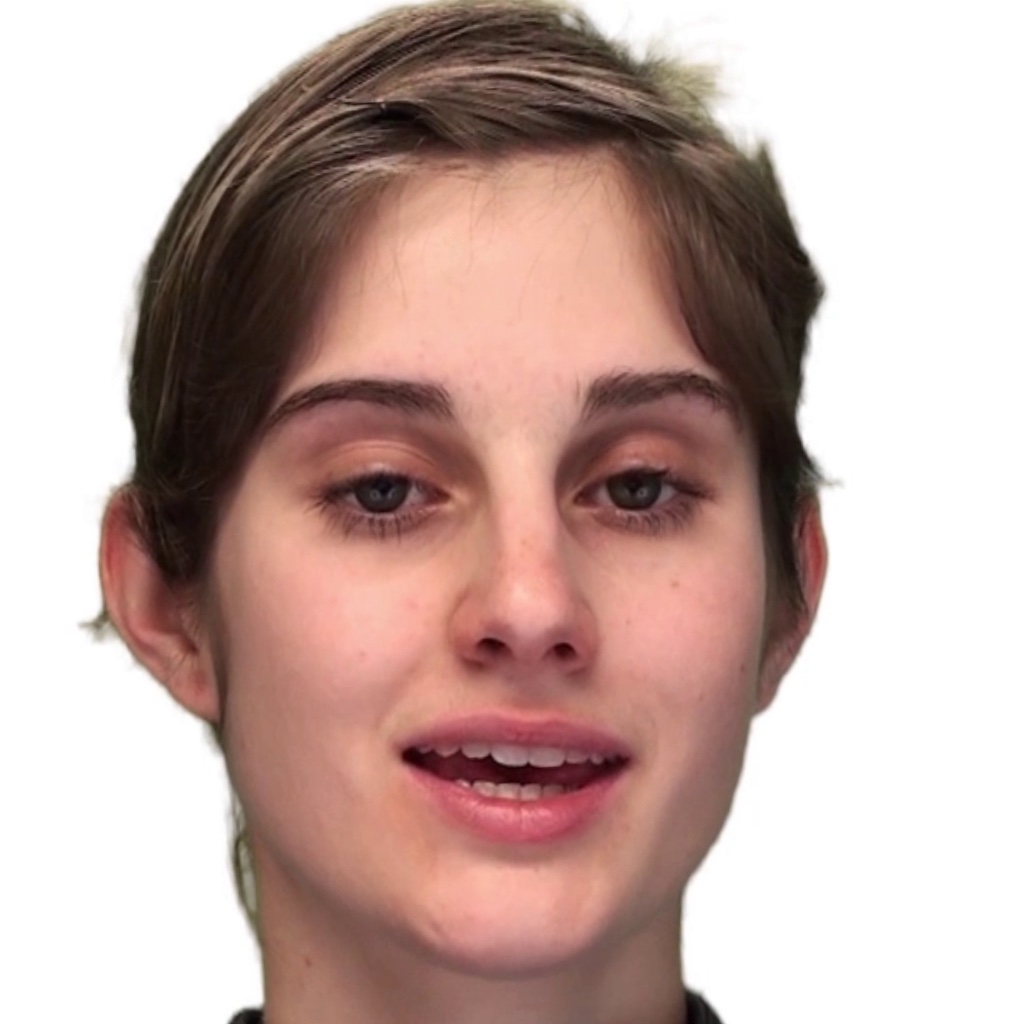}}\hfill\hspace{-5mm}
\frame{\includegraphics[trim=0 20 0 40, clip,width=.14\textwidth]{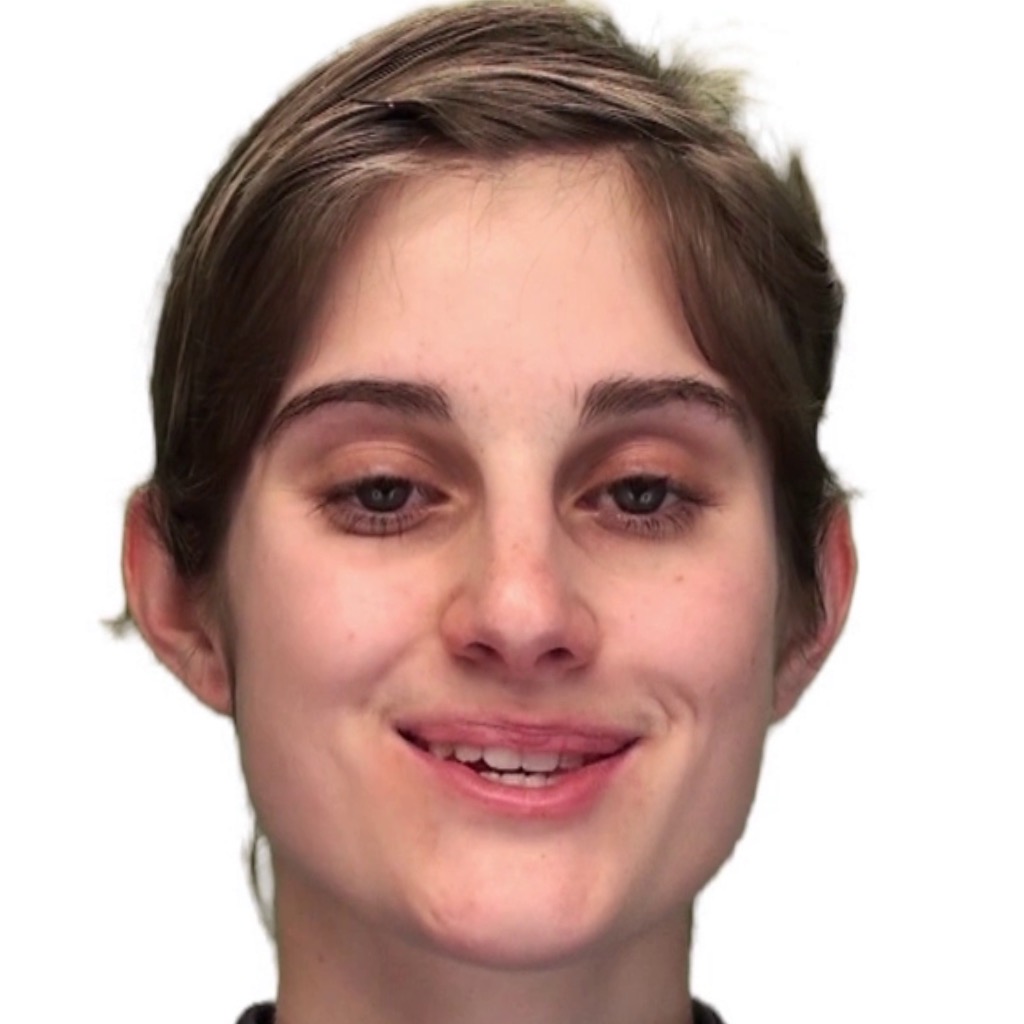}}\hfill\hspace{-5mm}
\frame{\includegraphics[trim=0 20 0 40, clip,width=.14\textwidth]{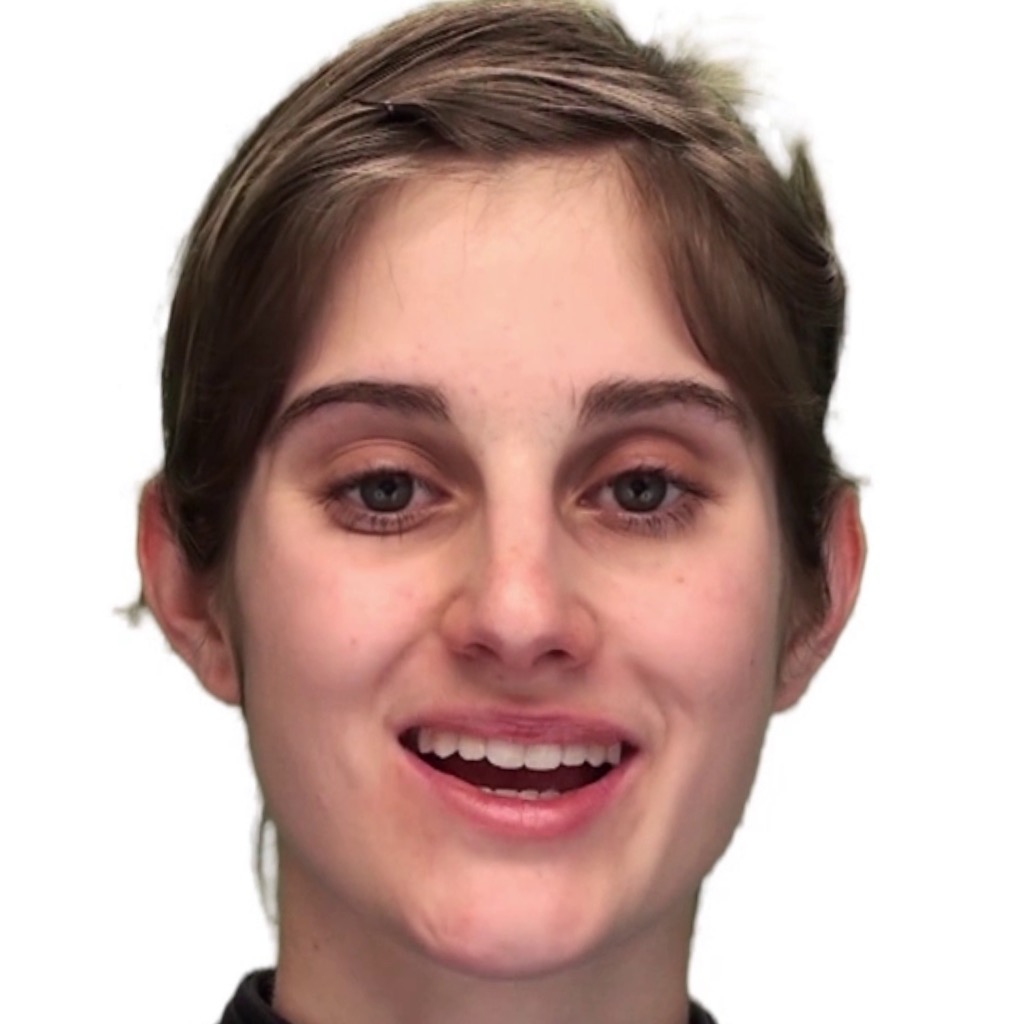}}\\

\vspace{-13.5mm}
\mpage{0.01}{\raisebox{60pt}{\rotatebox{90}{Ours}}}  \hfill
\frame{\includegraphics[trim=0 20 0 40, clip,width=.14\textwidth]{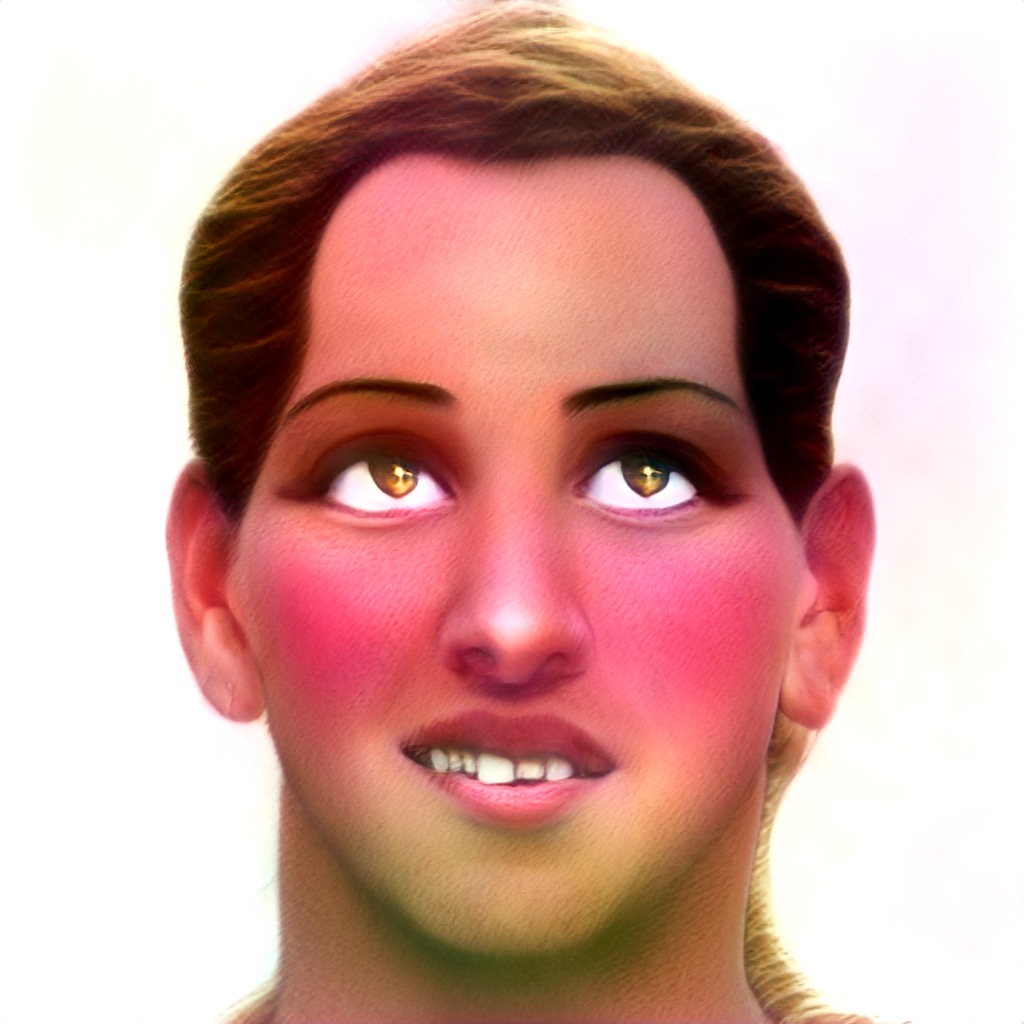}}\hfill\hspace{-5mm}
\frame{\includegraphics[trim=0 20 0 40, clip,width=.14\textwidth]{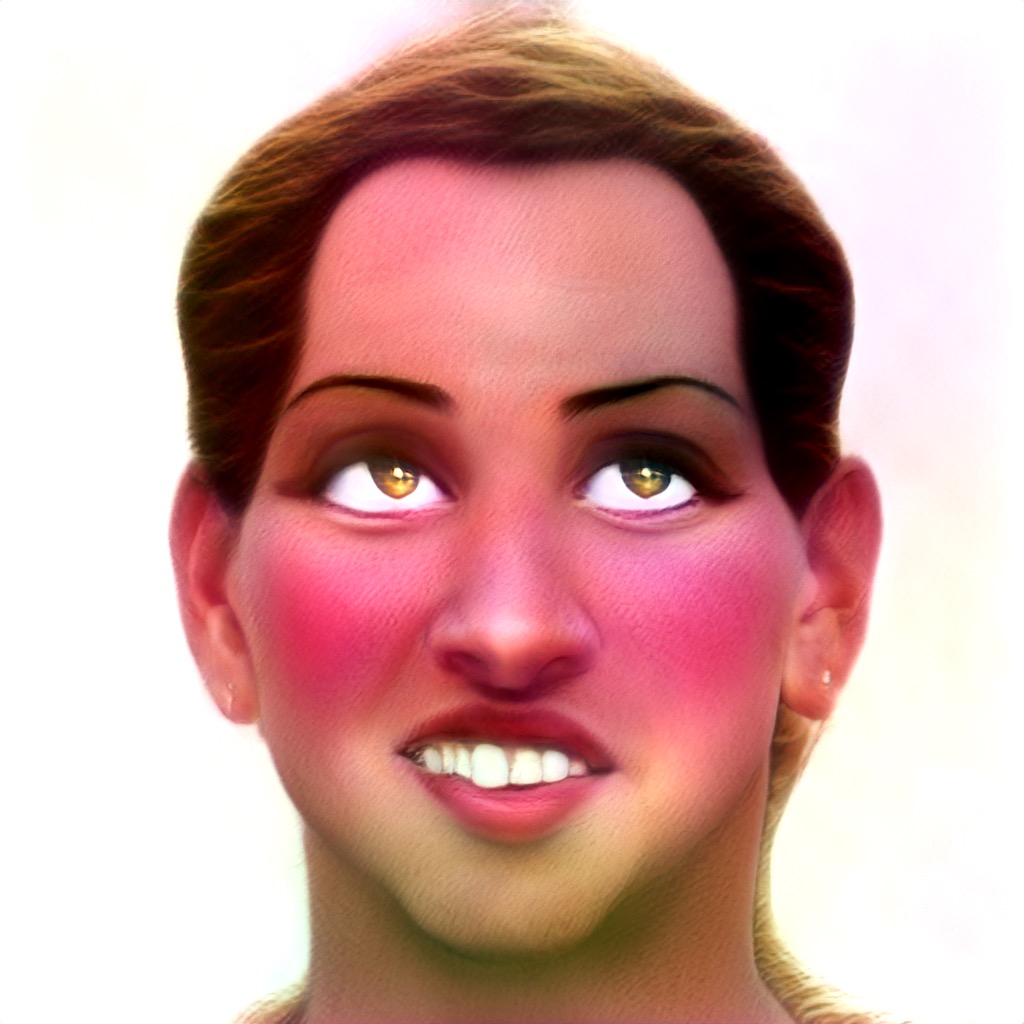}}\hfill\hspace{-5mm}
\frame{\includegraphics[trim=0 20 0 40, clip,width=.14\textwidth]{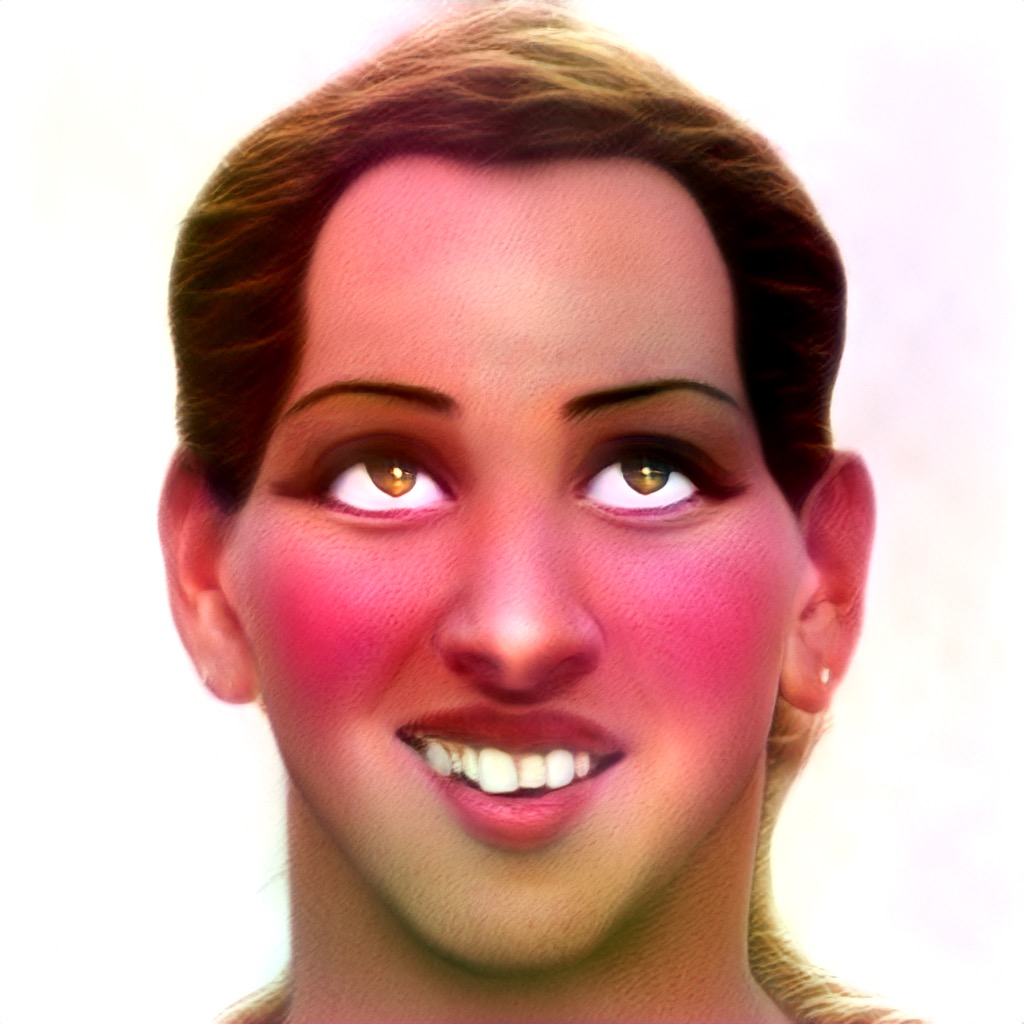}}\hfill
\hspace{2mm}
\frame{\includegraphics[trim=0 20 0 40, clip,width=.14\textwidth]{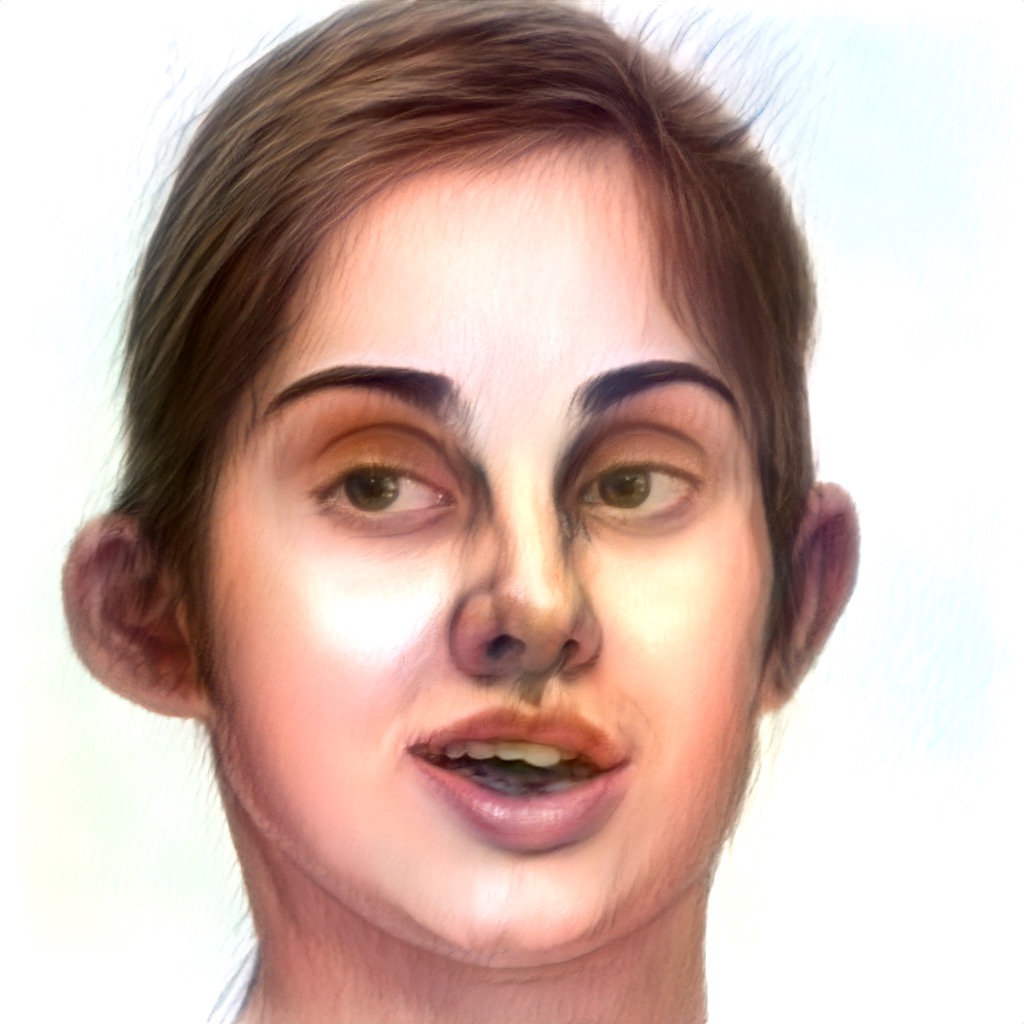}}\hfill\hspace{-5mm}
\frame{\includegraphics[trim=0 20 0 40, clip,width=.14\textwidth]{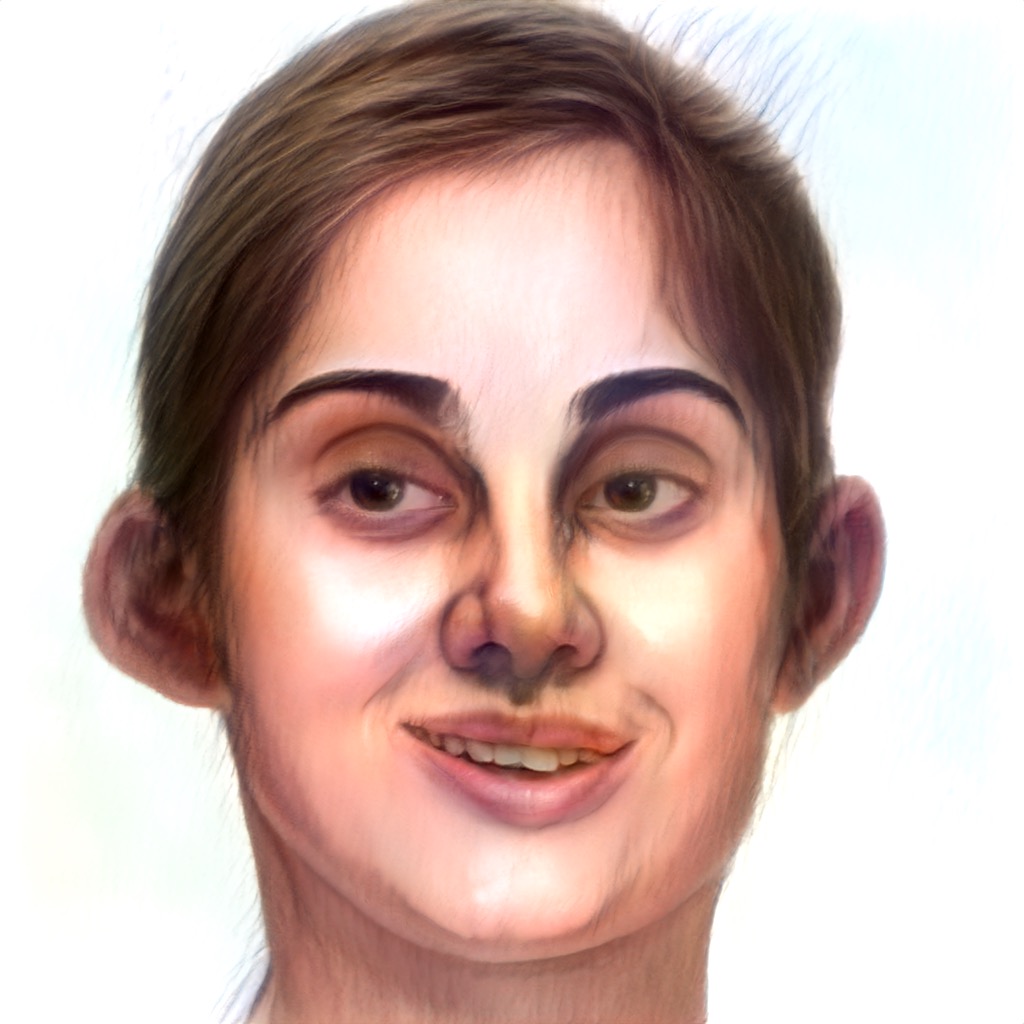}}\hfill\hspace{-5mm}
\frame{\includegraphics[trim=0 20 0 40, clip,width=.14\textwidth]{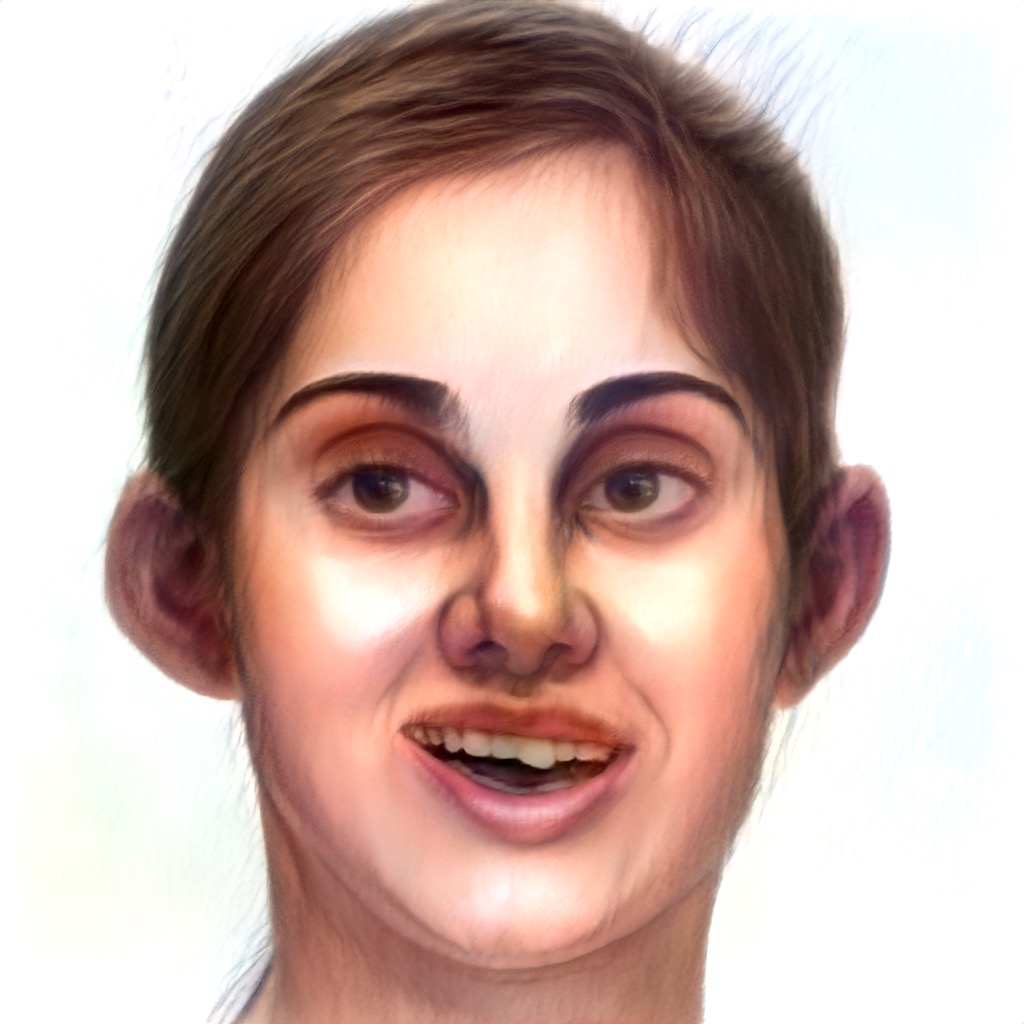}}\\

\vspace{-12.5mm}
\hspace{5mm}\mpage{0.45}{``Disney princess"}\hfill
\hspace{-5mm}\mpage{0.45}{``Sketch"}
\vspace{-2mm}

\caption{\textbf{Visual results 
on RAVDESS dataset~\cite{livingstone2018ryerson}.} 
We show both in-domain (``angry'' and ``eyeglasses'') and out-of-domain (``Disney princess'' and ``Sketch'') editing results. Our results maintain consistent changes with time preserving the temporal coherence.
}
\label{fig:results}
\vspace{-7mm}
\end{figure*}
\begin{figure*}[t]


\mpage{0.01}{\raisebox{60pt}{\rotatebox{90}{DE}}} \hfill
\frame{\includegraphics[trim=0 0 0 0, clip,width=.135\textwidth]{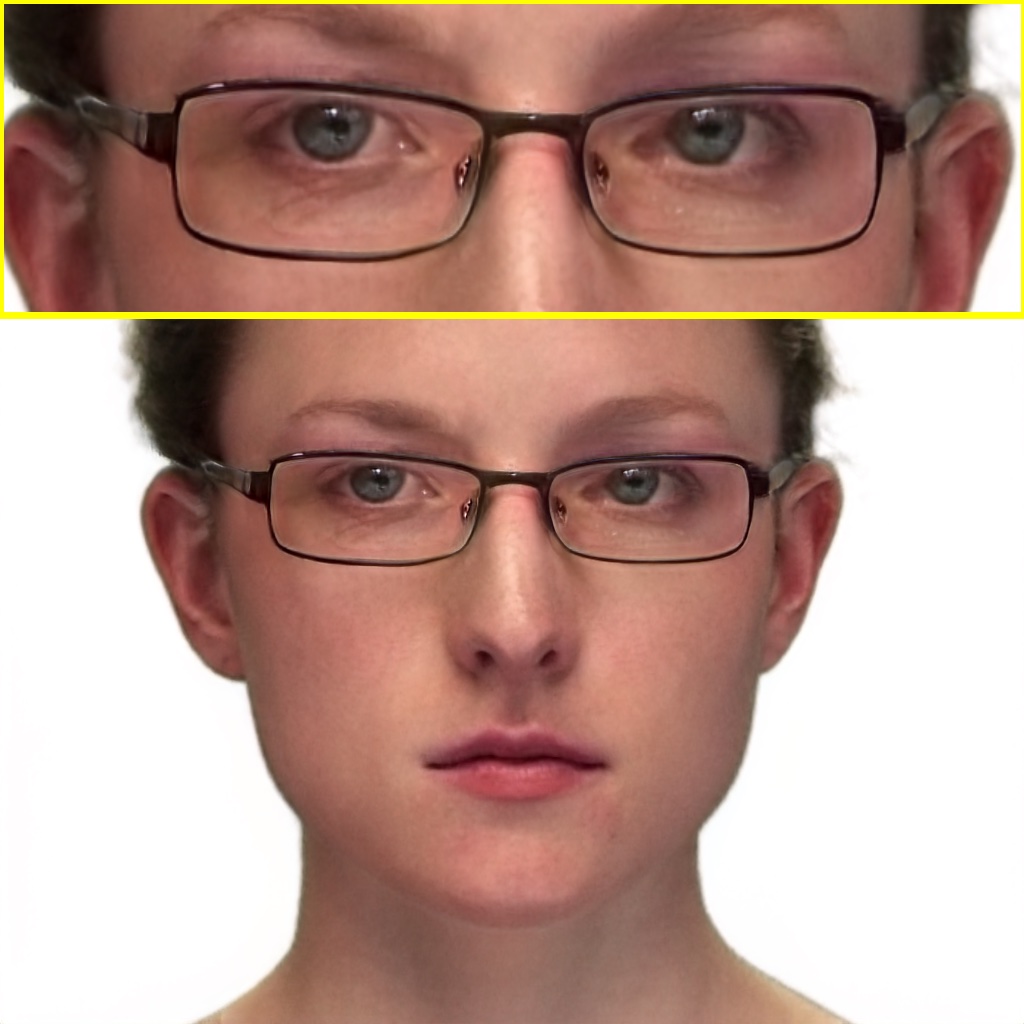}}\hfill\hspace{-5mm}
\frame{\includegraphics[trim=0 0 0 0, clip,width=.135\textwidth]{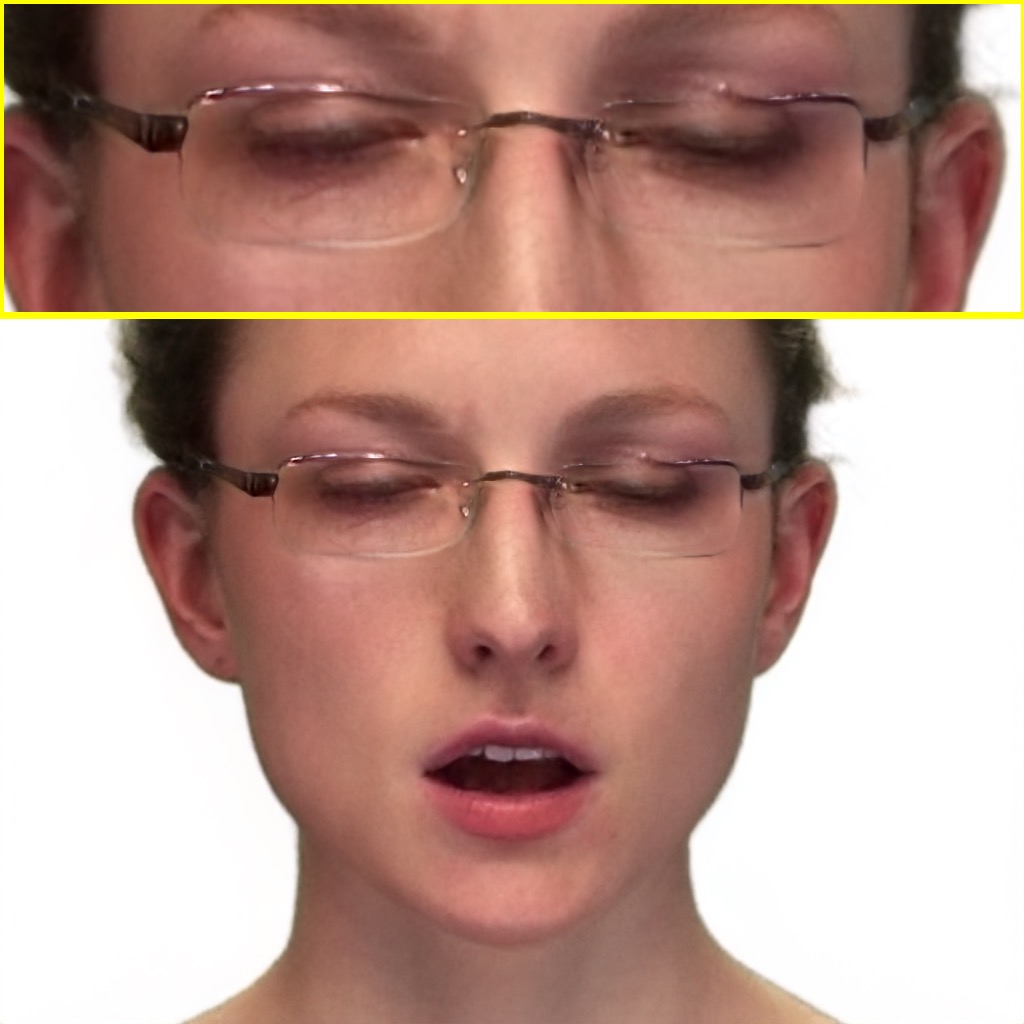}}\hfill\hspace{-5mm}
\frame{\includegraphics[trim=0 0 0 0, clip,width=.135\textwidth]{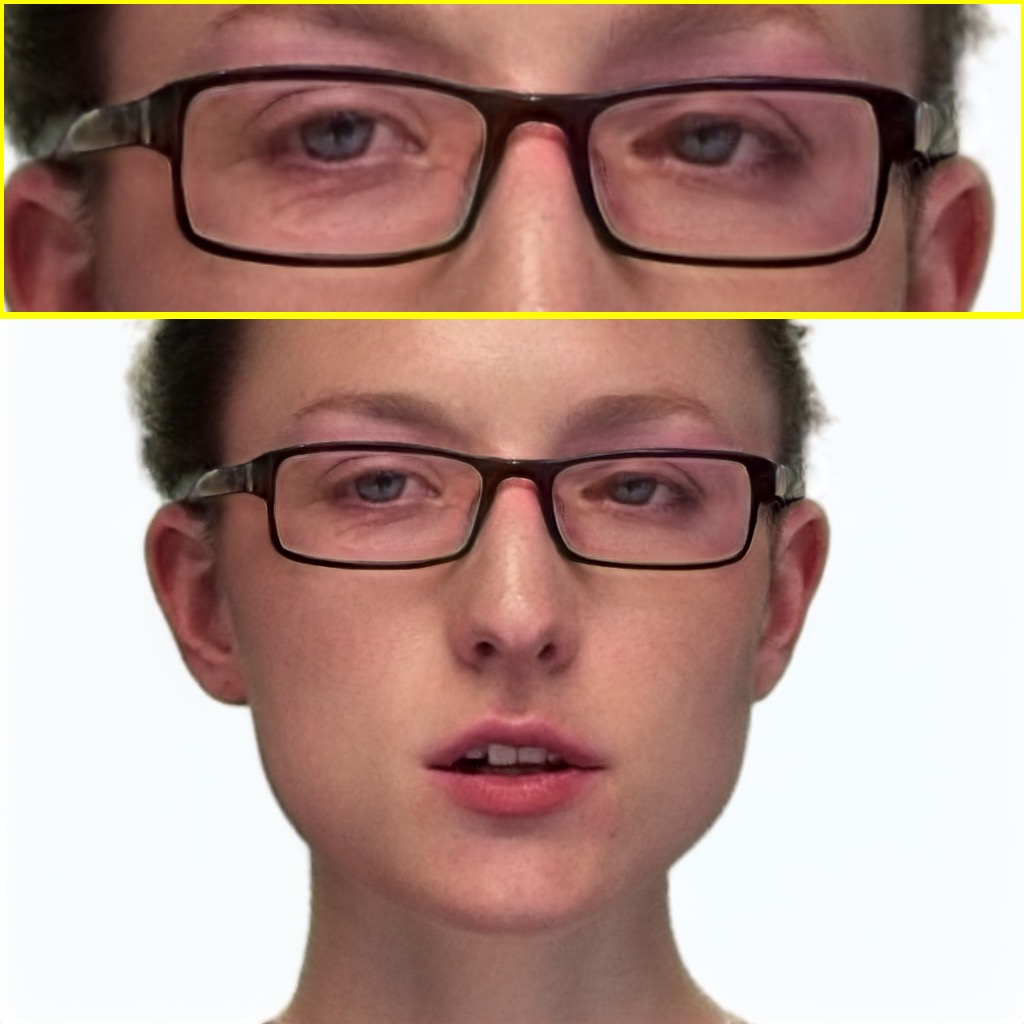}}\hfill\hspace{-5mm}
\frame{\includegraphics[trim=0 0 0 0, clip,width=.135\textwidth]{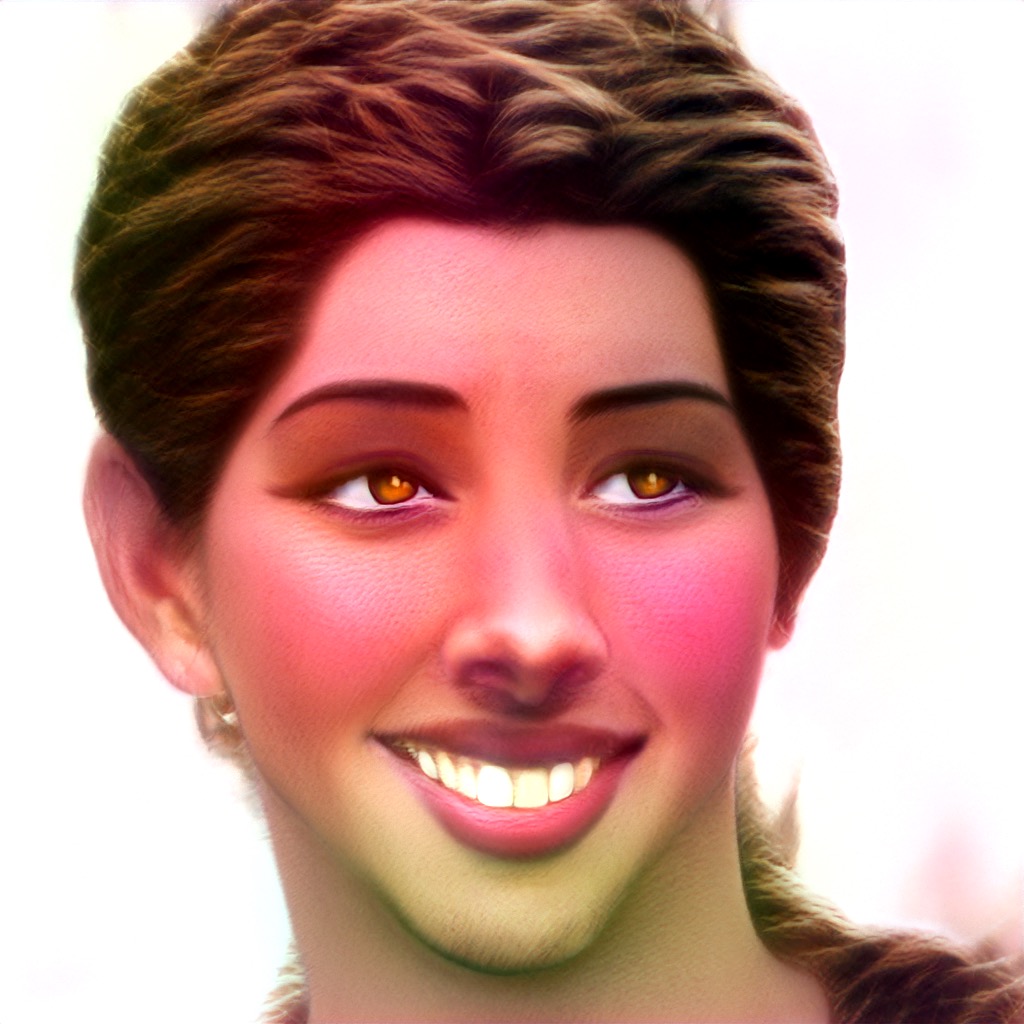}}\hfill\hspace{-5mm}
\frame{\includegraphics[trim=0 0 0 0, clip,width=.135\textwidth]{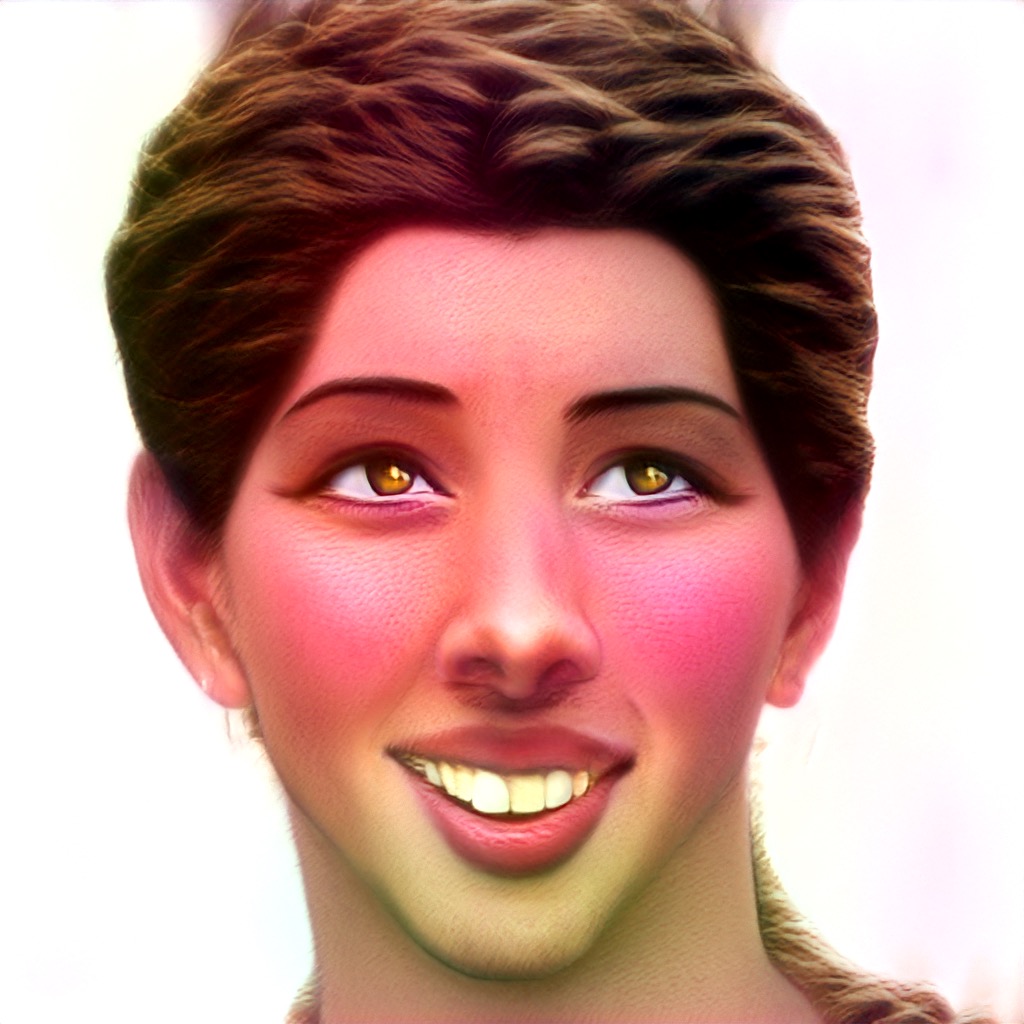}}\hfill\hspace{-5mm}
\frame{\includegraphics[trim=0 0 0 0, clip,width=.135\textwidth]{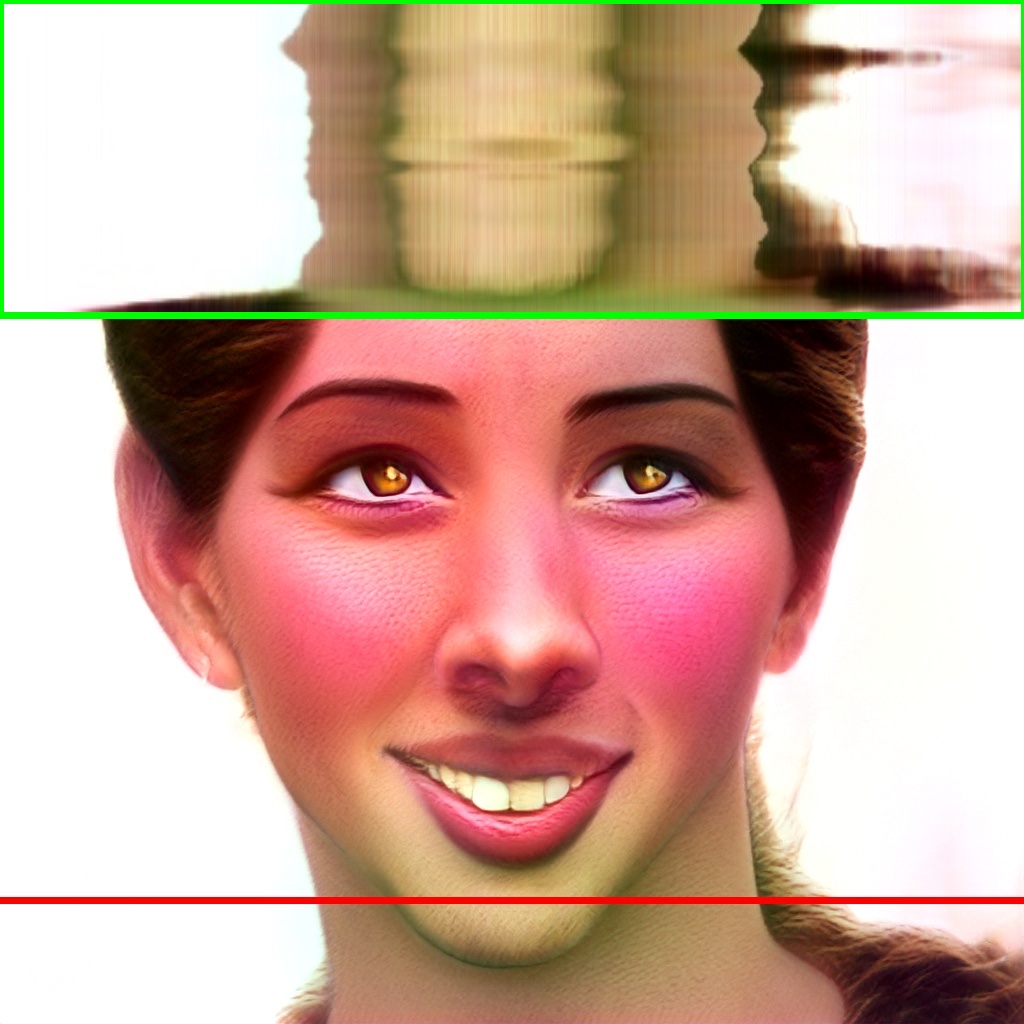}}\\

\vspace{-14mm}

\mpage{0.01}{\raisebox{60pt}{\rotatebox{90}{DVP~\cite{lei2020dvp}}}} \hfill
\frame{\includegraphics[trim=0 0 0 0, clip,width=.135\textwidth]{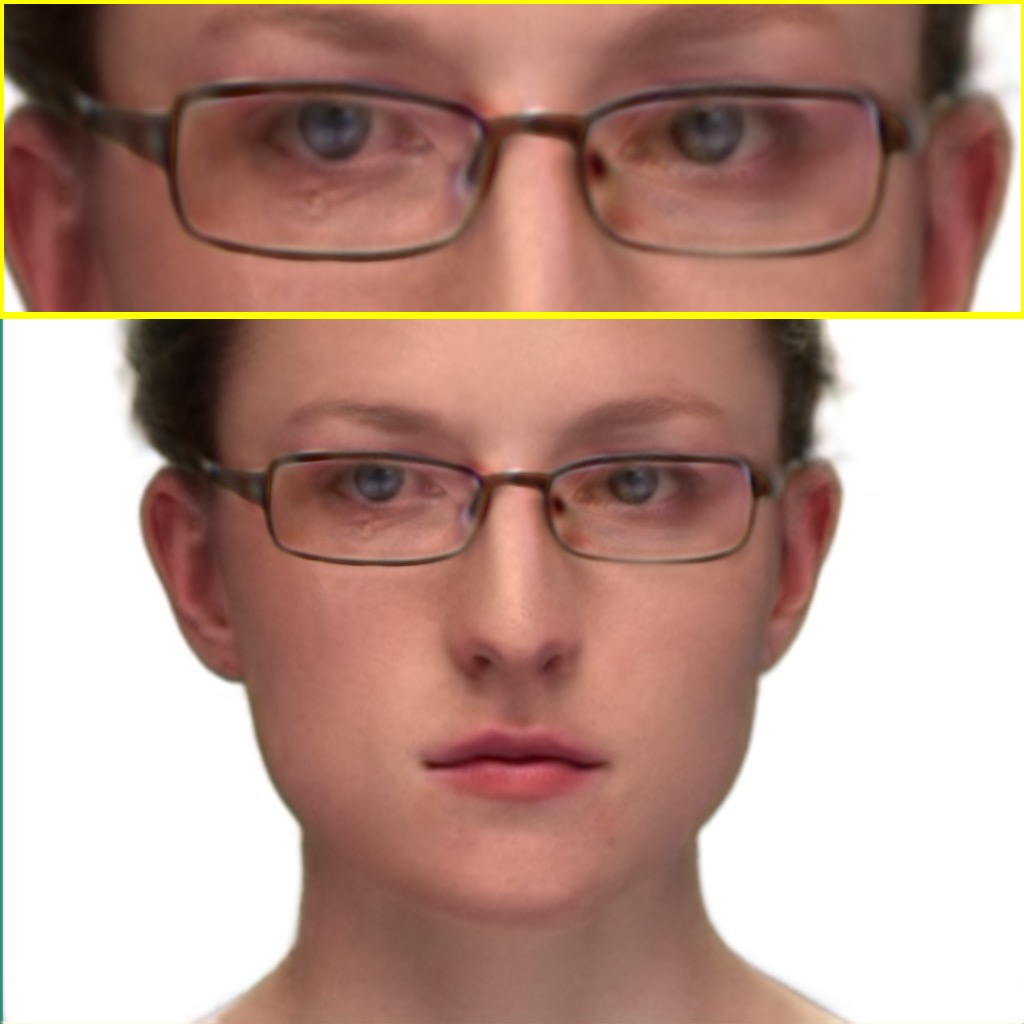}}\hfill\hspace{-5mm}
\frame{\includegraphics[trim=0 0 0 0, clip,width=.135\textwidth]{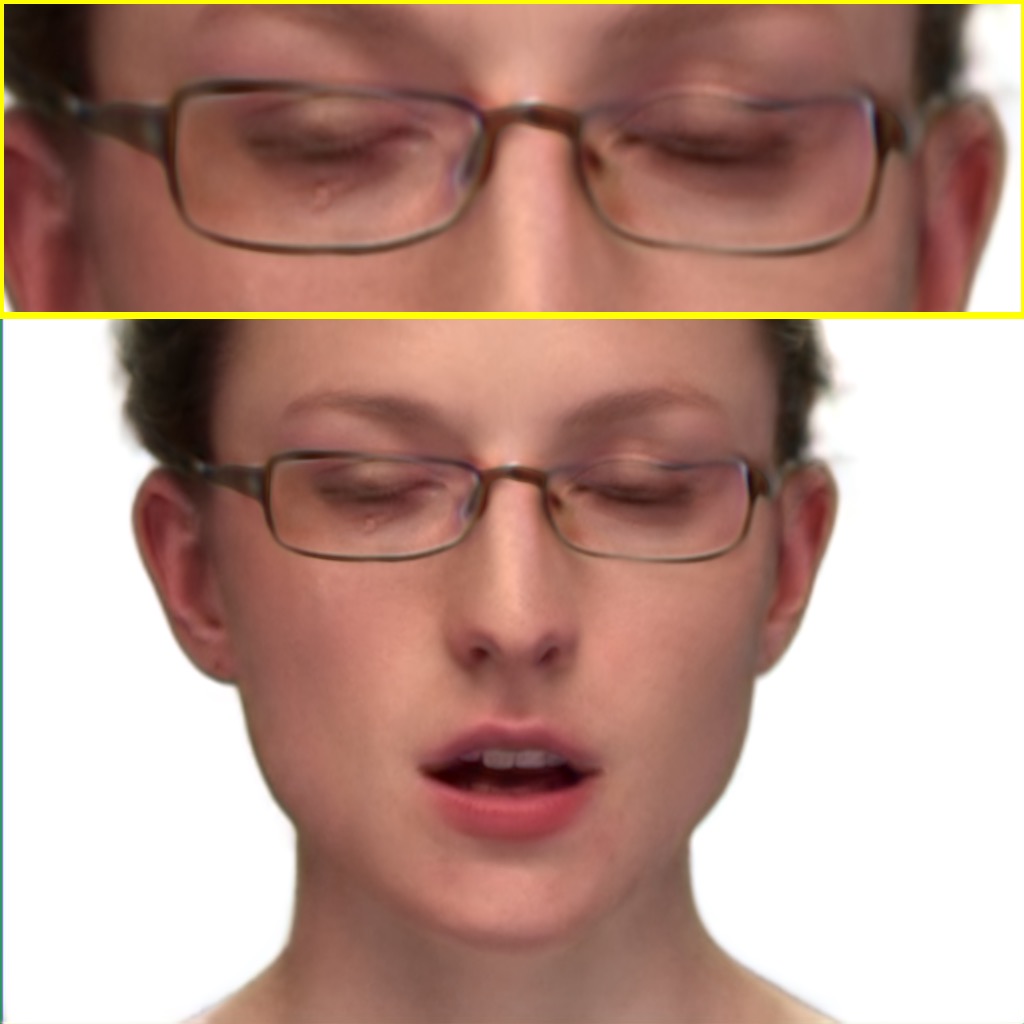}}\hfill\hspace{-5mm}
\frame{\includegraphics[trim=0 0 0 0, clip,width=.135\textwidth]{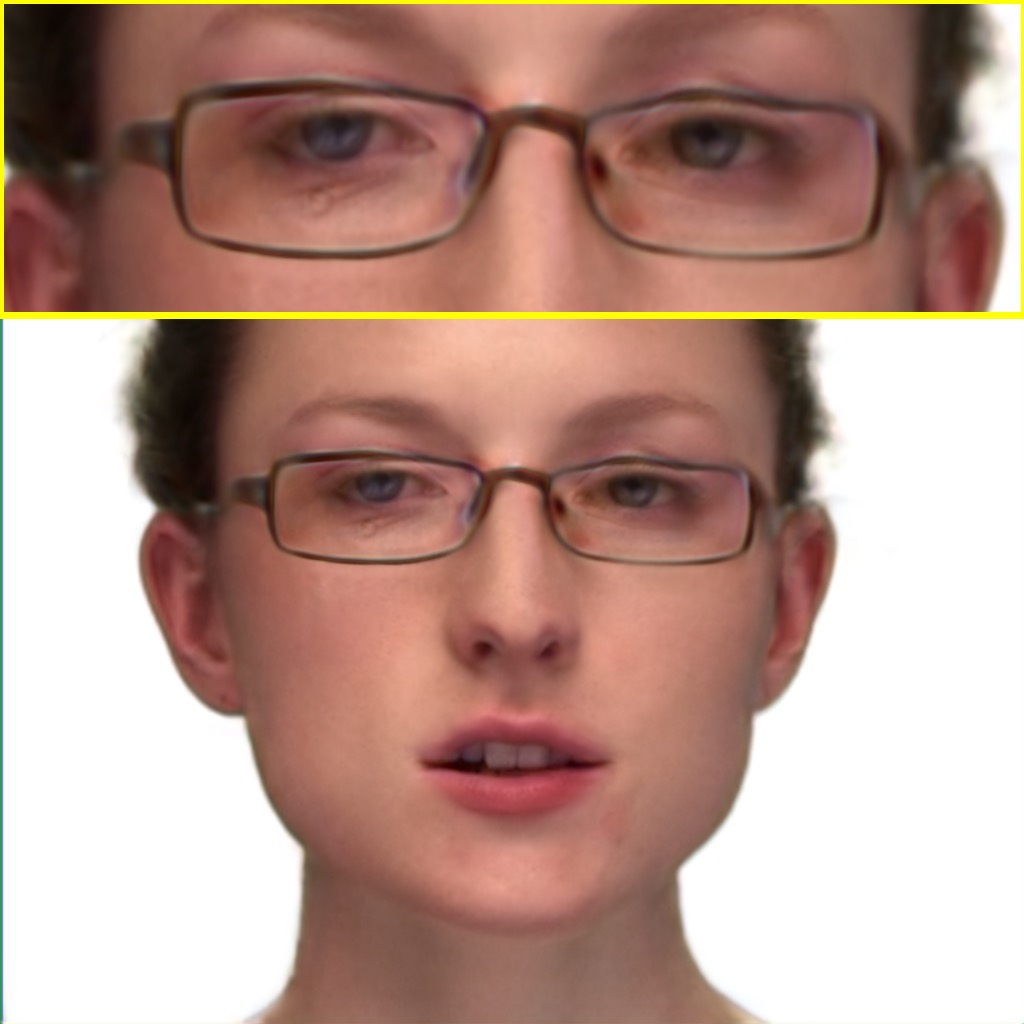}}\hfill\hspace{-5mm}
\frame{\includegraphics[trim=0 0 0 0,
clip,width=.135\textwidth]{{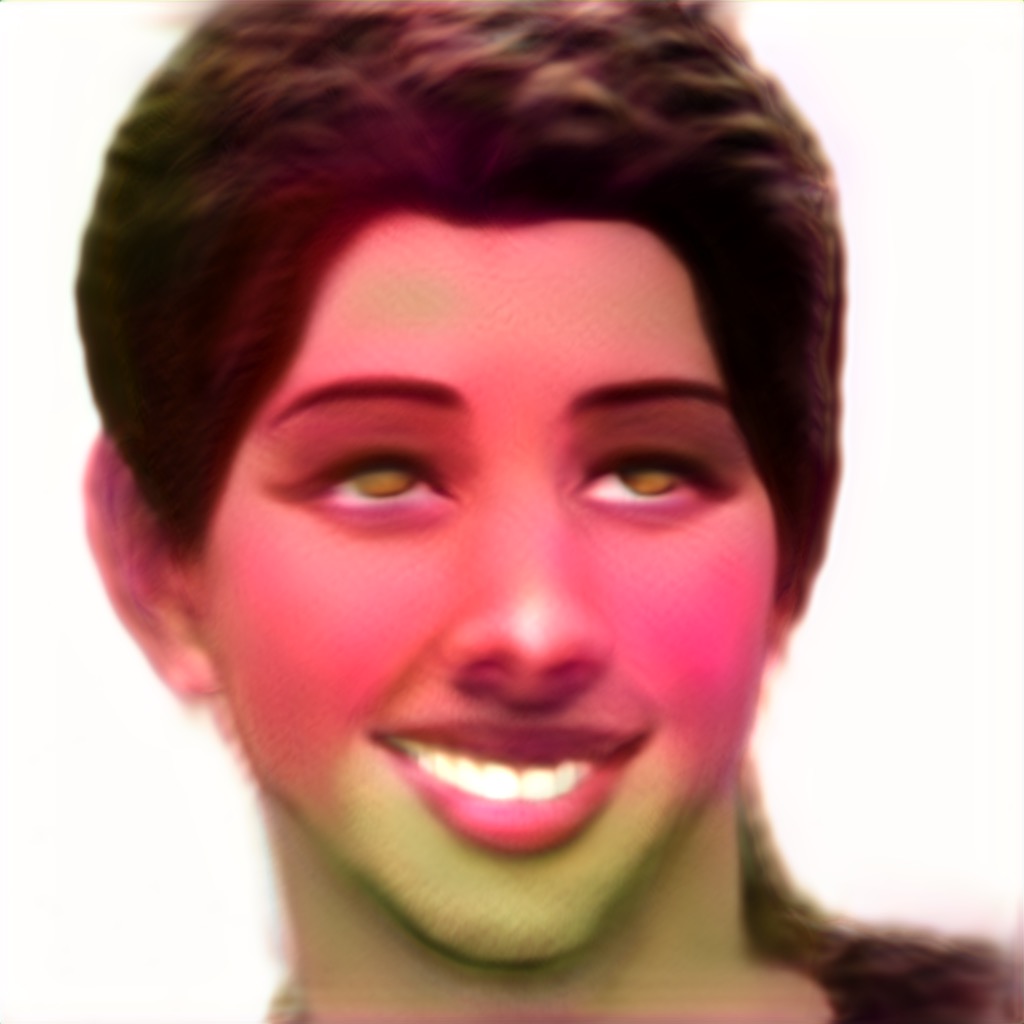}}}\hfill\hspace{-5mm}
\frame{\includegraphics[trim=0 0 0 0, clip,width=.135\textwidth]{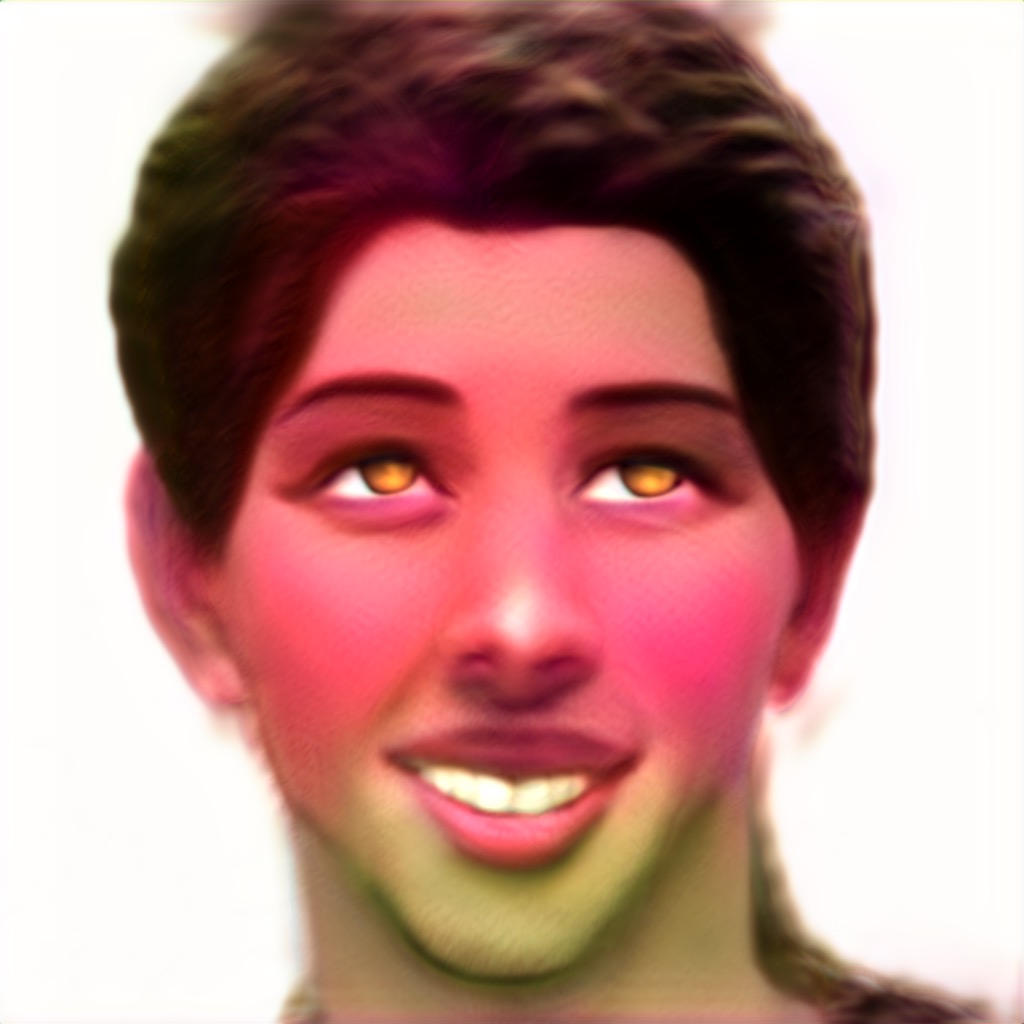}}\hfill\hspace{-5mm}
\frame{\includegraphics[trim=0 0 0 0, clip,width=.135\textwidth]{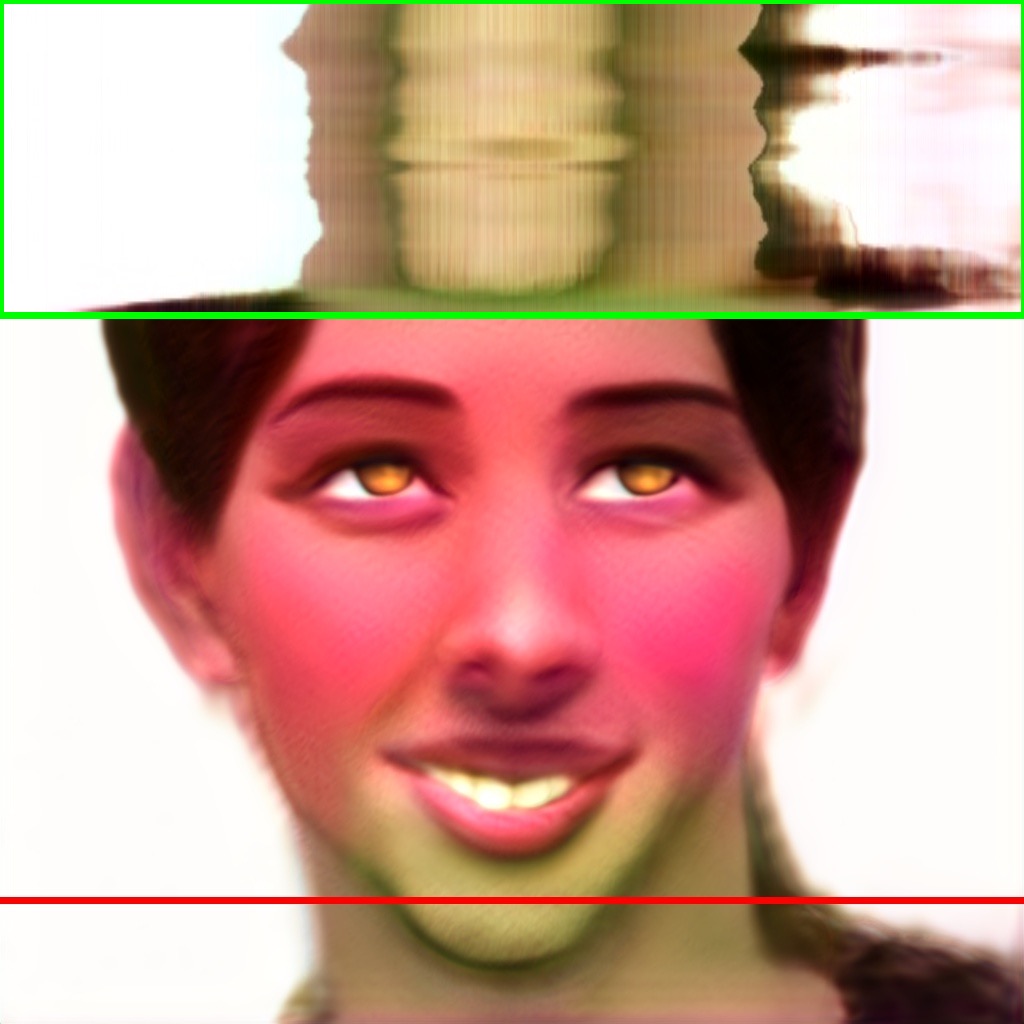}}\\

\vspace{-17mm}

\mpage{0.01}{\raisebox{60pt}{\rotatebox{90}{Ours}}} \hfill
\frame{\includegraphics[trim=0 0 0 0, clip,width=.135\textwidth]{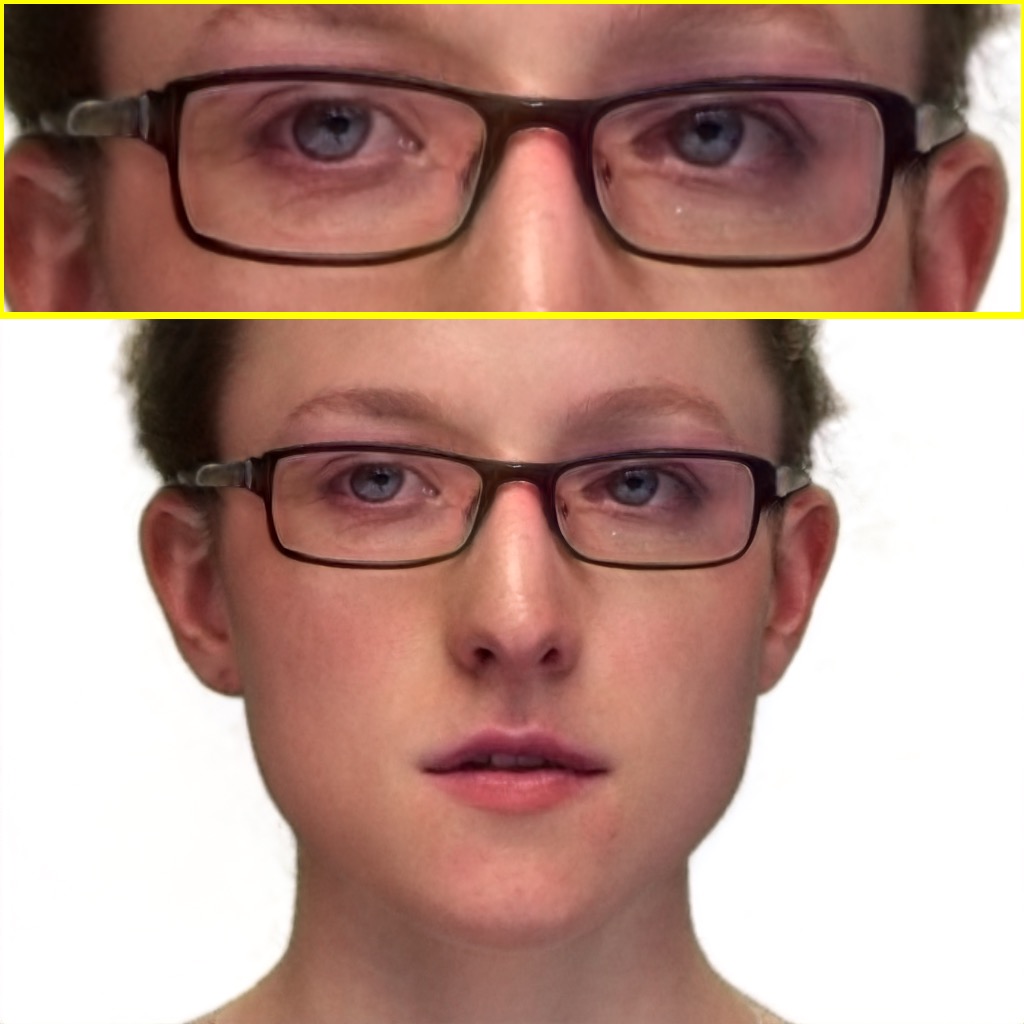}}\hfill\hspace{-5mm}
\frame{\includegraphics[trim=0 0 0 0, clip,width=.135\textwidth]{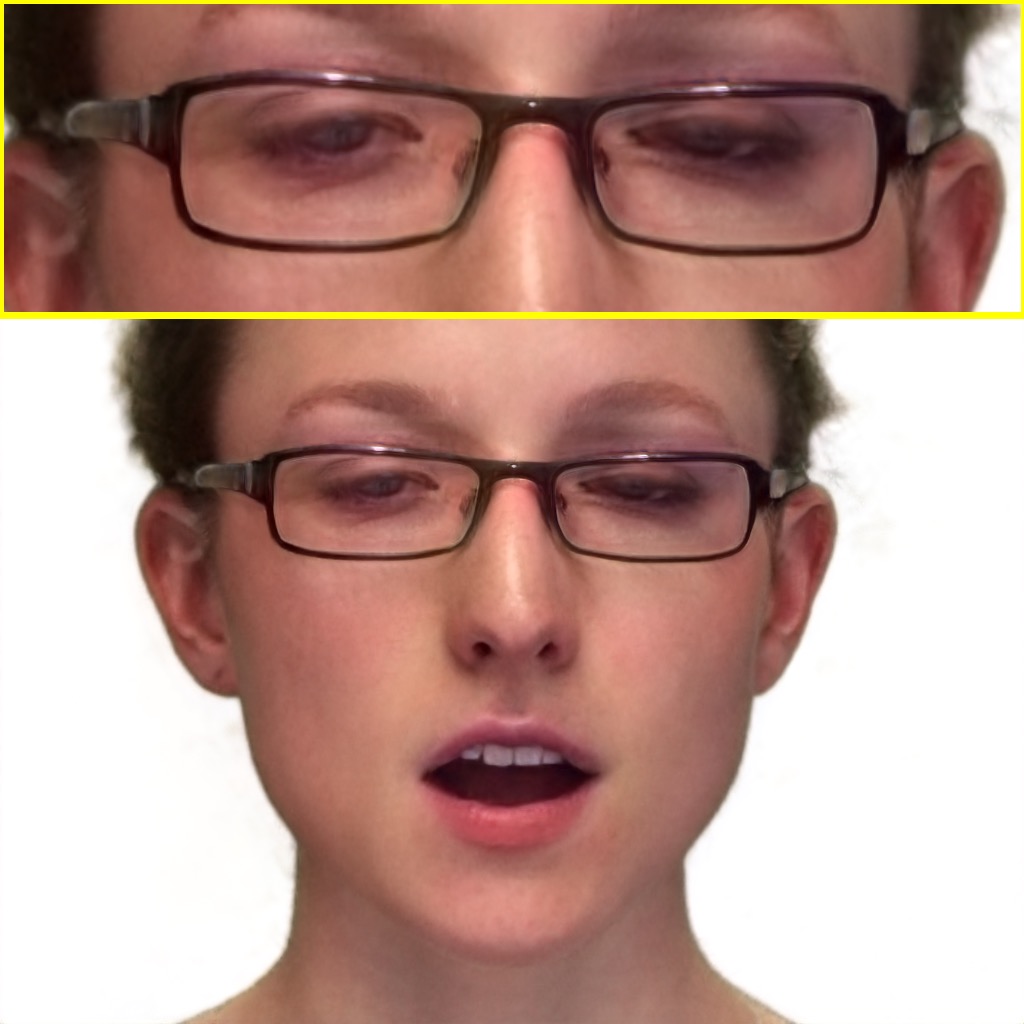}}\hfill\hspace{-5mm}
\frame{\includegraphics[trim=0 0 0 0, clip,width=.135\textwidth]{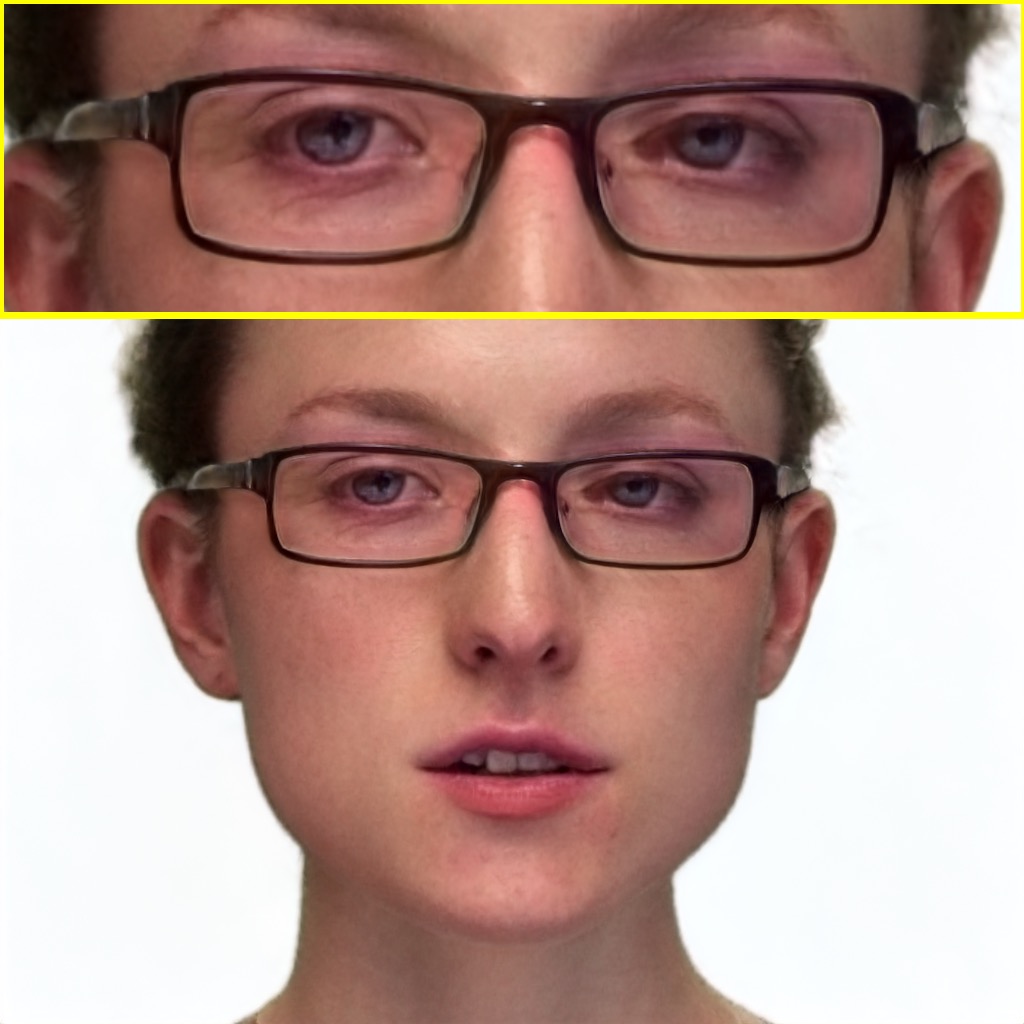}}\hfill\hspace{-5mm}
\frame{\includegraphics[trim=0 0 0 0, clip,width=.135\textwidth]{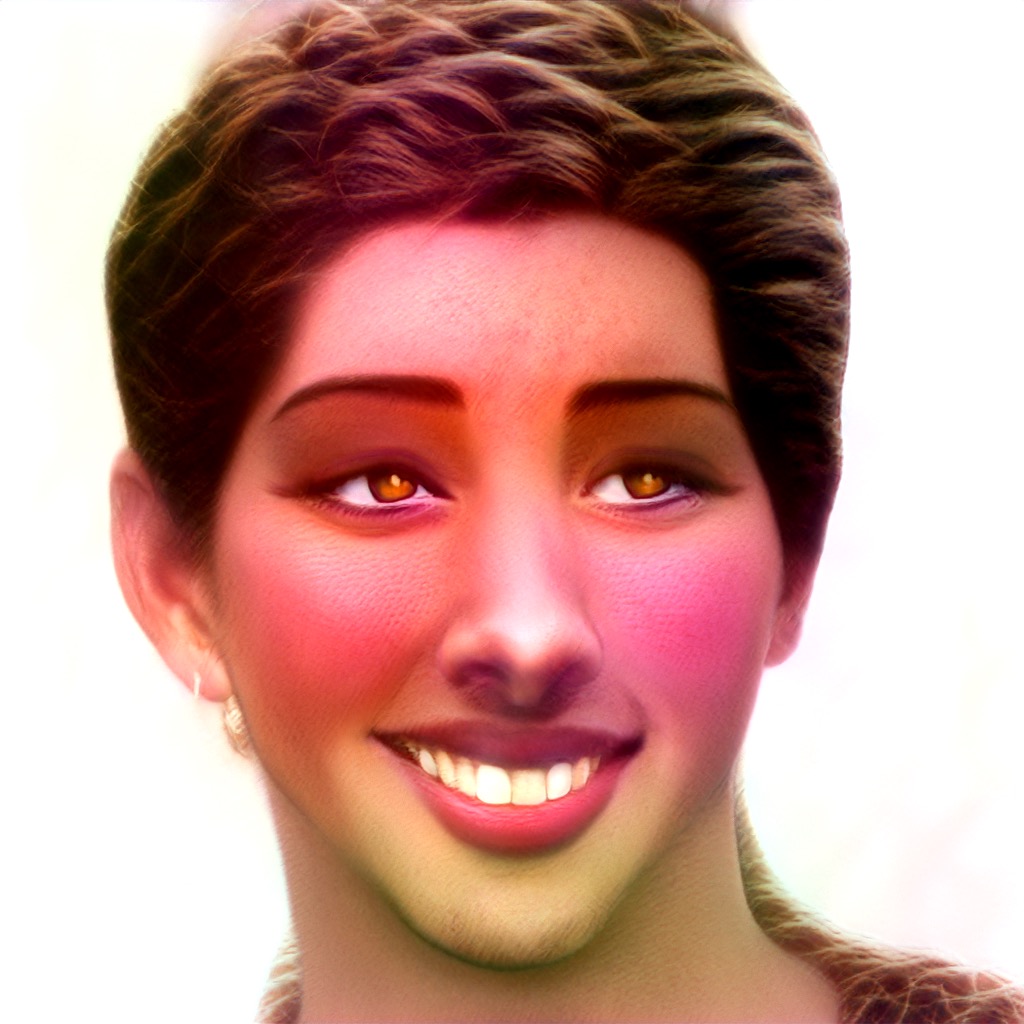}}\hfill\hspace{-5mm}
\frame{\includegraphics[trim=0 0 0 0, clip,width=.135\textwidth]{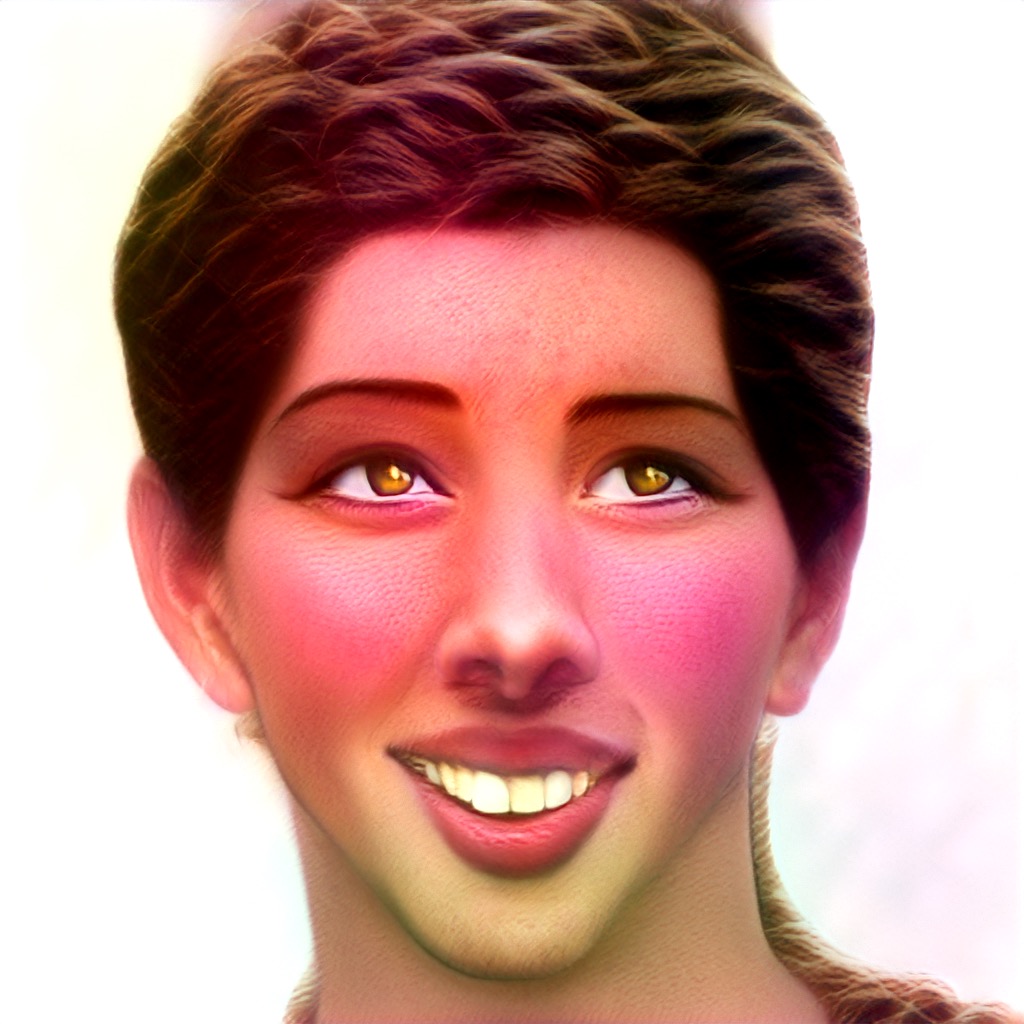}}\hfill\hspace{-5mm}
\frame{\includegraphics[trim=0 0 0 0, clip,width=.135\textwidth]{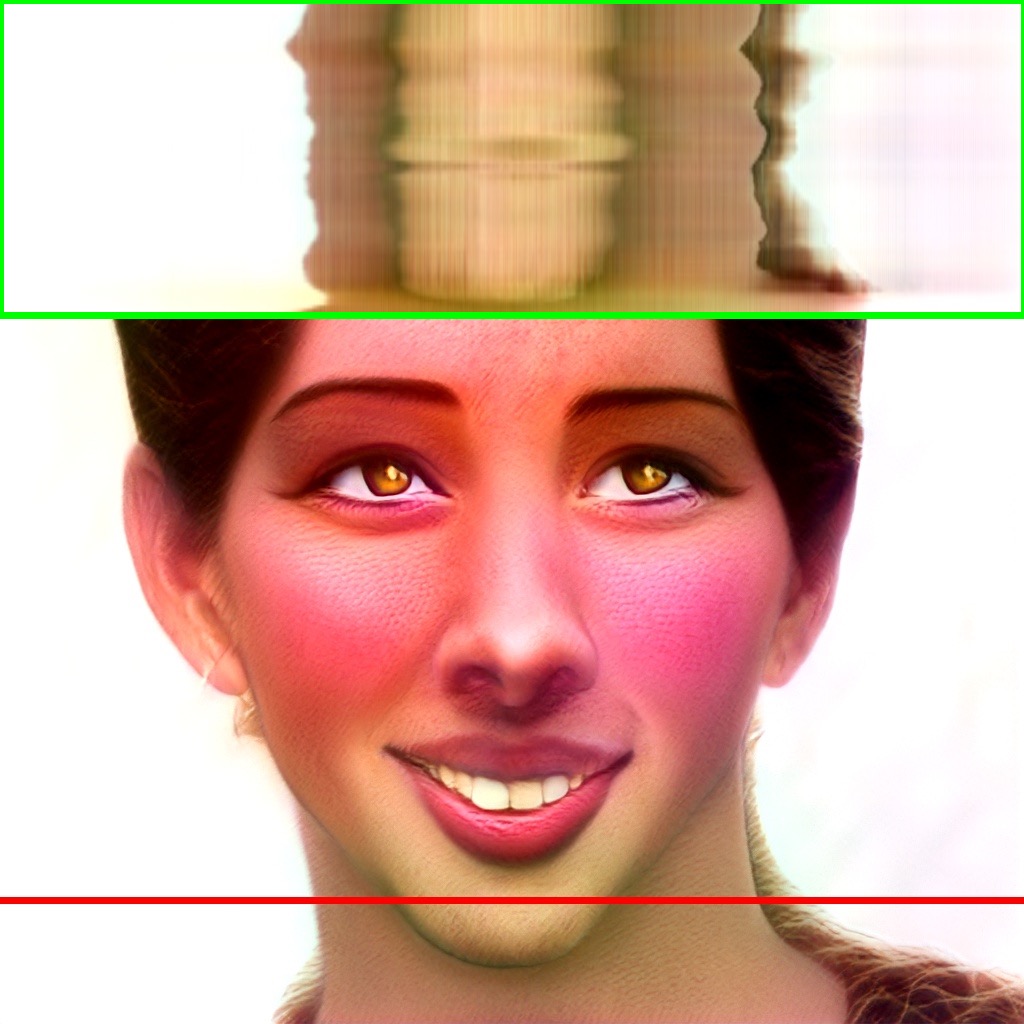}} \\

\vspace{-15mm}
\caption{
\textbf{Visual comparison with DVP~\cite{lei2020dvp}.}
DVP achieves temporal consistency by severely smoothing the image and hence losing its sharpness. Our method, however, can achieve a balance between consistency and sharpness. In ``eyeglasses'' example (left), DVP shows a different pair of eyeglasses across the time (zoom-in for better visualization), while ours remain a good consistency for the eyeglasses; in ``Disney princess'' (right), DVP shows a blurry result with an unstable x-t scanline, while ours is sharper and shows a stable consistency in the scanline.
}
\vspace{-6mm}
\label{fig:baseline_comp}
\end{figure*}
\begin{figure*}[t]
\centering
\mpage{0.01}{\raisebox{50pt}{\rotatebox{90}{Input}}} 
\frame{\includegraphics[trim=0 100 250 80, clip,width=.235\textwidth]{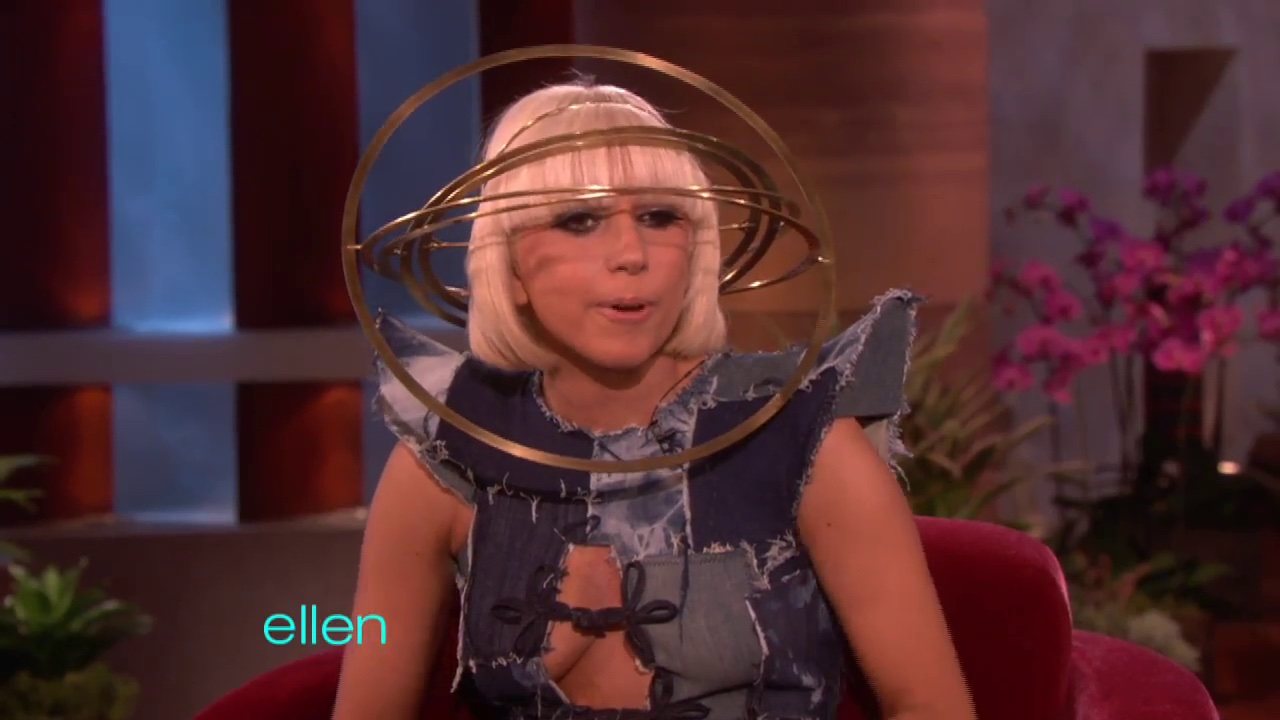}}\hfill\hspace{-5mm}
\frame{\includegraphics[trim=0 100 250 80, clip,width=.235\textwidth]{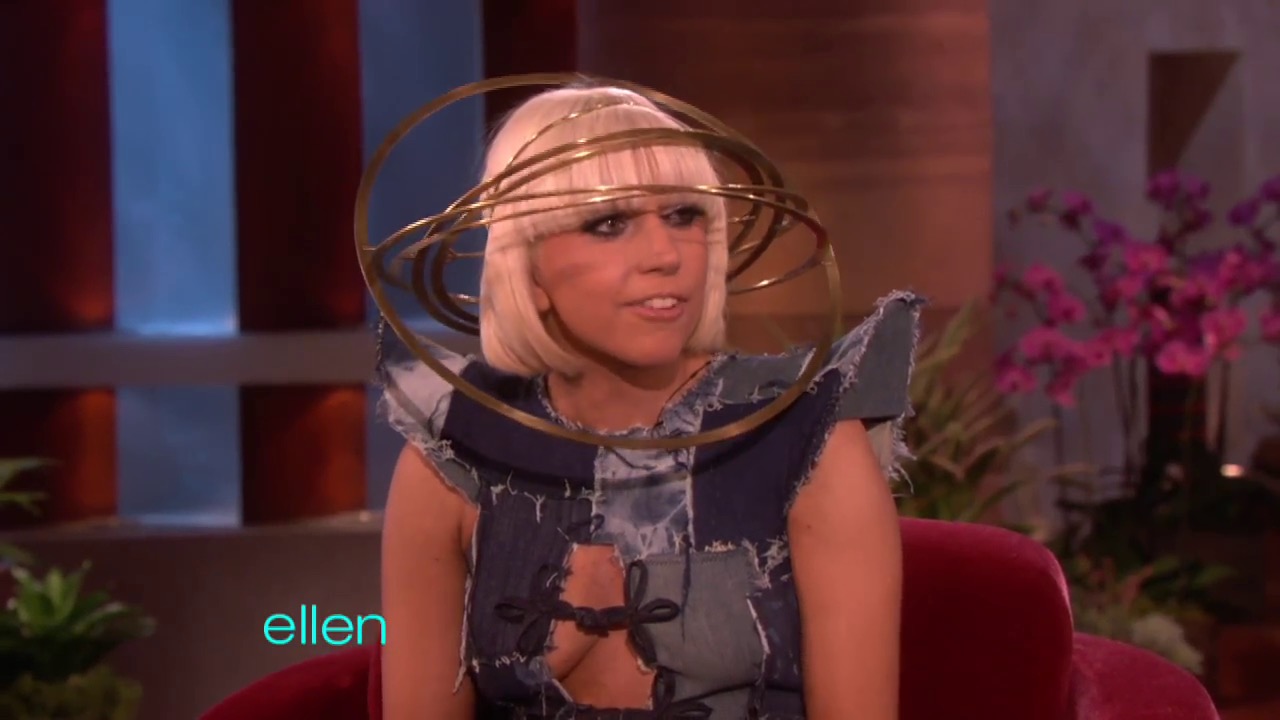}}\hfill\hspace{-5mm}
\frame{\includegraphics[trim=0 100 250 80, clip,width=.235\textwidth]{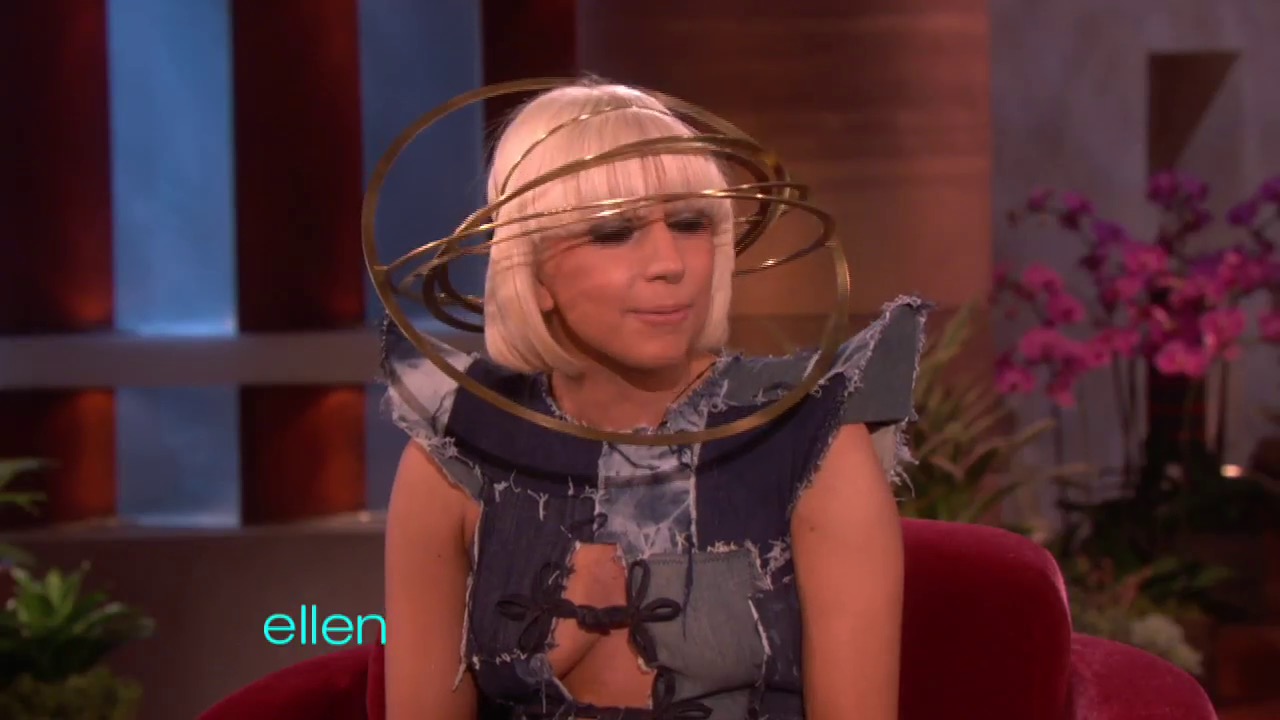}}\hfill\hspace{-5mm}
\frame{\includegraphics[trim=0 100 250 80, clip,width=.235\textwidth]{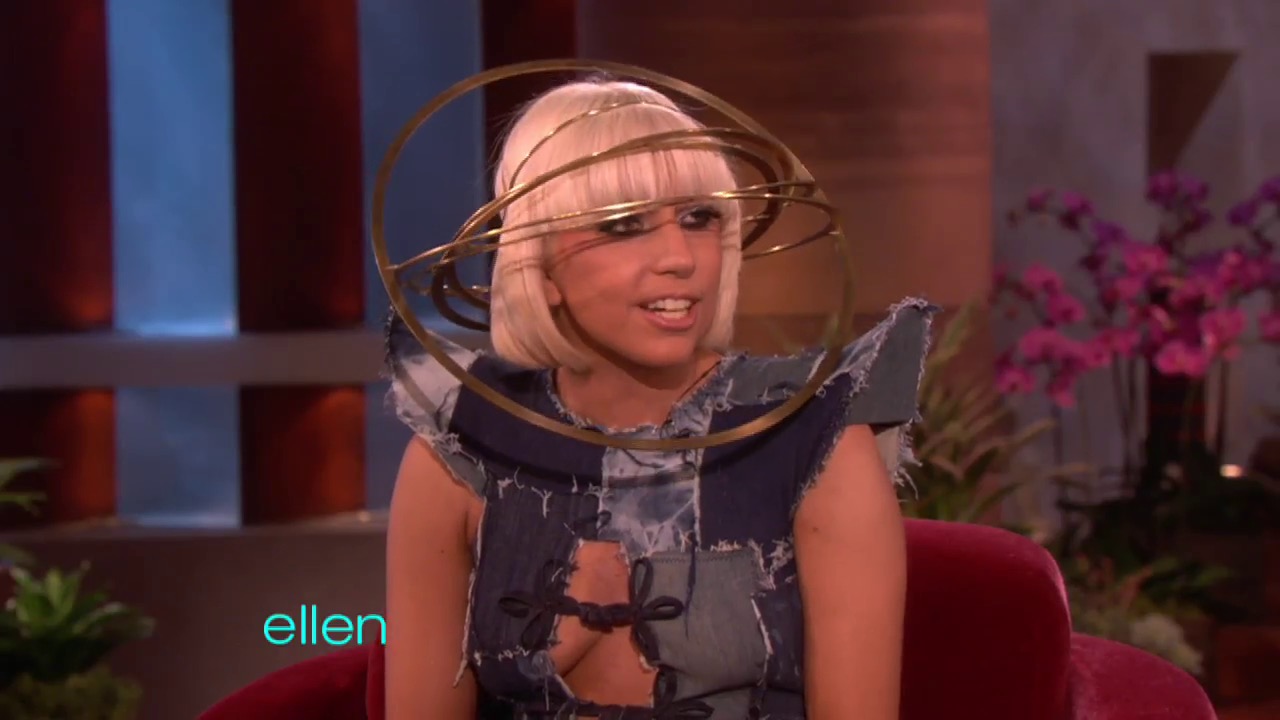}}\\

\vspace{-11mm}
\mpage{0.01}{\raisebox{50pt}{\rotatebox{90}{Ours}}} 
\frame{\includegraphics[trim=0 100 250 80, clip,width=.235\textwidth]{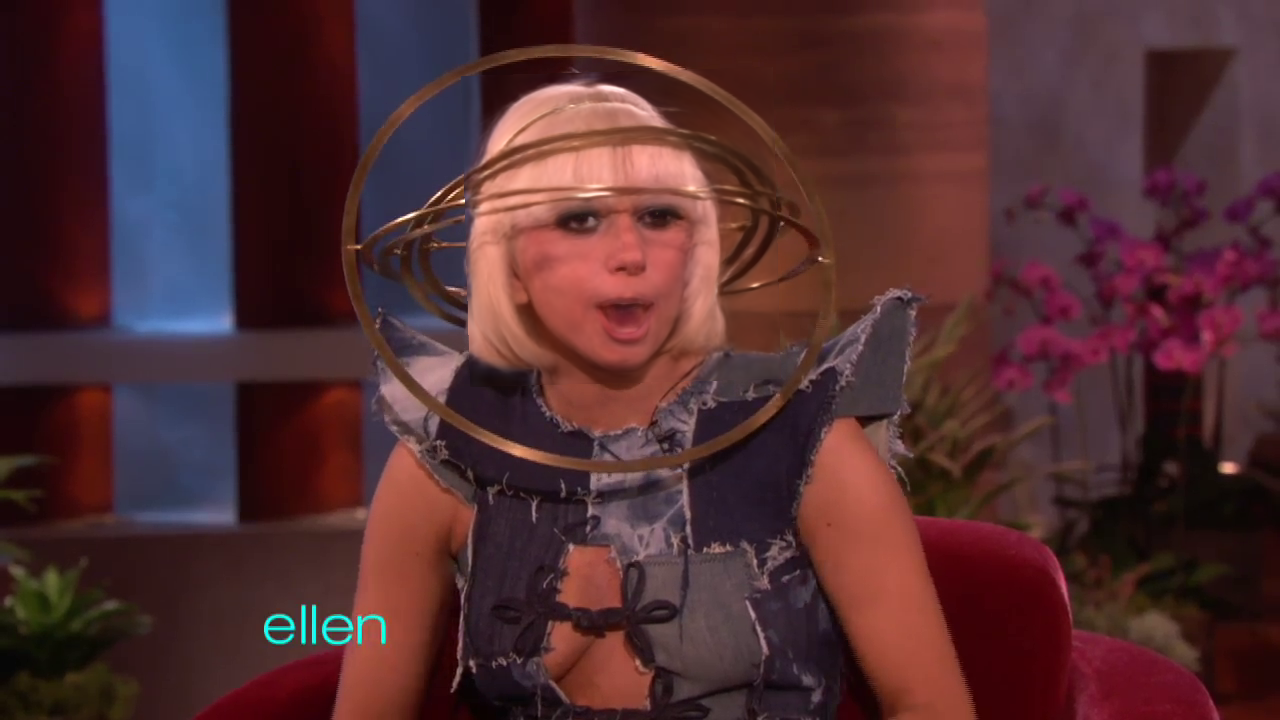}}\hfill\hspace{-5mm}
\frame{\includegraphics[trim=0 100 250 80, clip,width=.235\textwidth]{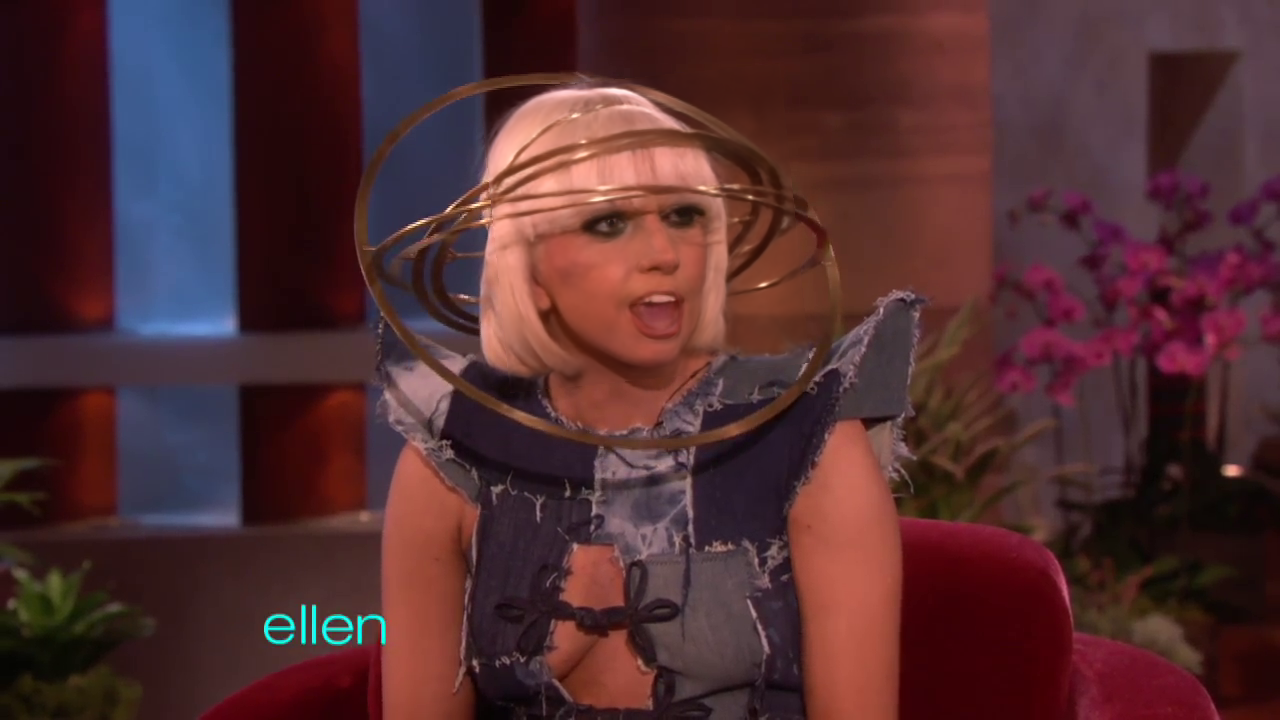}}\hfill\hspace{-5mm}
\frame{\includegraphics[trim=0 100 250 80, clip,width=.235\textwidth]{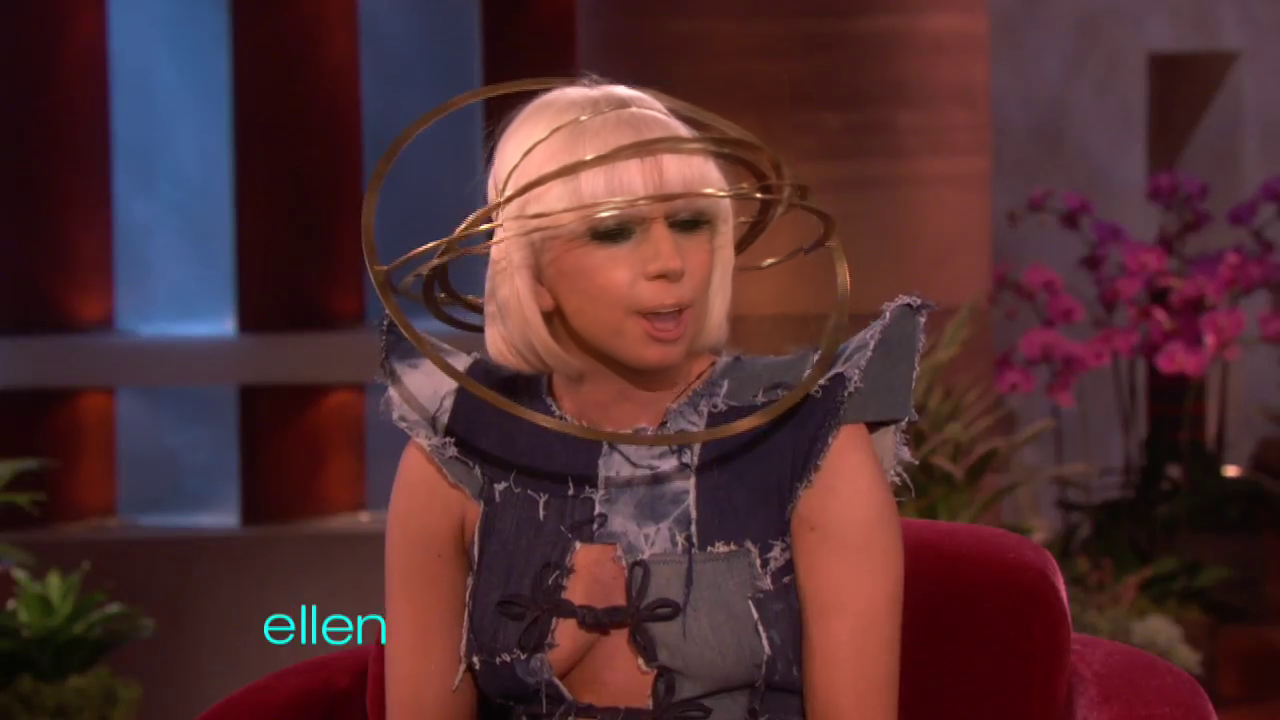}}\hfill\hspace{-5mm}
\frame{\includegraphics[trim=0 100 250 80, clip,width=.235\textwidth]{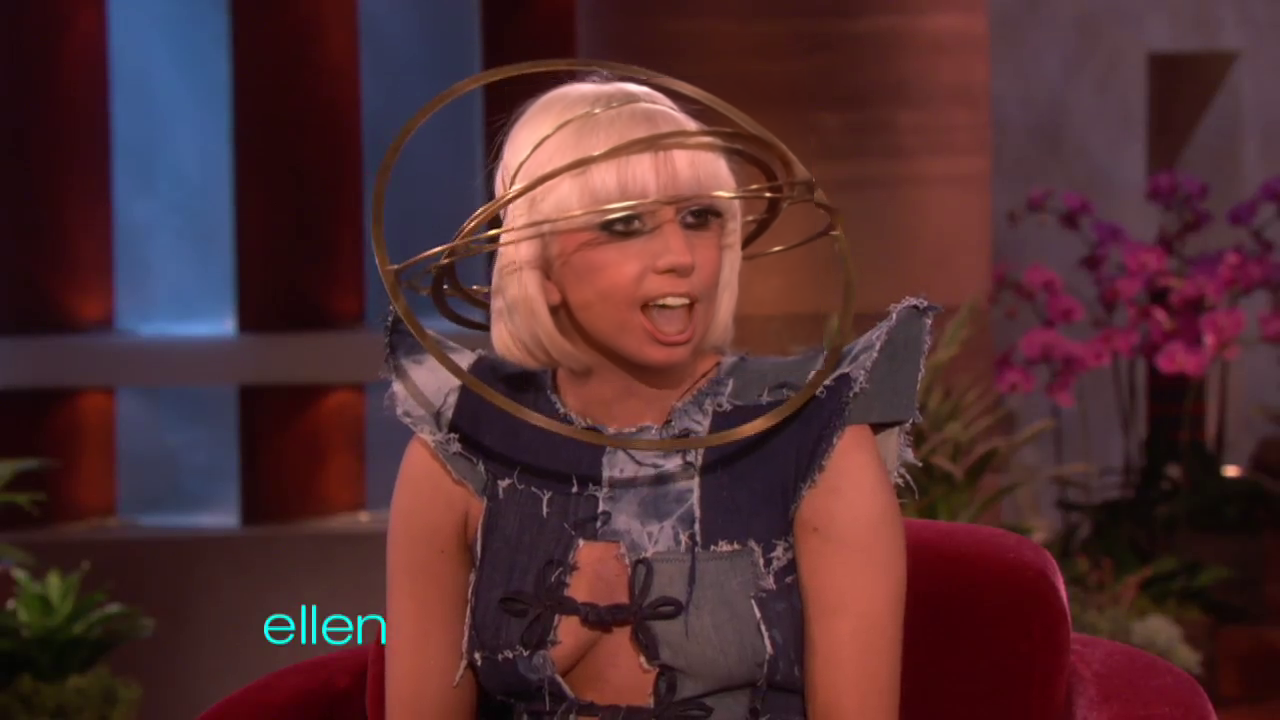}}\\

\vspace{-10mm}

\mpage{0.01}{\raisebox{50pt}{\rotatebox{90}{Input}}} 
\frame{\includegraphics[trim=0 100 250 80, clip,width=.235\textwidth]{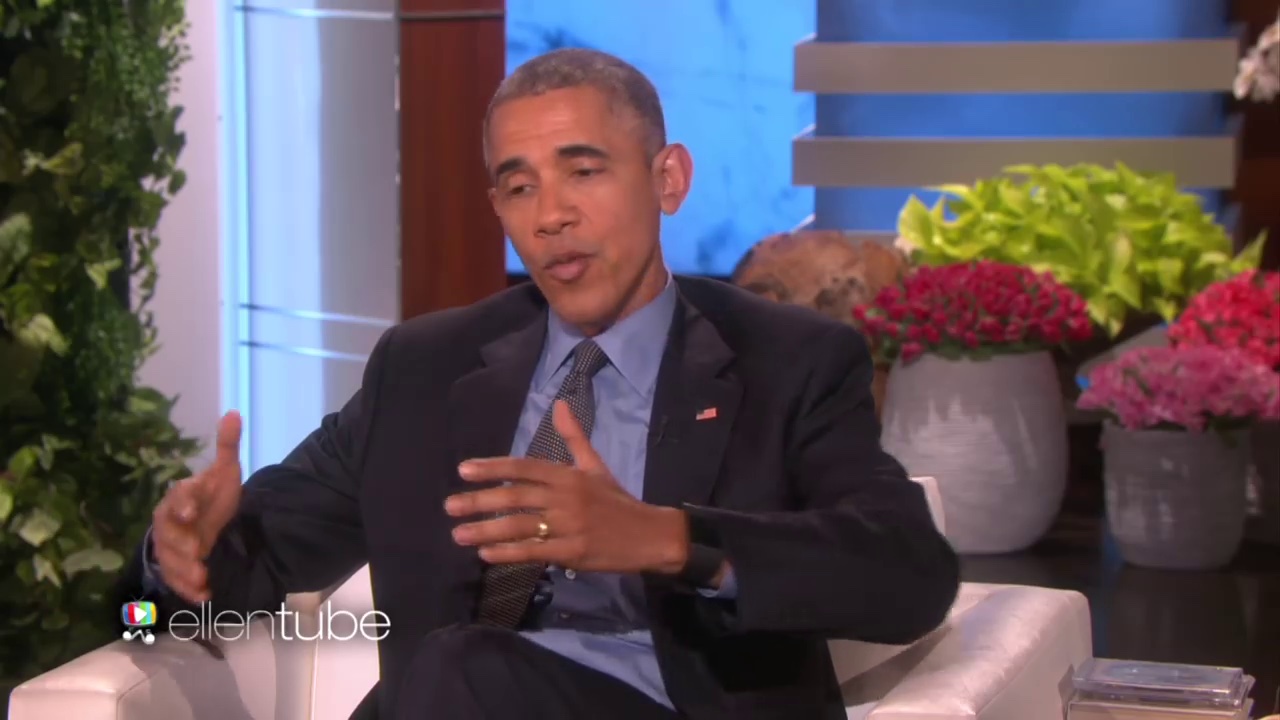}}\hfill\hspace{-5mm}
\frame{\includegraphics[trim=0 100 250 80, clip,width=.235\textwidth]{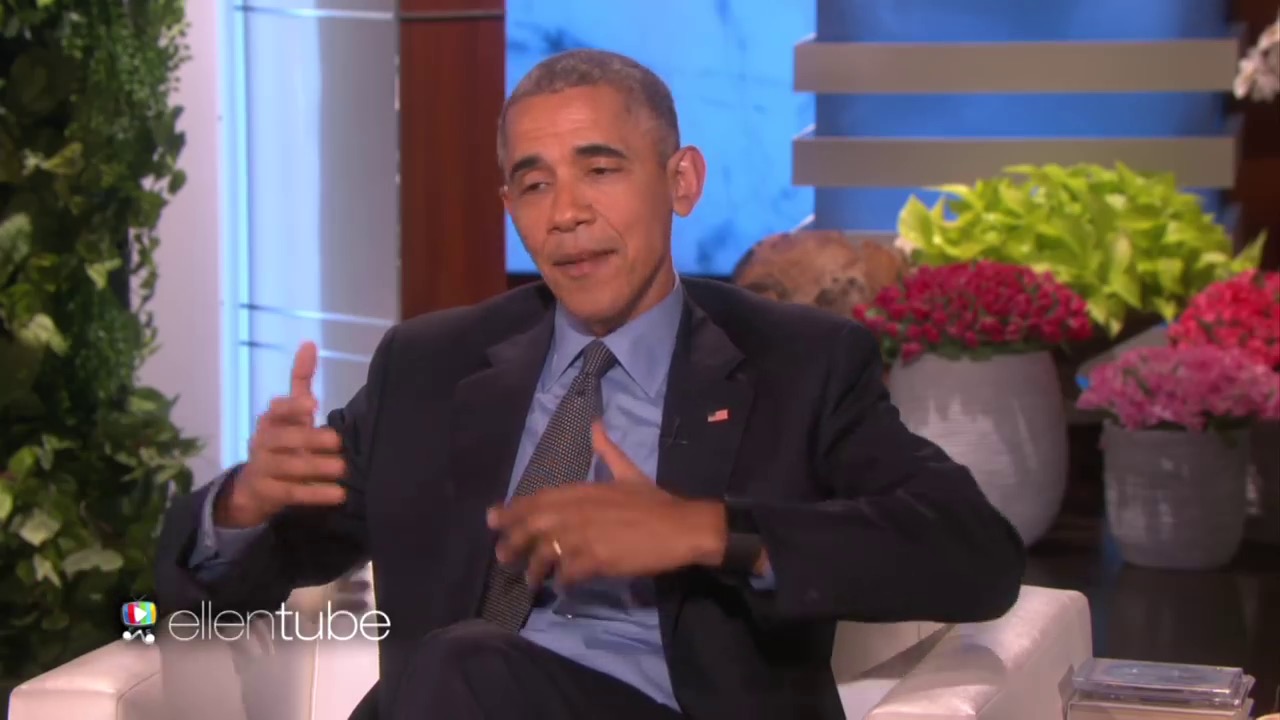}}\hfill\hspace{-5mm}
\frame{\includegraphics[trim=0 100 250 80, clip,width=.235\textwidth]{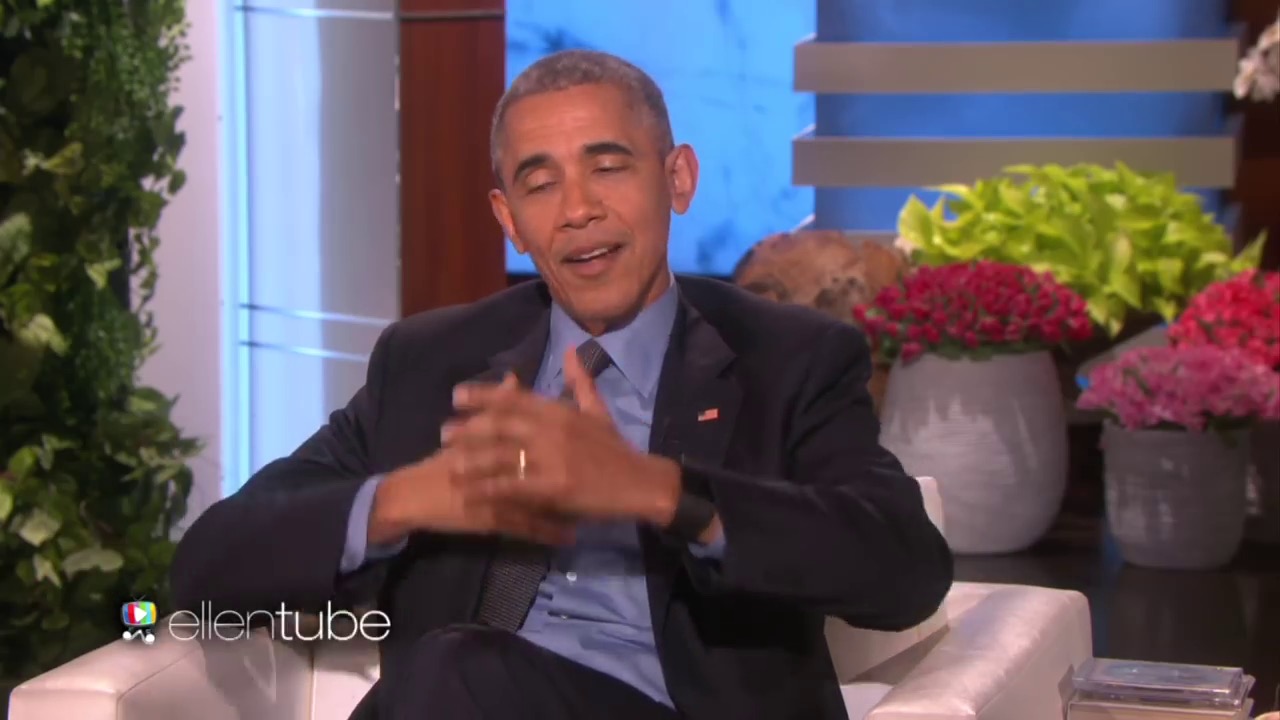}}\hfill\hspace{-5mm}
\frame{\includegraphics[trim=0 100 250 80, clip,width=.235\textwidth]{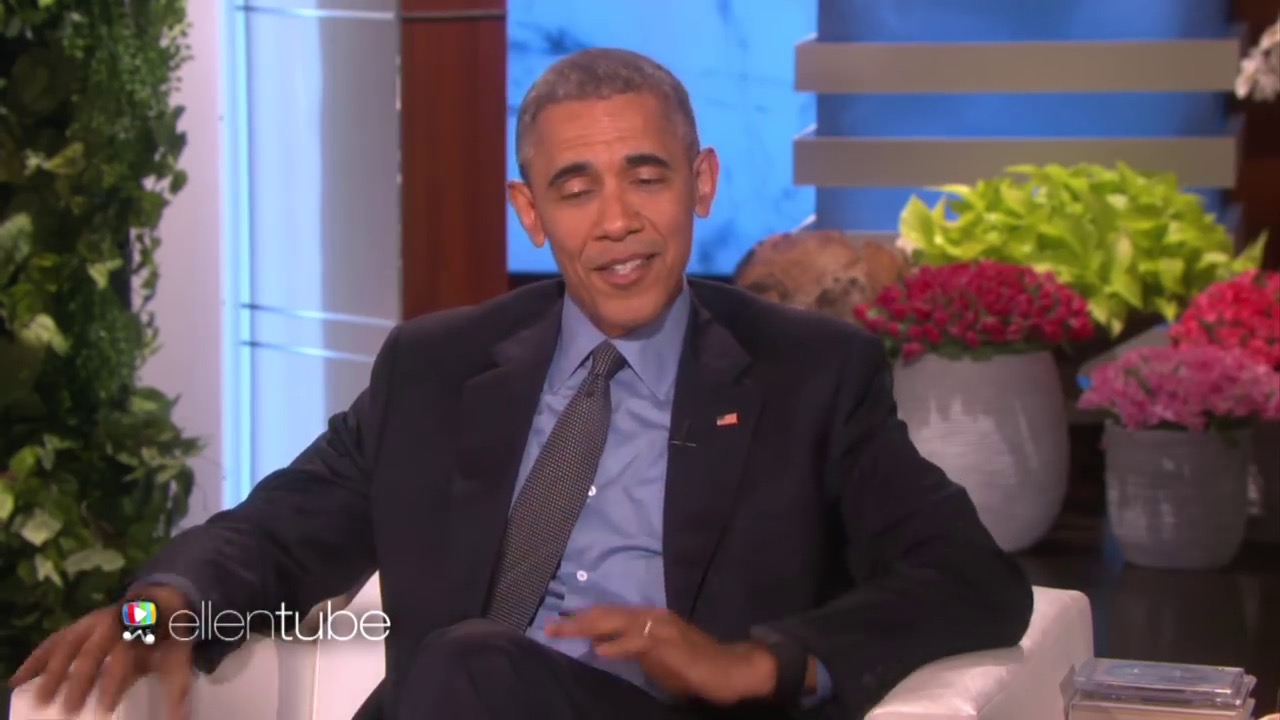}}\\

\vspace{-11mm}
\mpage{0.01}{\raisebox{50pt}{\rotatebox{90}{Ours}}} 
\frame{\includegraphics[trim=0 100 250 80, clip,width=.235\textwidth]{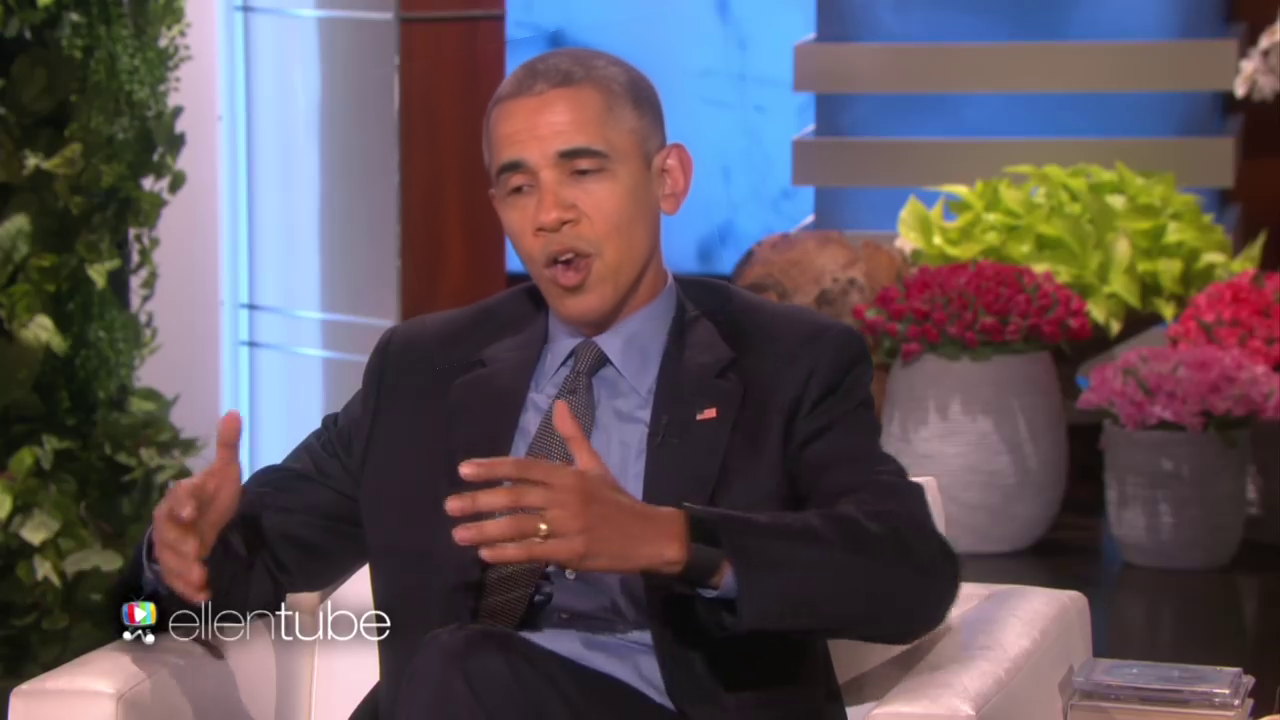}}\hfill\hspace{-5mm}
\frame{\includegraphics[trim=0 100 250 80, clip,width=.235\textwidth]{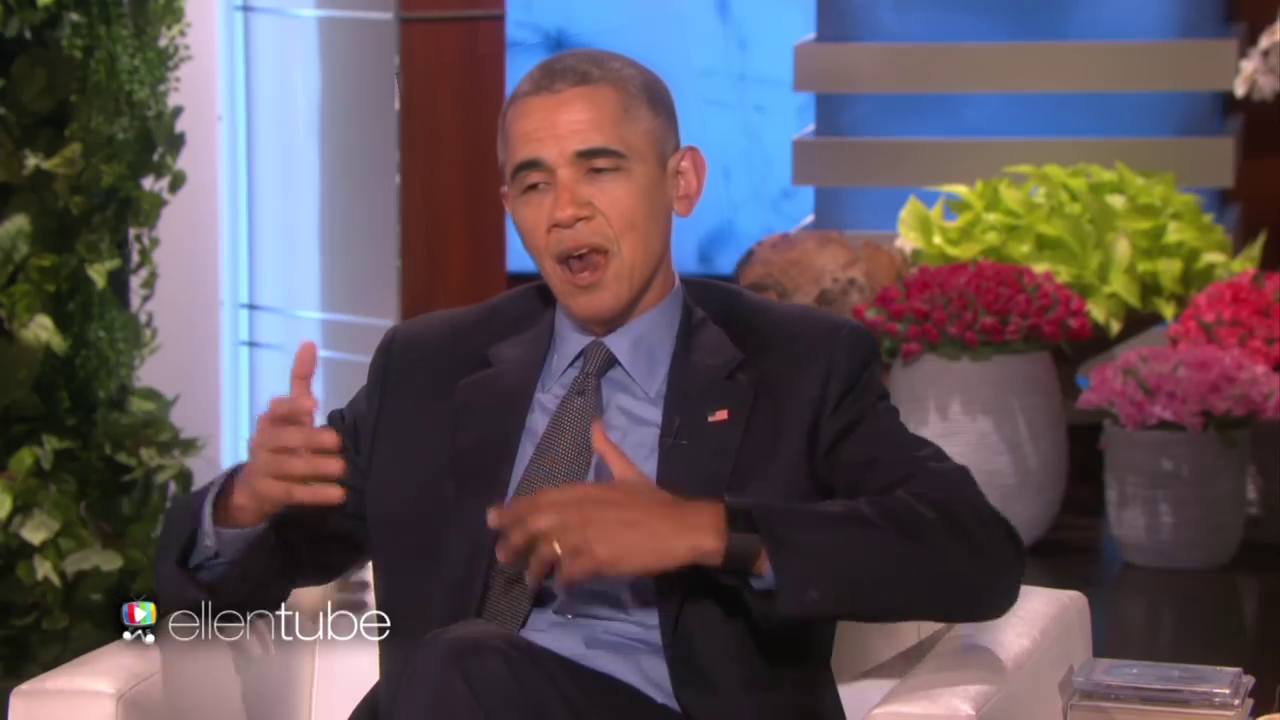}}\hfill\hspace{-5mm}
\frame{\includegraphics[trim=0 100 250 80, clip,width=.235\textwidth]{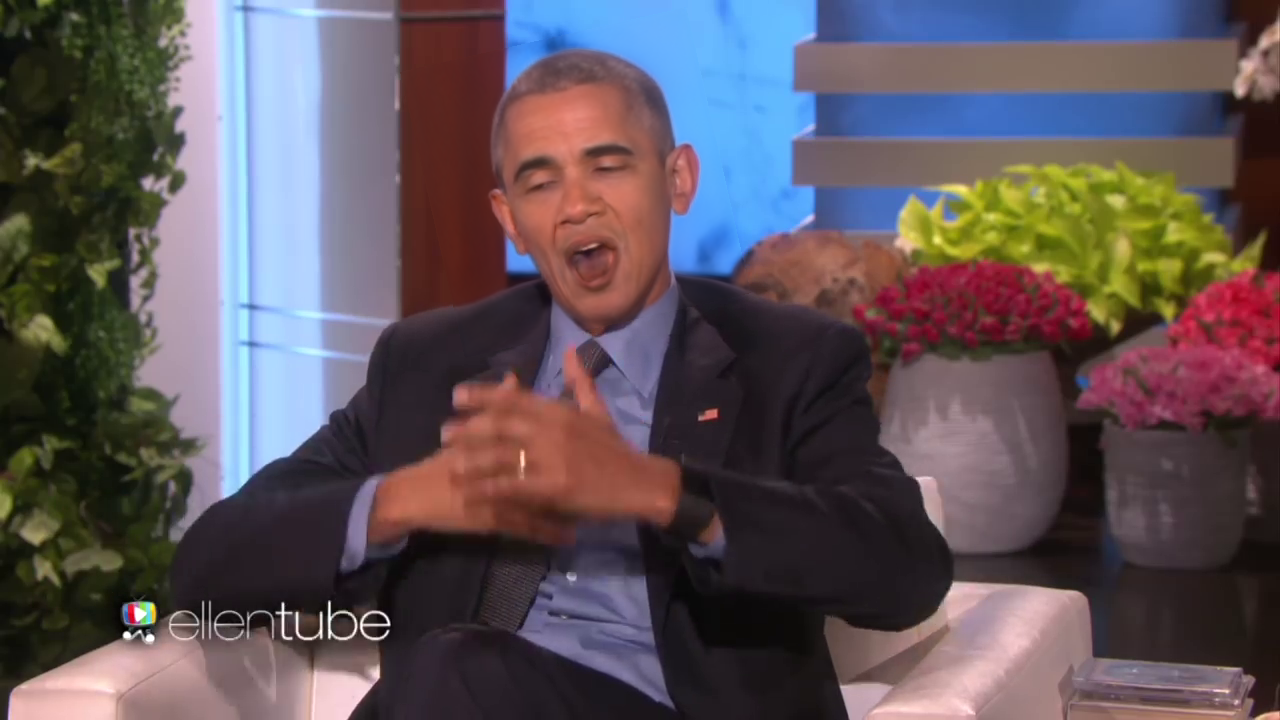}}\hfill\hspace{-5mm}
\frame{\includegraphics[trim=0 100 250 80, clip,width=.235\textwidth]{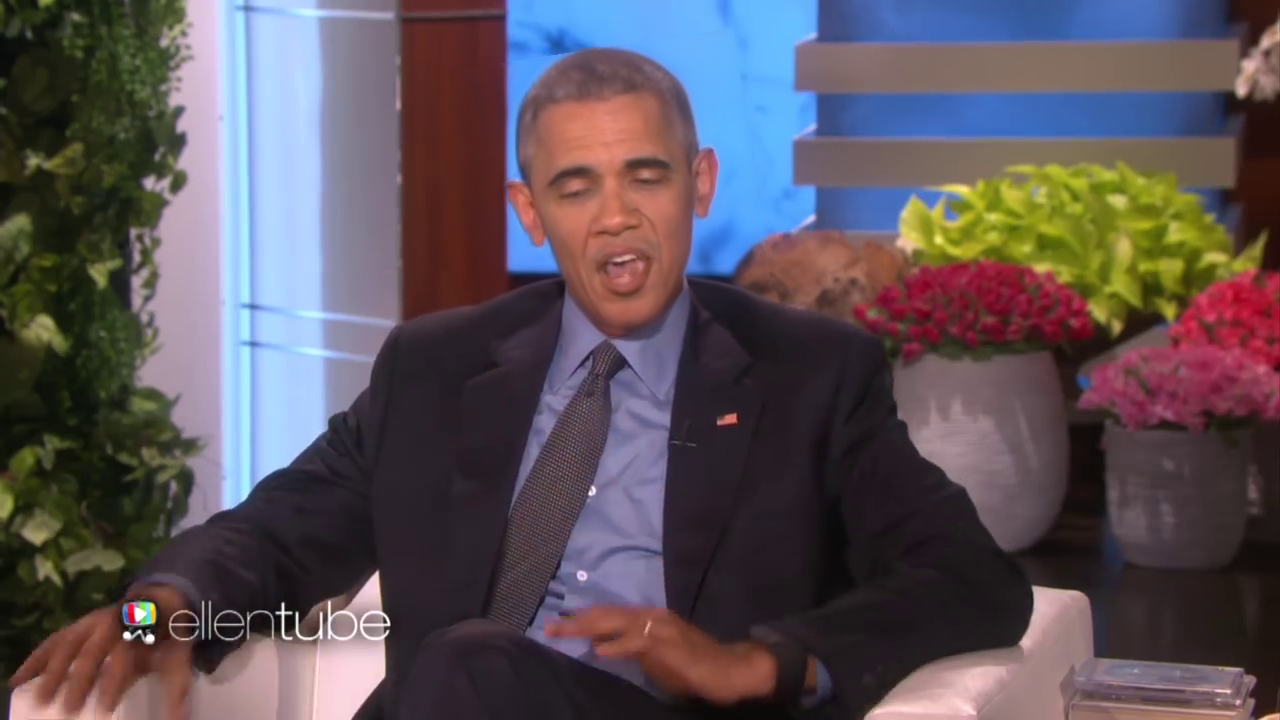}}\\

\vspace{-10mm}

\caption{{\textbf{Results on Internet videos}.} 
Results on the Internet videos.
We change the first person to ``surprised'' expression, and change the second person to ``angry''.
}
\vspace{-3mm}
\label{fig:diverse_qual_results}
\end{figure*}


\topic{Implementation details.}
We use StyleGAN-ADA~\cite{Karras2020ada} as our pre-trained generator.
We experiment with in-domain and out-of-domain editing techniques to validate our approach for different GAN inversion 
methods. 
Specifically, for in-domain editing, we use the PTI inversion~\cite{roich2021pivotal} (based on e4e~\cite{tov2021designing}) and StyleCLIP mapper~\cite{Patashnik_2021_ICCV}. 
For out-of-domain editing, we use the Restyle encoder~\cite{alaluf2021restyle} and the StyleGAN-NADA~\cite{gal2021stylegannada}. 
In the following, we show sample results from the video frames. 
We encourage the readers to view the videos in the supplementary material for video results.

\topic{Datasets.}
We conduct our metric evaluation using 20 videos randomly sampled from RAVDESS dataset~\cite{livingstone2018ryerson}. We conduct 5 types of in-domain editing for each video and 5 types of out-of-domain editing.
To further demonstrate the capabilities of our method to handle \emph{real} videos, we also apply our approach to Internet videos and show the visual results.

\topic{Metrics.} 
We aim to evaluate the method using two main aspects: 
1) temporal consistency and 
2) perceptual similarity with the semantically edited frames.
To evaluate temporal consistency, we measure the \emph{Warping Error} $E_{warp}$:
\begin{equation}\label{eq:metrics_Ewarp}
    E_{warp}(I_t, I_{t+1}) = \frac{1}{\sum_{i=1}^{N}M_t(p_i)} \cdot \sum_{i=1}^{N}M_t(p_i) ||I_t(p_i) - \hat{I}_{t+1}(p_i)||_2^2 \,, 
\end{equation}
where $\hat{I}_{t+1} = warp(I_{t+1}, F_{t \rightarrow t+1})$, $N$ is the number of pixels, $p_i$ is the $i$-th pixel, $M_t$ is a binary non-occlusion mask, which shows non-occluded pixels, we compute it using the forward-backward consistency error the threshold in~\cite{Lai-ECCV-2018,liu2020learning}.

We also measure the LPIPS perceptual similarity score~\cite{zhang2018unreasonable} (with AlexNet~\cite{krizhevsky2012imagenet}) between the directly edited video $V^{edit} = \{I^{edit}_{1}, I^{edit}_{2}, \cdots, I^{edit}_{T}\}$ and 
the output of our phase 2 $\{\hat{I}^{''}_{1}, \hat{I}^{''}_{2}, \cdots, \hat{I}^{''}_{T}\}$
by measuring the averaged perceptual similarity 
between the corresponding frames.

The purpose of these two metrics is to evaluate whether the method can achieve a balance between \emph{temporal consistency} and \emph{fidelity degradation}. 
This is an inherent trade-off.
Preserving all the details of per-frame editing inevitably leads to temporal flickering artifacts.
Focusing only on temporal consistency may easily lead to blurry videos.
Our goal is that the final output video is visually similar to the directly (per-frame) edited video. 

\begin{figure*}[t]
\centering

\frame{\includegraphics[trim=250 50 250 95, clip,width=.24\textwidth]{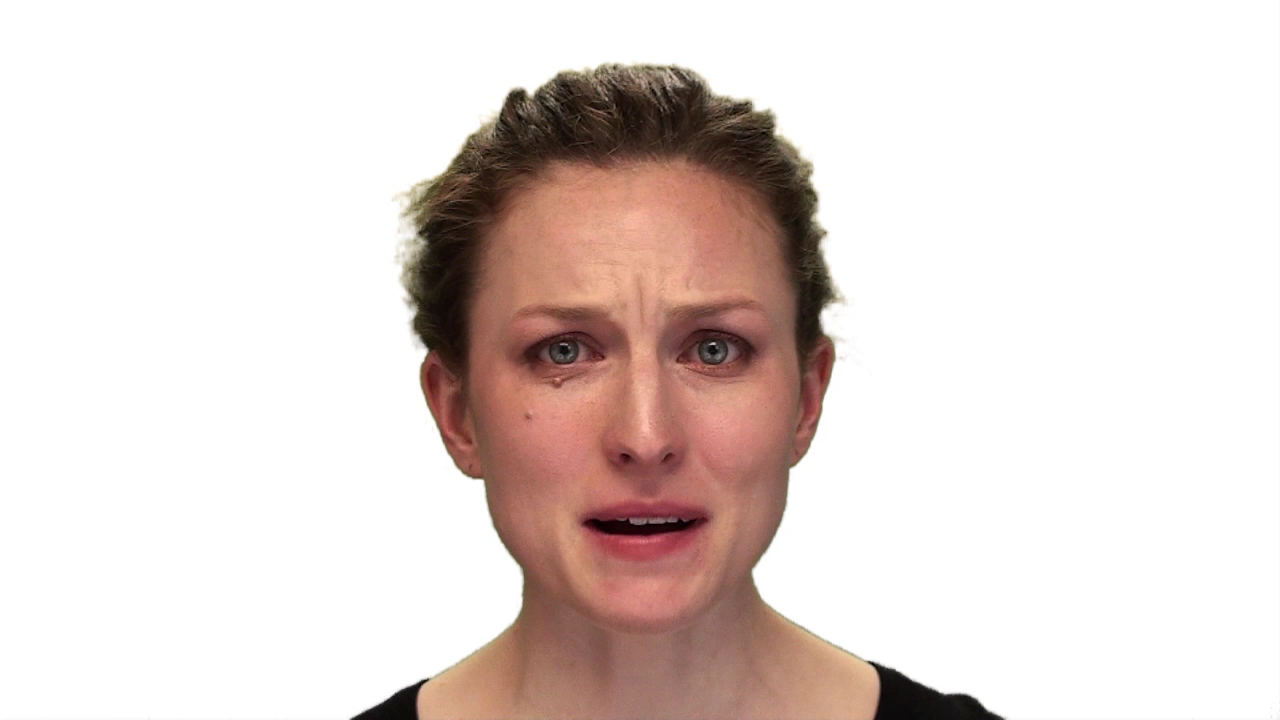}}\hfill\hspace{-2mm}
\frame{\includegraphics[trim=250 50 250 95, clip,width=.24\textwidth]{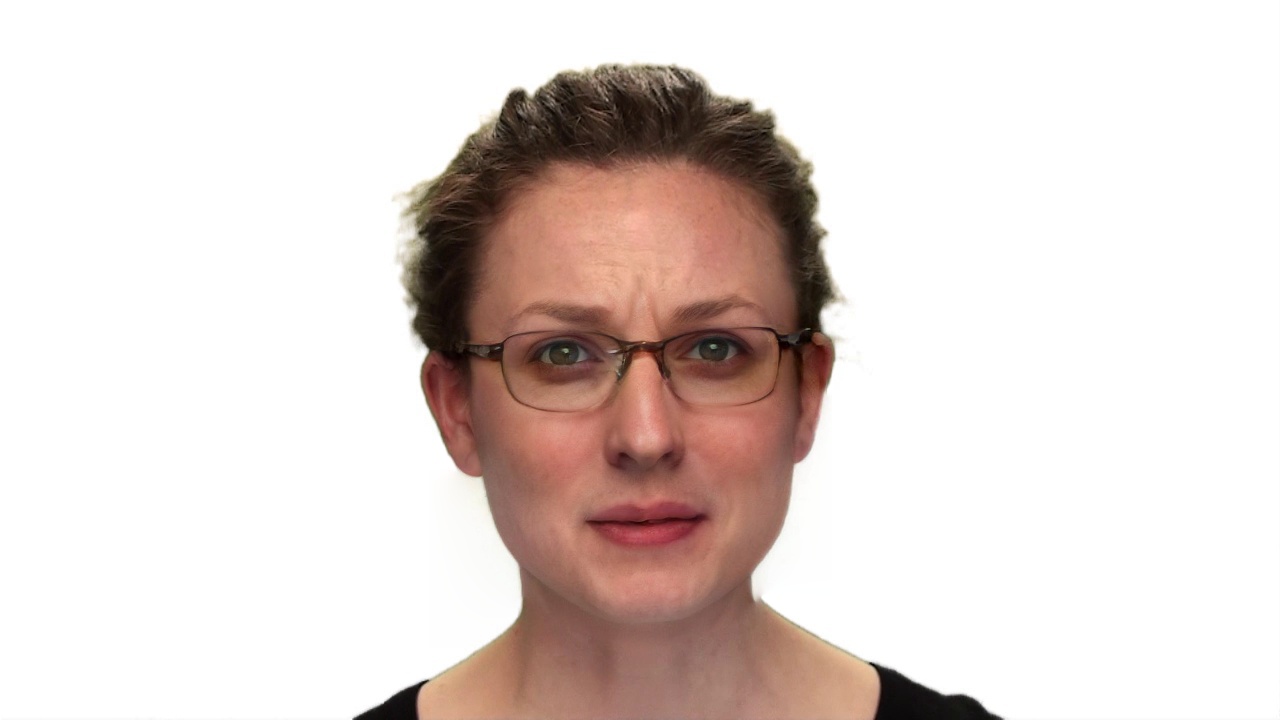}}\hfill\hspace{-2mm}
\frame{\includegraphics[trim=250 50 250 95,  clip,width=.24\textwidth]{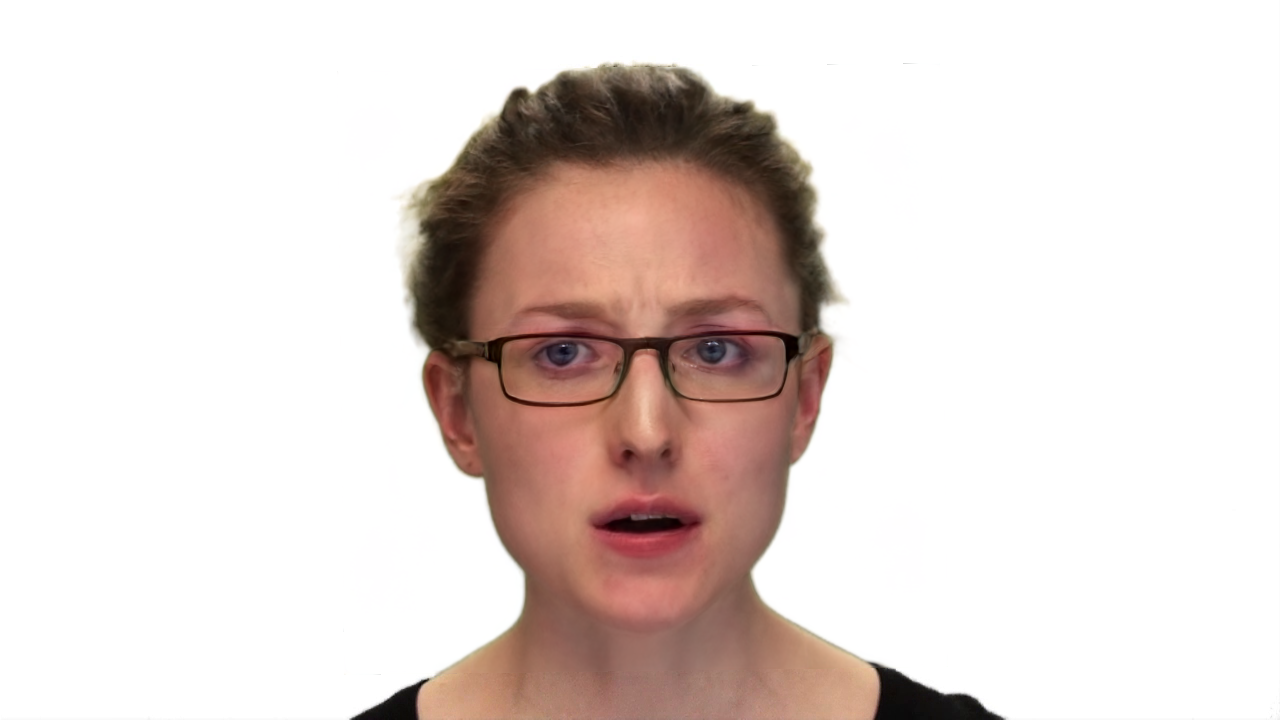}}\\

\mpage{0.2}{\small{Input}} 
\mpage{0.5}{\small{LT}}
\mpage{0.2}{\small{Ours}}
\vspace{-3mm}
\caption{{\textbf{Visual comparison with Latent Transformer (LT)~\cite{yao2021latent}}.} 
LT cannot preserve the person's identity very well. Our method can preserve the identity and achieves a temporal consistent video. 
}
\vspace{-5mm}
\label{fig:comp_yao}
\end{figure*}
\begin{figure}[t]
\centering
{\includegraphics[trim=0 0 0 0, clip,width=.16\textwidth]{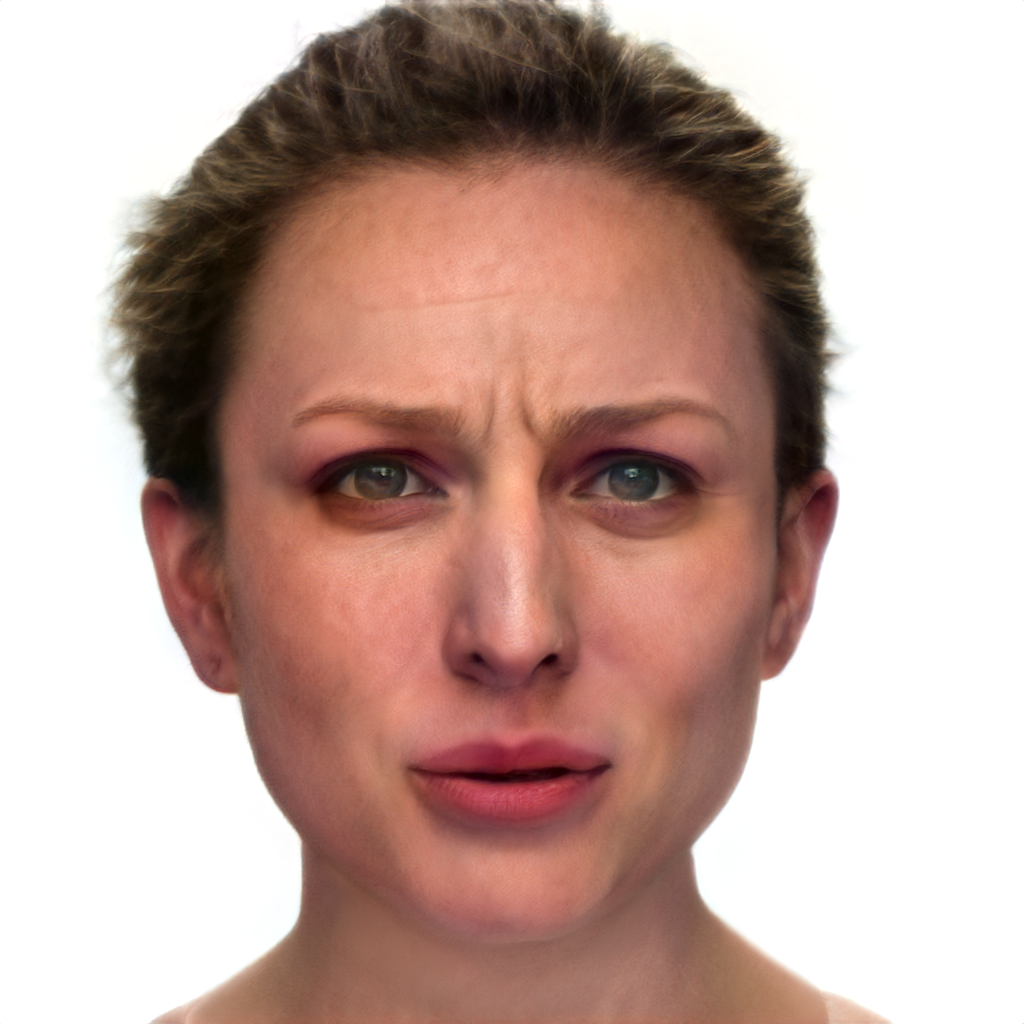}}\hfill 
{\includegraphics[trim=0 0 0 0, clip,width=.16\textwidth]{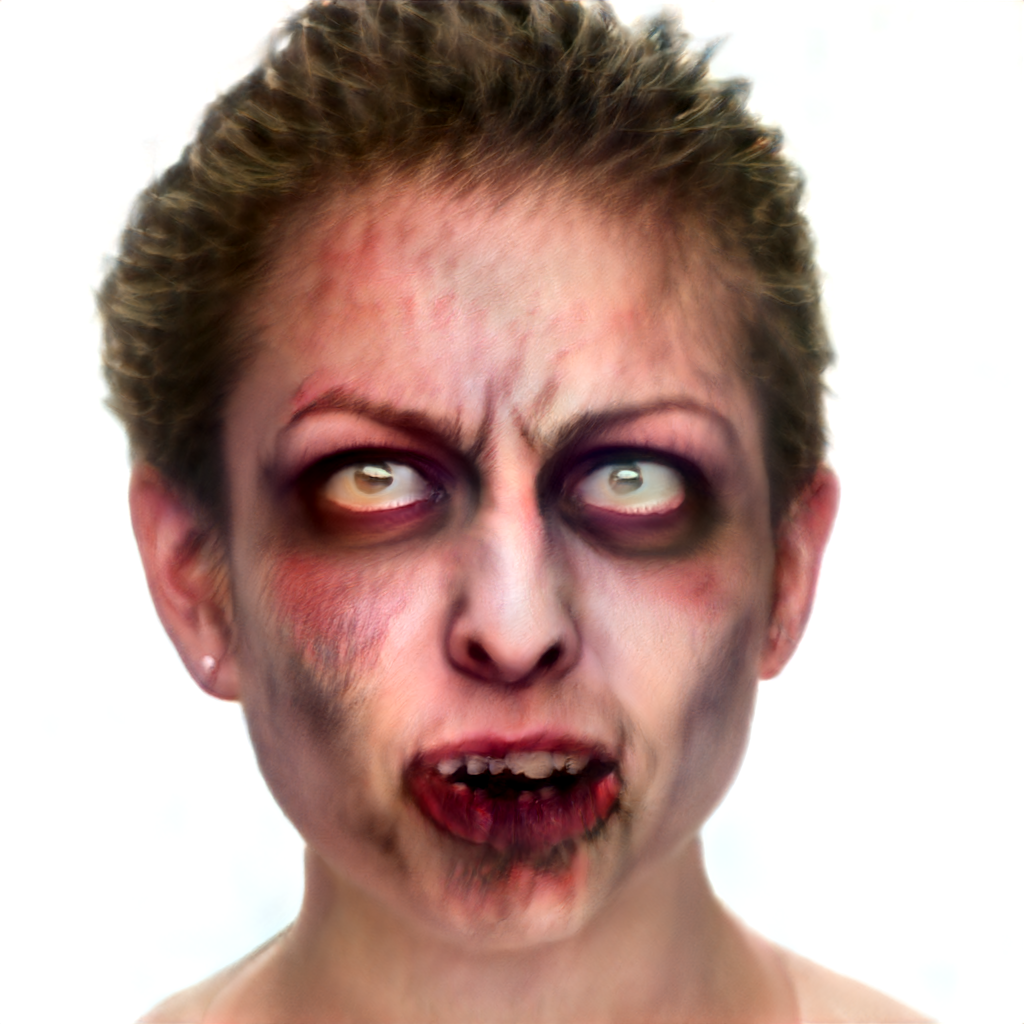}}\hfill
{\includegraphics[trim=0 0 0 0, clip,width=.16\textwidth]{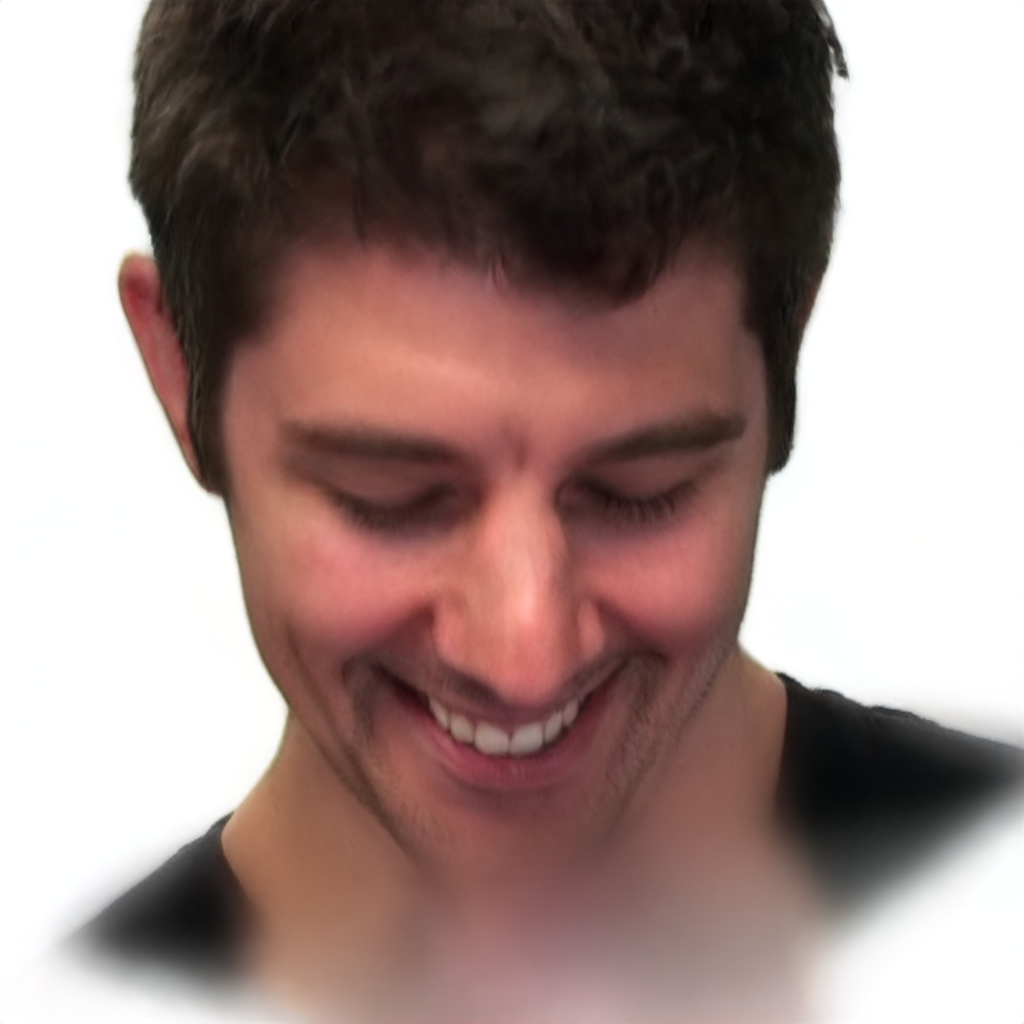}} \hfill
{\includegraphics[trim=0 0 0 0, clip,width=.16\textwidth]{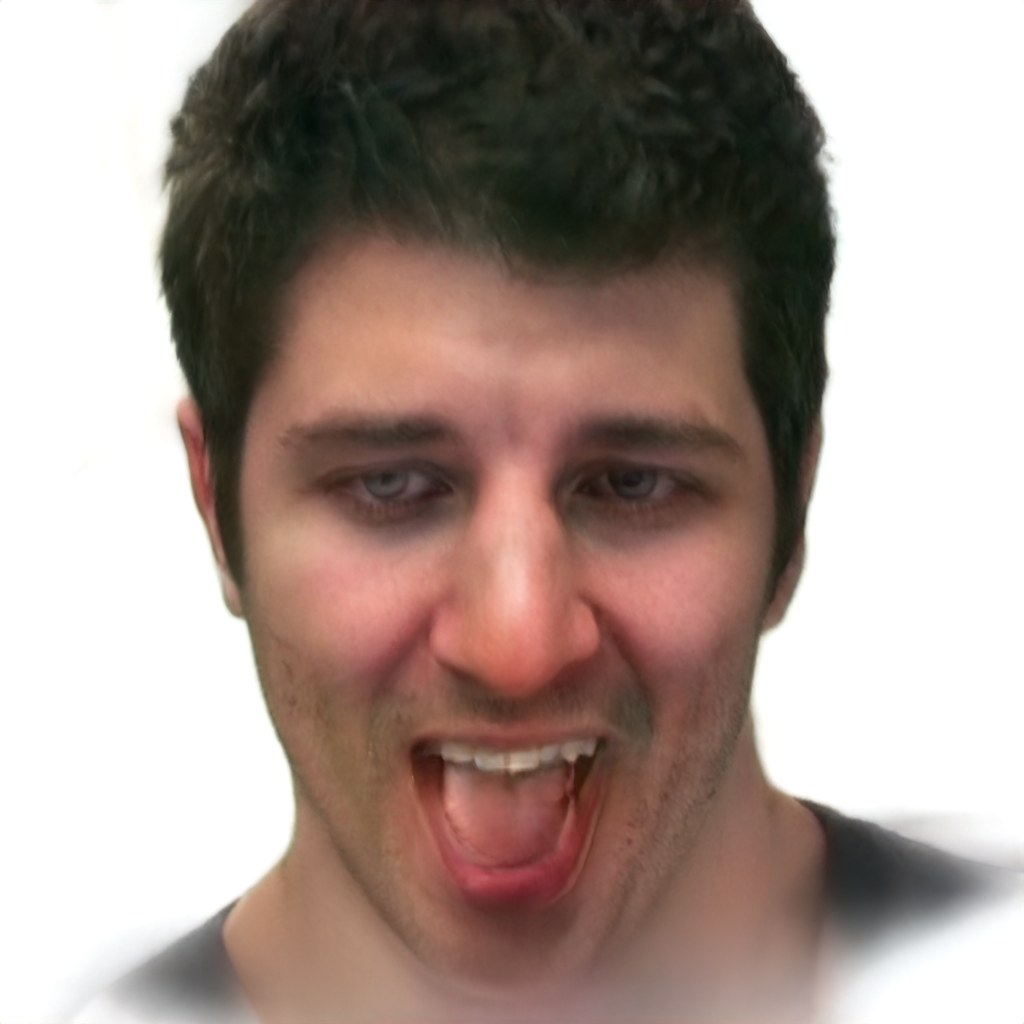}}\\

\mpage{0.01}{}\hspace{3mm}
\mpage{0.3}{(a) ``Zombie''}\hfill
\mpage{0.5}{(b) ``Rare pose''}
\vspace{-2mm}
\caption{\textbf{Limitations}. 
From (a), it can be seen that earrings are added by GAN editing prior to our flow-based temporal consistency approach. Since our approach builds on existing GAN inversion and editing techniques, it will be affected by their quality.
From (b), it can be seen that our method fails when there is a rare pose and a large motion.
}
\label{fig:limitations}
\end{figure}
\vspace{-3mm}
\subsection{Out-of-domain results}\label{sec:out_domain_result}
\begin{table}[t]\setlength{\tabcolsep}{7pt}
  \centering
  \scriptsize	
  \caption{\label{tab:baseline_comp_outdomain}\textbf{Out-of-domain editing comparison.}
  }
    \begin{tabular}{lrrrr}
    \toprule
          & \multicolumn{2}{c}{$E_{warp}\downarrow$} & \multicolumn{2}{c}{LPIPS$\downarrow$} \\
    \midrule
    Direct editing & \multicolumn{2}{c}{0.0098} & \multicolumn{2}{c}{0.0000}  \\
    \midrule
        Editing categories
          & \multicolumn{1}{c}{DVP~\cite{lei2020dvp}} & \multicolumn{1}{c}{Ours} & \multicolumn{1}{c}{DVP} & \multicolumn{1}{c}{Ours} \\
    \midrule
    Sketch          & 0.0036  & 0.0085 & 0.2404 & 0.1314 \\
    Pixar           & 0.0031  & 0.0025 & 0.1074 & 0.1178 \\
    Disney Princess & 0.0040  & 0.0078 & 0.2062 & 0.1204 \\
    Elf             & 0.0042  & 0.0108 & 0.2289 & 0.1310 \\
    Zombie          & 0.0040  & 0.0085 & 0.2033 & 0.1370 \\
    \midrule
    Average perfomance & 0.0038 & 0.0076 & 0.1972 & 0.1275 \\
    \bottomrule
    \end{tabular}%
\end{table}%
\textbf{Setup.} We first invert the videos frame by frame using the Restyle encoder~\cite{alaluf2021restyle} (psp-based~\cite{richardson2021encoding}). 
We then directly apply five different out-of-domain editing effects produced by StyleGAN-NADA~\cite{gal2021stylegannada}. 
We perform our two-phase optimization approach on the directly edited video using Adam optimizer~\cite{kingma2014adam}. 
For phase 1, we set the learning rate to $\alpha_{I}=0.005$, and update the latent codes for 5 epochs. 
In Eqn.~\ref{eq:mlp_update}, we set $\alpha=0.04$ for all the editing directions.
For phase 2, we set the learning rate to $\alpha_{II}=8 \times 10^{-4}$, and finetune $G$ for 5 epochs. We set the regularization weight $\lambda_r$ to $200$.

\topic{Evaluation.} 
Table~\ref{tab:baseline_comp_outdomain} shows that our method decreases the temporal error of the directly edited video.
The primary sources of inconsistency in out-of-domain editing can be seen in the flickering background and the details of the hair. 
We show our visual results in \figref{results}. 
Our method preserves the temporal consistency and maintains the sharpness of the input video. 

\begin{table}[t]\setlength{\tabcolsep}{7pt}
  \centering\scriptsize	
  \vspace{-1mm}
  \caption{\label{tab:baseline_comp_indomain}\textbf{In-domain editing comparison.}
  }
    \begin{tabular}{lrrrrrr}
    \toprule
          & \multicolumn{3}{c}{$E_{warp}\downarrow$} & \multicolumn{2}{c}{LPIPS$\downarrow$} \\
    \midrule
    Direct editing & \multicolumn{3}{c}{0.0076} & \multicolumn{2}{c}{0.0000} \\
    \midrule
        Editing categories
          & \multicolumn{1}{c}{DVP~\cite{lei2020dvp}} & \multicolumn{1}{c}{LT~\cite{yao2021latent}} & \multicolumn{1}{c}{Ours}  & \multicolumn{1}{c}{DVP} & \multicolumn{1}{c}{Ours} \\
    \midrule
    angry      & 0.0033 & - & 0.0032 & 0.2452 & 0.1100 \\
    beard      & 0.0038 & 0.0064 & 0.0030 & 0.2444 & 0.1033 \\
    eyeglasses & 0.0039 & 0.0066 & 0.0034 & 0.1226 & 0.1097 \\
    Depp       & 0.0037 & - & 0.0031 & 0.2452 & 0.2024 \\
    surprised  & 0.0035 & - & 0.0028 & 0.1415 & 0.1012 \\
    \midrule
    Average perfomance & 0.0036 & 0.0065 & 0.0031 & 0.1760 & 0.1253 \\
    \bottomrule
    \end{tabular}%
\end{table}%
\subsection{In-domain editing results}
\label{sec:in_domain_result}
\textbf{Setup.} We first invert the videos frame by frame by using the PTI method~\cite{roich2021pivotal}.
We then directly apply five different semantic editing directions discovered by StyleCLIP mapper~\cite{Patashnik_2021_ICCV}. 
Next, we perform our two-phase optimization approach on the directly edited video using Adam optimizer~\cite{kingma2014adam}. 
For phase 1, we set the learning rate $\alpha_{I}=0.05$, and update $f_{\theta}$ for 10 epochs. In Eqn.~\ref{eq:mlp_update}, we set $\alpha=0.12$ for the ``eyeglasses'', and $\alpha=0.04$ for the rest of the semantic directions.
For phase 2, we set the learning rate of $G$ to $\alpha_{II}=0.0001$, and finetune $G$ for 5 epochs. We set the regularization weight $\lambda_r$ to $200$.

\topic{Evaluation.} 
Table~\ref{tab:baseline_comp_indomain} shows that our approach improves the temporal consistency over the directly edited video baseline by a large margin. 
We also report two overlapped editing directions of Latent Transformer~\cite{yao2021latent} (LT) in Table~\ref{tab:baseline_comp_indomain}. 
LT processes video editing frame by frame without any temporal constraints, it is similar to our direct editing, so we skip the LPIPS scores here. 
Our method also outperforms LT by a large margin. 
When dealing with in-domain editing, the primary source of inconsistency is the details of the newly added attributes, e.g., glasses or beard and some background flickering. 
We show sample visual results in \figref{results}, where the introduced changes are consistent among the different frames. 

\begin{table}[t]\setlength{\tabcolsep}{2.5pt}
  \centering \scriptsize	
  \caption{\label{tab:ablation_study}\textbf{Two-stage optimization strategy ablation study.} 
  }
    \begin{tabular}{cc|cc|cc}
    \toprule
    \multicolumn{2}{c|}{Optimization stage} & \multicolumn{2}{c|}{In-domain editing} & \multicolumn{2}{c}{Out-of-domain editing} \\
    \midrule
    Update ${W_t^{edit}}$ & Update $G$ & $E_{warp} \downarrow$ & LPIPS$\downarrow$ & $E_{warp} \downarrow$ & LPIPS$\downarrow$ \\
    \midrule
      -    &   -    & 0.0076 & 0.0000     & 0.0098 & 0.0000 \\
      \checkmark    &   -    & 0.0064 & 0.2108 & 0.0094 & 0.1428 \\
       -   &   \checkmark    & \textbf{0.0027} & 0.2463 & \textbf{0.0057} & 0.1375 \\
     \checkmark  &   \checkmark    & 0.0031 & \textbf{0.1253} & 0.0076 & \textbf{0.1275} \\
    \bottomrule
    \end{tabular}%
\end{table}%

\subsection{Two-phase optimization strategy ablation study}
\begin{figure}[h]
\vspace{-2mm}


\frame{\includegraphics[trim=0 0 0 0, clip,width=.18\textwidth]{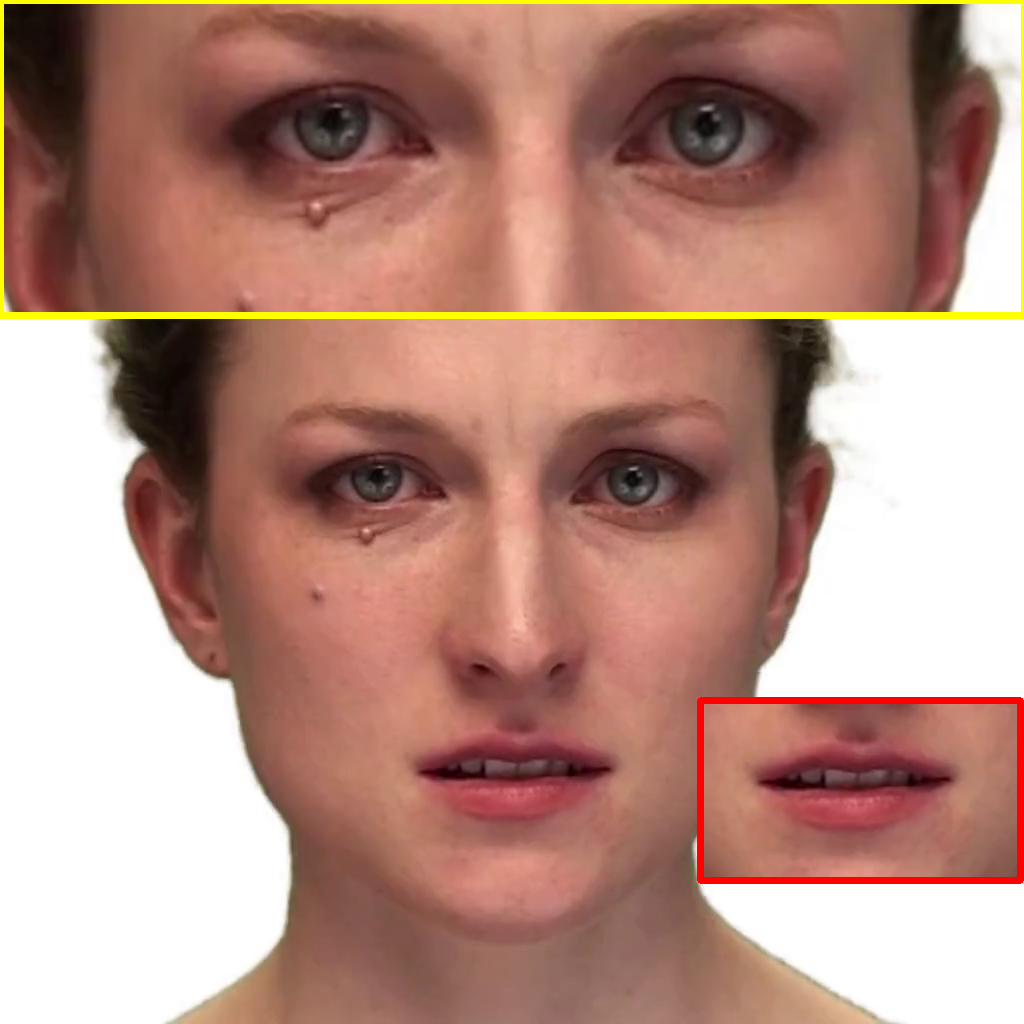}}\hfill
\frame{\includegraphics[trim=0 0 0 0, clip,width=.18\textwidth]{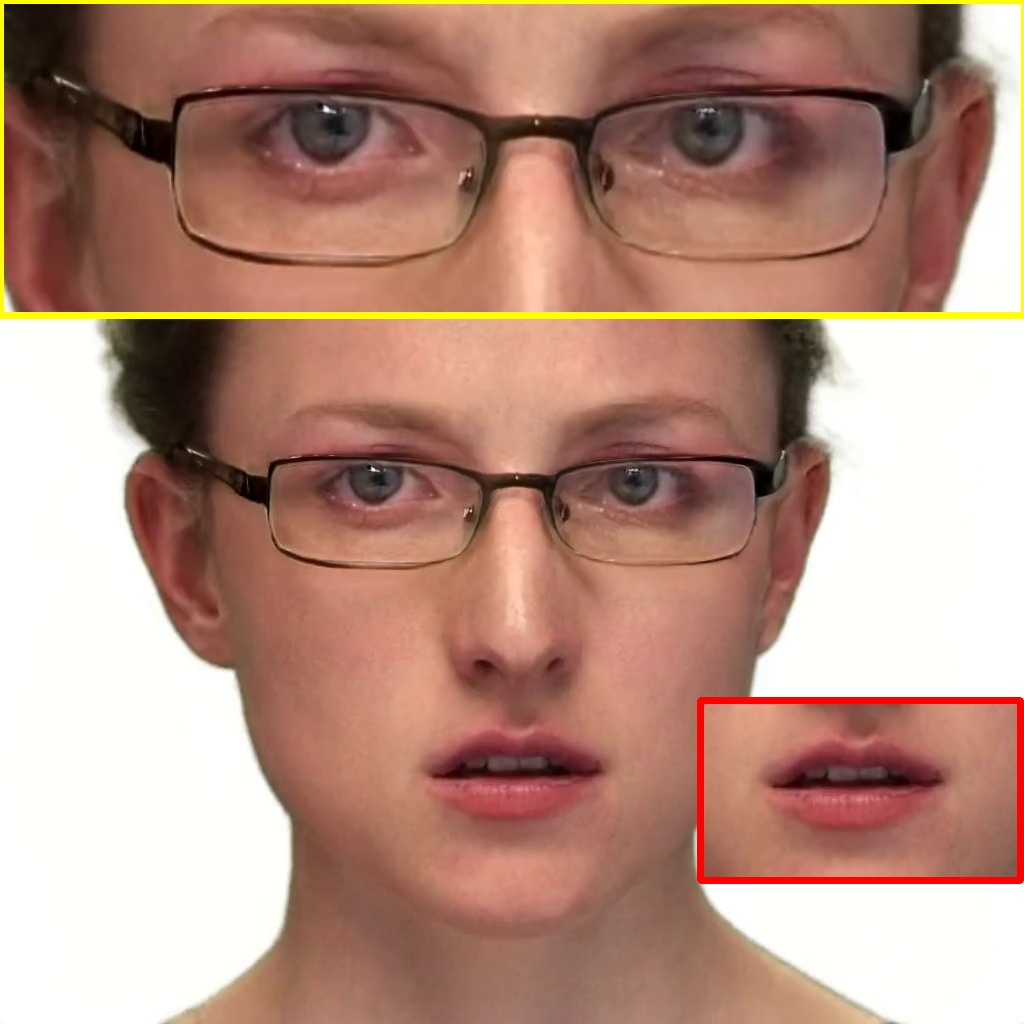}}\hfill
\frame{\includegraphics[trim=0 0 0 0, clip,width=.18\textwidth]{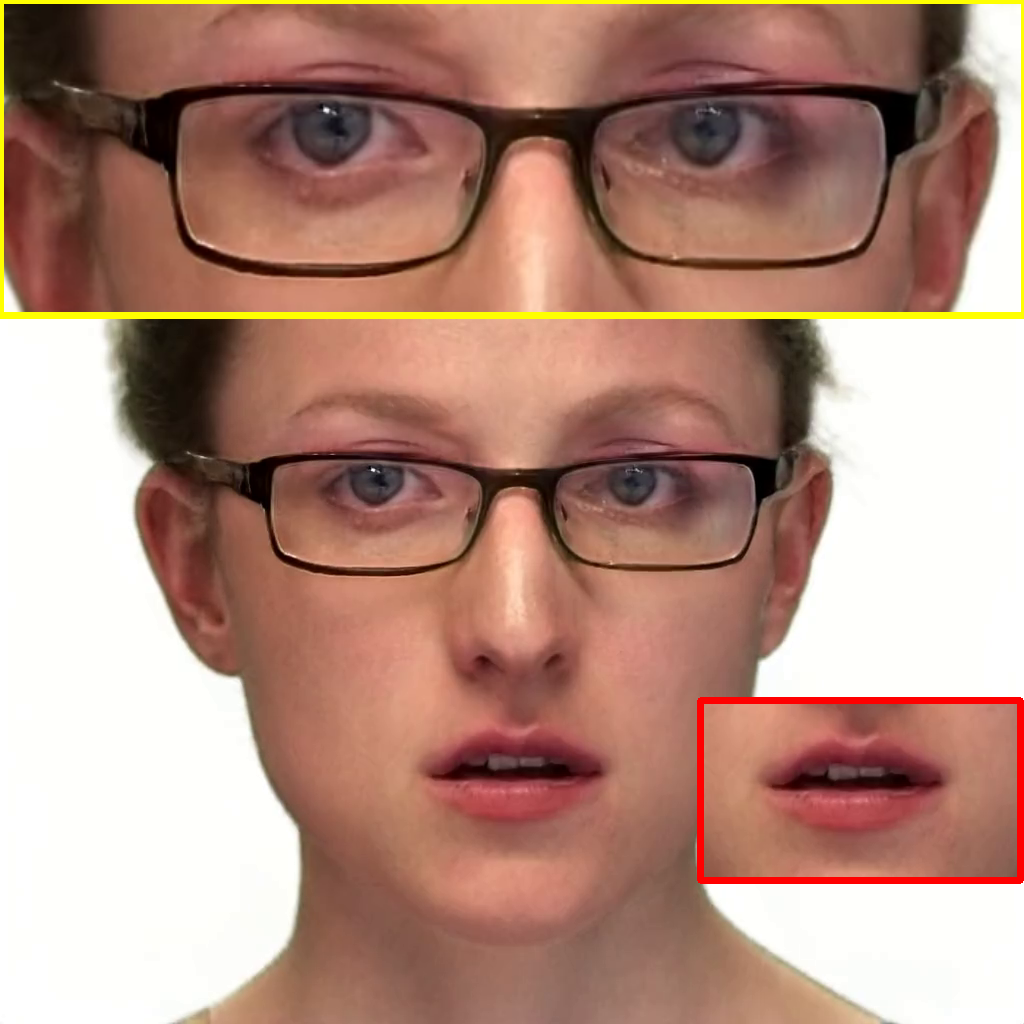}}\hfill
\frame{\includegraphics[trim=0 0 0 0, clip,width=.18\textwidth]{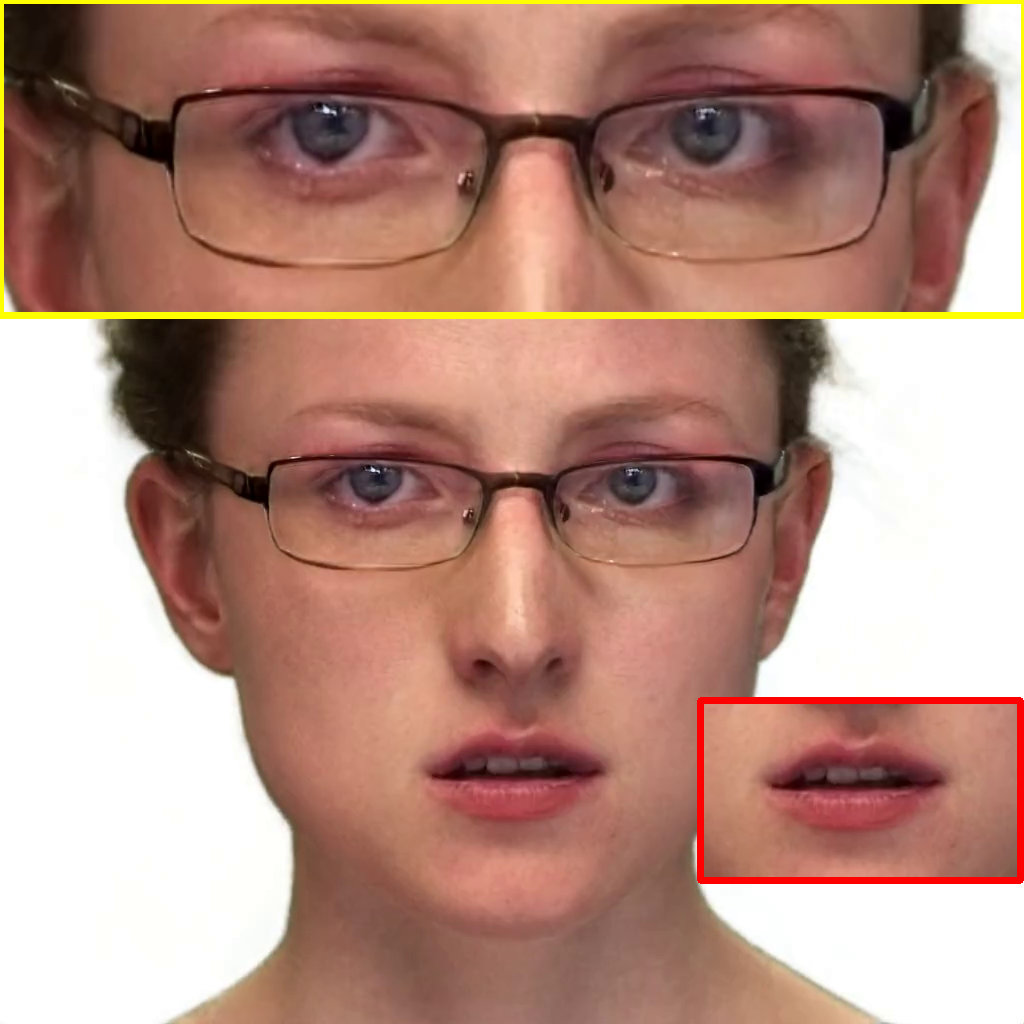}}\hfill
\frame{\includegraphics[trim=0 0 0 0, clip,width=.18\textwidth]{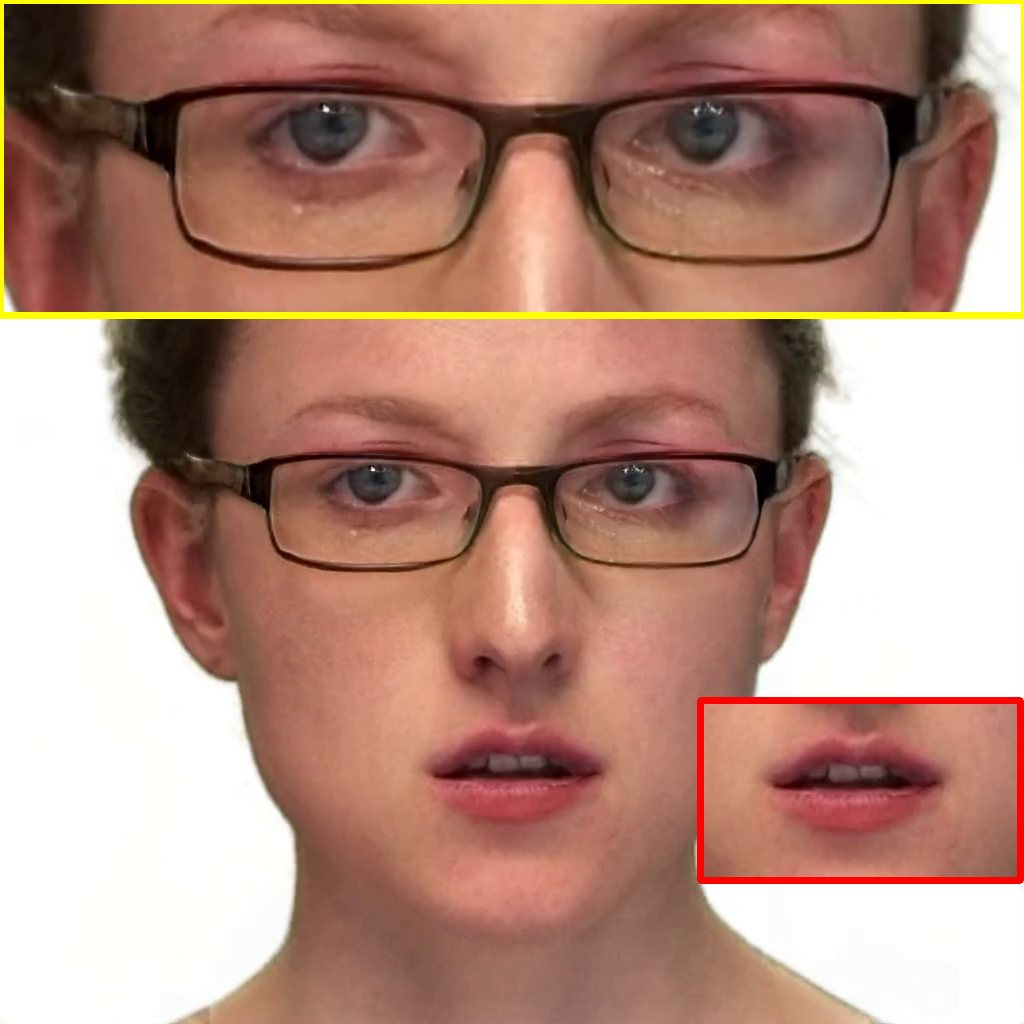}}\\
\mpage{0.15}{\small{Input}}\hfill
\mpage{0.1}{\small{Direct editing}}\hfill
\mpage{0.12}{\small{Only $W$}}\hfill
\mpage{0.12}{\small{Only $G$}}\hfill
\mpage{0.12}{\small{Ours}}
\vspace{-3mm}

\caption{
\textbf{Demonstration of ablation study.}
Unlike other shown approaches, ours achieves temporal consistency while preserving the original attributes. Note the ``eyeglasses'' and the opening of the ``mouth''.
}
\label{fig:ablation_demo}
\end{figure}

We demonstrate the effect of our two-phase optimization strategy of updating the latent codes first and following that with finetuning the generator $G$. 
We compare our two-phase approach to:
(1) No optimization (i.e., direct editing), 
(2) update latent code only (phase 1), and 
(3) finetune generator $G$ only.
We show the results in Table~\ref{tab:ablation_study}.
When we only update generator $G$, we can achieve a low warping error $E_{warp}$. 
However, this is not desirable since finetuning $G$ pushes the video to be consistent globally without modifying the local details. 
Therefore, the output video is different from the directly edited video (i.e., high LPIPS distance). 
Thus, we follow our two-phase optimization of a) updating the latent codes via an MLP $f_{\theta}$ (to improve local consistency), b) finetuning the generator $G$ (to modify the global effect). 
We also show a qualitative comparison in Figure~\ref{fig:ablation_demo}. We can see our method can maintain eyeglasses consistency and recover the mouth affected by direct editing. 

\subsection{Comparison with Latent Transformer}
We compare our method with Latent Transformer (LT)~\cite{yao2021latent}. 
We show a qualitative comparison in Fig.~\ref{fig:comp_yao}. 
LT edits the video by updating the projected latent code \emph{independently} for each frame without using temporal constraints.
Our method, in contrast, uses flow-based loss to improve the temporal consistency, and our second phase uses a perceptual difference mask as a regularization to preserve the facial details other than the edited parts.
As a result, our method can improve temporal consistency and preserve personal identity.

\subsection{Comparison with Deep Video Prior (DVP)}
We compare our method with DVP~\cite{lei2020dvp}, a state-of-the-art blind video consistency approach, using their default setting.  
We show the in-domain editing comparison in Table~\ref{tab:baseline_comp_indomain} and the out-of-domain editing comparison in Table~\ref{tab:baseline_comp_outdomain}. 
For warping error $E_{warp}$, our method achieves improved results for in-domain editing and comparable results for out-of-domain editing. 
However, in terms of LPIPS distance, our visual results are more similar to the directly edited video for both in-domain and out-of-domain editing. 
We show visual comparison in \figref{baseline_comp}. 
DVP can achieve temporally consistent results (i.e., low $E_{warp}$).
However, this is at the cost of losing local details in the ``eyeglasses'' example 
or excessively smoothing the results to get a blurry video as in the ``Disney Princess'' example.

\subsection{Limitations}
We show several limitations of our approach in~\figref{limitations}. 
First, our approach relies on plausible results from existing GAN inversion and editing techniques.
We show an example of added earrings in~\figref{limitations}(a), and an example of a rare pose in~\figref{limitations}(a).
Second, the GANs used in our experiments require the objects to be spatially aligned and thus may not yet be suitable for inverting and editing unconstrained videos. 
Third, our method relies on a high-quality GAN model that may be computationally expensive to train and often require diverse training images.
Our full method (phases 1, 2, and 3) takes 40 minutes on a 150-frame video, on a single NVIDIA P6000 GPU.
\vspace{-3mm}

\section{Conclusions}
\label{sec:conclusion}
We have presented a novel method for video-based semantic editing by leveraging image-based GAN inversion and editing. 
Our approach starts from direct per-frame editing, and we refine the editing results by a flow-based method to minimize the bi-directional photometric loss. 
Our core approach is two-phase, by adjusting the latent codes via an MLP and tuning $G$ to achieve temporal consistency.
We show that our method can achieve temporal consistency and preserve its similarity to the direct editing results. 
Finally, our model-agnostic method is applicable to different GAN inversion and manipulation techniques.

\topic{Potential negative impacts.}
Malicious use of our technique may lead to video manipulation of 
public figures for spreading misinformation.

\clearpage
%
%
\bibliographystyle{splncs04}
\bibliography{main}

\begin{thebibliography}{10}
\providecommand{\url}[1]{\texttt{#1}}
\providecommand{\urlprefix}{URL }
\providecommand{\doi}[1]{https://doi.org/#1}

\bibitem{abdal2019image2stylegan}
Abdal, R., Qin, Y., Wonka, P.: Image2stylegan: How to embed images into the
  stylegan latent space? In: ICCV (2019)

\bibitem{abdal2020image2stylegan++}
Abdal, R., Qin, Y., Wonka, P.: Image2stylegan++: How to edit the embedded
  images? In: CVPR (2020)

\bibitem{abdal2021styleflow}
Abdal, R., Zhu, P., Mitra, N.J., Wonka, P.: Styleflow: Attribute-conditioned
  exploration of stylegan-generated images using conditional continuous
  normalizing flows. ACM Transactions on Graphics (TOG)  \textbf{40}(3),  1--21
  (2021)

\bibitem{afifi2021histogan}
Afifi, M., Brubaker, M.A., Brown, M.S.: Histogan: Controlling colors of
  gan-generated and real images via color histograms. In: Proceedings of the
  IEEE Conference on Computer Vision and Pattern Recognition (2021)

\bibitem{alaluf2021only}
Alaluf, Y., Patashnik, O., Cohen-Or, D.: Only a matter of style: Age
  transformation using a style-based regression model. arXiv preprint
  arXiv:2102.02754  (2021)

\bibitem{alaluf2021restyle}
Alaluf, Y., Patashnik, O., Cohen-Or, D.: Restyle: A residual-based stylegan
  encoder via iterative refinement. In: Proceedings of the IEEE/CVF
  International Conference on Computer Vision (ICCV) (October 2021)

\bibitem{alaluf2022third}
Alaluf, Y., Patashnik, O., Wu, Z., Zamir, A., Shechtman, E., Lischinski, D.,
  Cohen-Or, D.: Third time's the charm? image and video editing with stylegan3.
  arXiv preprint arXiv:2201.13433  (2022)

\bibitem{bau2020semantic}
Bau, D., Strobelt, H., Peebles, W., Zhou, B., Zhu, J.Y., Torralba, A., et~al.:
  Semantic photo manipulation with a generative image prior. arXiv preprint
  arXiv:2005.07727  (2020)

\bibitem{bonneel2015blind}
Bonneel, N., Tompkin, J., Sunkavalli, K., Sun, D., Paris, S., Pfister, H.:
  Blind video temporal consistency. ACM TOG  \textbf{34}(6), ~1--9 (2015)

\bibitem{brock2018large}
Brock, A., Donahue, J., Simonyan, K.: Large scale gan training for high
  fidelity natural image synthesis  (2019)

\bibitem{chai2021latent}
Chai, L., Wulff, J., Isola, P.: Using latent space regression to analyze and
  leverage compositionality in gans. In: International Conference on Learning
  Representations (2021)

\bibitem{chen2017coherent}
Chen, D., Liao, J., Yuan, L., Yu, N., Hua, G.: Coherent online video style
  transfer. In: ICCV (2017)

\bibitem{collins2020editing}
Collins, E., Bala, R., Price, B., Susstrunk, S.: Editing in style: Uncovering
  the local semantics of gans. In: Proceedings of the IEEE/CVF Conference on
  Computer Vision and Pattern Recognition. pp. 5771--5780 (2020)

\bibitem{daras2020your}
Daras, G., Odena, A., Zhang, H., Dimakis, A.G.: Your local gan: Designing two
  dimensional local attention mechanisms for generative models. In: Proceedings
  of the IEEE/CVF Conference on Computer Vision and Pattern Recognition. pp.
  14531--14539 (2020)

\bibitem{gal2021stylegannada}
Gal, R., Patashnik, O., Maron, H., Chechik, G., Cohen-Or, D.: Stylegan-nada:
  Clip-guided domain adaptation of image generators (2021)

\bibitem{goodfellow2014generative}
Goodfellow, I., Pouget-Abadie, J., Mirza, M., Xu, B., Warde-Farley, D., Ozair,
  S., Courville, A., Bengio, Y.: Generative adversarial nets. In: Advances in
  neural information processing systems. pp. 2672--2680 (2014)

\bibitem{gu2020image}
Gu, J., Shen, Y., Zhou, B.: Image processing using multi-code gan prior. In:
  Proceedings of the IEEE/CVF conference on computer vision and pattern
  recognition. pp. 3012--3021 (2020)

\bibitem{gulrajani2017improved}
Gulrajani, I., Ahmed, F., Arjovsky, M., Dumoulin, V., Courville, A.C.: Improved
  training of wasserstein gans (2017)

\bibitem{guo2020towards}
Guo, J., Zhu, X., Yang, Y., Yang, F., Lei, Z., Li, S.Z.: Towards fast, accurate
  and stable 3d dense face alignment. In: ECCV (2020)

\bibitem{huang2016temporally}
Huang, J.B., Kang, S.B., Ahuja, N., Kopf, J.: Temporally coherent completion of
  dynamic video. ACM TOG  \textbf{35}(6),  1--11 (2016)

\bibitem{huh2020ganprojection}
Huh, M., Zhang, R., Zhu, J.Y., Paris, S., Hertzmann, A.: Transforming and
  projecting images to class-conditional generative networks. In: ECCV (2020)

\bibitem{haerkoenen2020ganspace}
Härkönen, E., Hertzmann, A., Lehtinen, J., Paris, S.: Ganspace: Discovering
  interpretable gan controls. In: Proc. NeurIPS (2020)

\bibitem{jang2021stylecarigan}
Jang, W., Ju, G., Jung, Y., Yang, J., Tong, X., Lee, S.: Stylecarigan:
  caricature generation via stylegan feature map modulation. ACM Transactions
  on Graphics (TOG)  \textbf{40}(4),  1--16 (2021)

\bibitem{karras2018progressive}
Karras, T., Aila, T., Laine, S., Lehtinen, J.: Progressive growing of gans for
  improved quality, stability, and variation. In: International Conference on
  Learning Representations (2018)

\bibitem{karras2020training}
Karras, T., Aittala, M., Hellsten, J., Laine, S., Lehtinen, J., Aila, T.:
  Training generative adversarial networks with limited data. arXiv preprint
  arXiv:2006.06676  (2020)

\bibitem{Karras2020ada}
Karras, T., Aittala, M., Hellsten, J., Laine, S., Lehtinen, J., Aila, T.:
  Training generative adversarial networks with limited data. In: Proc. NeurIPS
  (2020)

\bibitem{Karras2021}
Karras, T., Aittala, M., Laine, S., H\"ark\"onen, E., Hellsten, J., Lehtinen,
  J., Aila, T.: Alias-free generative adversarial networks. In: Proc. NeurIPS
  (2021)

\bibitem{karras2019style}
Karras, T., Laine, S., Aila, T.: A style-based generator architecture for
  generative adversarial networks. In: CVPR (2019)

\bibitem{karras2020analyzing}
Karras, T., Laine, S., Aittala, M., Hellsten, J., Lehtinen, J., Aila, T.:
  Analyzing and improving the image quality of stylegan. In: CVPR (2020)

\bibitem{kasten2021layered}
Kasten, Y., Ofri, D., Wang, O., Dekel, T.: Layered neural atlases for
  consistent video editing. ACM TOG  (2021)

\bibitem{kingma2014adam}
Kingma, D.P., Ba, J.: Adam: A method for stochastic optimization. arXiv
  preprint arXiv:1412.6980  (2014)

\bibitem{krizhevsky2012imagenet}
Krizhevsky, A., Sutskever, I., Hinton, G.E.: Imagenet classification with deep
  convolutional neural networks. Advances in neural information processing
  systems  \textbf{25},  1097--1105 (2012)

\bibitem{kwong2021unsupervised}
Kwong, S., Huang, J., Liao, J.: Unsupervised image-to-image translation via
  pre-trained stylegan2 network. IEEE Transactions on Multimedia  (2021)

\bibitem{lai2018learning}
Lai, W.S., Huang, J.B., Wang, O., Shechtman, E., Yumer, E., Yang, M.H.:
  Learning blind video temporal consistency. In: ECCV (2018)

\bibitem{Lai-ECCV-2018}
Lai, W.S., Huang, J.B., Wang, O., Shechtman, E., Yumer, E., Yang, M.H.:
  Learning blind video temporal consistency. In: European Conference on
  Computer Vision (2018)

\bibitem{lei2020dvp}
Lei, C., Xing, Y., Chen, Q.: Blind video temporal consistency via deep video
  prior. In: Advances in Neural Information Processing Systems (2020)

\bibitem{li2021dystyle}
Li, B., Cai, S., Liu, W., Zhang, P., Hua, M., He, Q., Yi, Z.: Dystyle: Dynamic
  neural network for multi-attribute-conditioned style editing. arXiv preprint
  arXiv:2109.10737  (2021)

\bibitem{liu2020learning}
Liu, Y.L., Lai, W.S., Yang, M.H., Chuang, Y.Y., Huang, J.B.: Learning to see
  through obstructions. In: CVPR (2020)

\bibitem{livingstone2018ryerson}
Livingstone, S.R., Russo, F.A.: The ryerson audio-visual database of emotional
  speech and song (ravdess): A dynamic, multimodal set of facial and vocal
  expressions in north american english. PloS one  \textbf{13}(5),  e0196391
  (2018)

\bibitem{luo2017learning}
Luo, J., Xu, Y., Tang, C., Lv, J.: Learning inverse mapping by autoencoder
  based generative adversarial nets. In: International Conference on Neural
  Information Processing. pp. 207--216. Springer (2017)

\bibitem{mao2017least}
Mao, X., Li, Q., Xie, H., Lau, R.Y., Wang, Z., Paul~Smolley, S.: Least squares
  generative adversarial networks. In: Proceedings of the IEEE international
  conference on computer vision. pp. 2794--2802 (2017)

\bibitem{miyato2018spectral}
Miyato, T., Kataoka, T., Koyama, M., Yoshida, Y.: Spectral normalization for
  generative adversarial networks  (2018)

\bibitem{Nitzan2020FaceID}
Nitzan, Y., Bermano, A., Li, Y., Cohen-Or, D.: Face identity disentanglement
  via latent space mapping. ACM Transactions on Graphics (TOG)  \textbf{39},  1
  -- 14 (2020)

\bibitem{Patashnik_2021_ICCV}
Patashnik, O., Wu, Z., Shechtman, E., Cohen-Or, D., Lischinski, D.: Styleclip:
  Text-driven manipulation of stylegan imagery. In: Proceedings of the IEEE/CVF
  International Conference on Computer Vision (ICCV). pp. 2085--2094 (October
  2021)

\bibitem{raj2019gan}
Raj, A., Li, Y., Bresler, Y.: Gan-based projector for faster recovery with
  convergence guarantees in linear inverse problems. In: Proceedings of the
  IEEE/CVF International Conference on Computer Vision. pp. 5602--5611 (2019)

\bibitem{rav2008unwrap}
Rav-Acha, A., Kohli, P., Rother, C., Fitzgibbon, A.: Unwrap mosaics: A new
  representation for video editing. ACM TOG  (2008)

\bibitem{rho2022neural}
Rho, D., Cho, J., Ko, J.H., Park, E.: Neural residual flow fields for efficient
  video representations. arXiv preprint arXiv:2201.04329  (2022)

\bibitem{richardson2021encoding}
Richardson, E., Alaluf, Y., Patashnik, O., Nitzan, Y., Azar, Y., Shapiro, S.,
  Cohen-Or, D.: Encoding in style: a stylegan encoder for image-to-image
  translation. In: IEEE/CVF Conference on Computer Vision and Pattern
  Recognition (CVPR) (June 2021)

\bibitem{roich2021pivotal}
Roich, D., Mokady, R., Bermano, A.H., Cohen-Or, D.: Pivotal tuning for
  latent-based editing of real images. arXiv preprint arXiv:2106.05744  (2021)

\bibitem{saha2021loho}
Saha, R., Duke, B., Shkurti, F., Taylor, G.W., Aarabi, P.: Loho: Latent
  optimization of hairstyles via orthogonalization. In: Proceedings of the
  IEEE/CVF Conference on Computer Vision and Pattern Recognition. pp.
  1984--1993 (2021)

\bibitem{shen2020interfacegan}
Shen, Y., Yang, C., Tang, X., Zhou, B.: Interfacegan: Interpreting the
  disentangled face representation learned by gans. TPAMI  (2020)

\bibitem{shen2021closedform}
Shen, Y., Zhou, B.: Closed-form factorization of latent semantics in gans. In:
  CVPR (2021)

\bibitem{teed2020raft}
Teed, Z., Deng, J.: Raft: Recurrent all-pairs field transforms for optical
  flow. arXiv preprint arXiv:2003.12039  (2020)

\bibitem{tewari2020pie}
Tewari, A., Elgharib, M., Bernard, F., Seidel, H.P., P{\'e}rez, P.,
  Zollh{\"o}fer, M., Theobalt, C., et~al.: Pie: Portrait image embedding for
  semantic control. arXiv preprint arXiv:2009.09485  (2020)

\bibitem{tewari2020stylerig}
Tewari, A., Elgharib, M., Bharaj, G., Bernard, F., Seidel, H.P., P{\'e}rez, P.,
  Zollhofer, M., Theobalt, C.: Stylerig: Rigging stylegan for 3d control over
  portrait images. In: Proceedings of the IEEE/CVF Conference on Computer
  Vision and Pattern Recognition. pp. 6142--6151 (2020)

\bibitem{tov2021designing}
Tov, O., Alaluf, Y., Nitzan, Y., Patashnik, O., Cohen-Or, D.: Designing an
  encoder for stylegan image manipulation. ACM Transactions on Graphics (TOG)
  \textbf{40}(4),  1--14 (2021)

\bibitem{tzaban2022stitch}
Tzaban, R., Mokady, R., Gal, R., Bermano, A.H., Cohen-Or, D.: Stitch it in
  time: Gan-based facial editing of real videos. arXiv preprint
  arXiv:2201.08361  (2022)

\bibitem{viazovetskyi2020stylegan2}
Viazovetskyi, Y., Ivashkin, V., Kashin, E.: Stylegan2 distillation for
  feed-forward image manipulation. In: European Conference on Computer Vision.
  pp. 170--186. Springer (2020)

\bibitem{wu2021coarse}
Wu, Y., Yang, Y.L., Xiao, Q., Jin, X.: Coarse-to-fine: facial structure editing
  of portrait images via latent space classifications. ACM Transactions on
  Graphics (TOG)  \textbf{40}(4),  1--13 (2021)

\bibitem{Wu_2021_CVPR}
Wu, Z., Lischinski, D., Shechtman, E.: Stylespace analysis: Disentangled
  controls for stylegan image generation. In: Proceedings of the IEEE/CVF
  Conference on Computer Vision and Pattern Recognition (CVPR). pp.
  12863--12872 (June 2021)

\bibitem{xia2021gan}
Xia, W., Zhang, Y., Yang, Y., Xue, J.H., Zhou, B., Yang, M.H.: Gan inversion: A
  survey. arXiv preprint arXiv:2101.05278  (2021)

\bibitem{yao2021latent}
Yao, X., Newson, A., Gousseau, Y., Hellier, P.: A latent transformer for
  disentangled face editing in images and videos. In: Proceedings of the
  IEEE/CVF international conference on computer vision. pp. 13789--13798 (2021)

\bibitem{yuksel2021latentclr}
Y{\"u}ksel, O.K., Simsar, E., Er, E.G., Yanardag, P.: Latentclr: A contrastive
  learning approach for unsupervised discovery of interpretable directions.
  arXiv preprint arXiv:2104.00820  (2021)

\bibitem{zhang2018unreasonable}
Zhang, R., Isola, P., Efros, A.A., Shechtman, E., Wang, O.: The unreasonable
  effectiveness of deep features as a perceptual metric. In: Proceedings of the
  IEEE conference on computer vision and pattern recognition. pp. 586--595
  (2018)

\bibitem{zhu2020indomain}
Zhu, J., Shen, Y., Zhao, D., Zhou, B.: In-domain gan inversion for real image
  editing. In: Proceedings of European Conference on Computer Vision (ECCV)
  (2020)

\bibitem{zhu2016generative}
Zhu, J.Y., Kr{\"a}henb{\"u}hl, P., Shechtman, E., Efros, A.A.: Generative
  visual manipulation on the natural image manifold. In: ECCV (2016)

\end{thebibliography}
\end{document}